\documentclass[acmtocl,acmnow]{acmtrans2m}

\usepackage{epsfig}
\usepackage{latexsym}
\usepackage{verbatim}
\usepackage{url}

\def\iif{\hbox{\bf if } \; }
\def\tthen{\; \hbox{\bf then } \; }
\def\eelse{\; \hbox{\bf else } \; }
\def\causes{\hbox{\bf causes}}
\def\caused{\hbox{\bf caused}}
\def\executable{\hbox{\bf executable}}

\def\naf{ not \;}
\def\wwhile{\hbox{\bf while } \; }

\def\pick{ \hbox{\bf pick} }
\def\ddo{\; \hbox{\bf do } \; }

\def\Null{\hbox{\bf null}}

\def\until{\hbox{\bf until}}
\def\always{\hbox{\bf always}}
\def\eventually{\hbox{\bf eventually}}
\def\next{\hbox{\bf next}}

\def\aand{\hbox{\bf and}}
\def\oor{\hbox{\bf or}}
\def\negation{\hbox{\bf negation} }

\def\la{\leftarrow}

\def\call{{\cal L}}

\def\cala{{\cal A}}

\def\cala{{\cal A}}

\def\calb{{\cal B}}

\def\initially{\hbox{\bf initially}}
\def\lan{\langle}
\def\ran{\rangle}

\def\beq{\begin{equation}}
\def\eeq#1{\label{#1}\end{equation}}

\def\st{\medskip\noindent}
\def\ni{\noindent}

\def\qed{$\hfill\Box$}

\long\def\COMMENT#1\ENDCOMMENT{\message{(Commented text...)}\par}

\newtheorem{theorem}{Theorem}[section]

\newtheorem{corollary}[theorem]{Corollary}

\newtheorem{lemma}[theorem]{Lemma}
\newdef{definition}[theorem]{Definition}
\newdef{remark}[theorem]{Remark}
\newtheorem{example}[theorem]{Example}

\newtheorem{observation}[theorem]{Observation}

\begin{document}

\markboth{Tran Cao Son, Chitta Baral, Nam Tran, and Sheila
McIlraith} {Domain-Dependent Knowledge in Answer Set Planning}

\title{Domain-Dependent Knowledge in
Answer Set Planning}

\author{
TRAN CAO SON \\
New Mexico State University \\
CHITTA BARAL and NAM TRAN \\
Arizona State University \and
SHEILA MCILRAITH \\
University of Toronto}

\begin{bottomstuff}
Author's address: T.~C.~Son, Computer Science Department, PO Box
30001, MSC CS, New Mexico State University, Las Cruces, NM 88003,
USA. \newline C.~Baral and N.~Tran, Computer Science and
Engineering, Arizona State University, Tempe, AZ 85287, USA.
\newline S.~McIlraith, Department of Computer Science,
University of Toronto,  Toronto, Canada M5S 5H3.
\end{bottomstuff}

\begin{abstract}
In this paper we consider three different kinds of
domain-dependent control knowledge (temporal, procedural and
HTN-based) that are useful in planning. Our approach is
declarative and relies on the language of logic programming with
answer set semantics (AnsProlog*).  AnsProlog* is designed to plan
without control knowledge.  We show how temporal, procedural and
HTN-based control knowledge can be incorporated into AnsProlog* by
the modular addition of a small number of domain-dependent rules,
without the need to modify the planner.  We formally prove the
correctness of our planner, both in the absence and presence of
the control knowledge. Finally, we perform some initial
experimentation that demonstrates the potential reduction in
planning time that can be achieved when procedural domain
knowledge is used to solve planning problems with large plan
length.
\end{abstract}






\maketitle

\section{Introduction and Motivation}

The simplest formulation of planning -- referred to as classical
planning -- entails finding a sequence of actions that takes a
world from a completely known initial state to a state that
satisfies certain goal conditions. The inputs to a corresponding
planner are the descriptions (in a compact description language
such as STRIPS \cite{fik71}) of the effects of actions on the
world, the description of the initial state and the description of
the goal conditions, and the output is a plan (if it exists)
consisting of a sequence of actions.
The complexity of classical planning is known to
be undecidable in the general case \cite{cha87,erol95}. It reduces
to PSPACE-complete for finite and deterministic domains
\cite{byl94}. By making certain assumptions such as fixing the
length of plans, and requiring actions to be deterministic the
complexity reduces to NP-complete.

\st The ability to plan is widely recognized to be an important
characteristic of an intelligent entity.  Thus, when developing
intelligent systems, we often need to incorporate planning
capabilities, despite their inherent complexity.  Since the
complexity is due to the exponential size of the search space,
planning approaches that overcome this complexity require
efficient and intelligent search.  This is at the crux of three
common and successful approaches to planning: (i) using heuristics
\cite{bon00,hof00,blum97} that are derived from the problem description,
(ii) translating the planning problem into a model finding
problem in a suitable logic and using model finding techniques
for that logic \cite{kau98b}, and (iii) using domain-dependent control
knowledge\footnote{This is alternatively referred to in the
literature as `domain-dependent knowledge', `control knowledge',
`domain knowledge', and `domain constraints'. We also sometimes
use these shortened terms in this paper.}
\cite{bacchus00,doherty99:time,nau99}. The use of
domain-dependent control knowledge
has led to several successful planners, including
TLPlan \cite{bacchus00}, TALplan \cite{doherty99:time}
and SHOP \cite{nau99}, which have performed very well on planning
benchmarks.
Strictly speaking, planners that use control knowledge are no longer
considered to be classical planners since they require the addition
of domain-dependent control knowledge to the problem specification.
Nevertheless, such planners
are predicted to be the most scalable types of planning systems
in the long term \cite{AImag}.
In this paper we integrate the second and the third approaches
identified above by translating a planning problem with domain-dependent
control knowledge into a problem of model finding in logic programming.

\st We integrate domain-dependent control knowledge into our
planner in such a way that planning can still be performed without
this extra control knowledge.  The control knowledge may simply improve the
speed with which a plan is generated or may result in the
generation of plans with particular desirable characteristics.
In this respect%
\footnote{We differ from \cite{bacchus00} in another aspect.
Unlike in \cite{bacchus00} where the domain knowledge is used by
the search program thus controlling the search, our domain
knowledge is encoded as a logic program which is directly added to
the logic program encoding planning. In such an approach there is
no guarantee that the added rules will reduce the search during
answer set computation; although our experimentation shows that it
does for large plan lengths. The paper \cite{hua99} also comments
on this aspect.}
our approach is similar in spirit to the planning systems TLPlan
\cite{bacchus00}, and TALplan \cite{doherty99:time}, but differs
from typical Hierarchical Task Network (HTN) planners (e.g., SHOP
\cite{nau99}) because HTN planners require integration of
domain-dependent control knowledge into the specification of the
planning problem.  As such, HTN planning cannot be performed
without the existence of this knowledge, unlike our planner.

\st In this paper, we explore three kinds of domain control
knowledge: temporal knowledge, procedural knowledge, and HTN-based
knowledge. Our treatment of temporal knowledge is similar to that
used in both the TLPlan and TALplan systems. Our formulation of
procedural knowledge is inspired by GOLOG, referred to
alternatively as a logic programming language, or an action
execution language \cite{rei96}. Although our procedural knowledge
is similar to the syntax of a GOLOG program, how this knowledge is
used in planning is quite different. Similarly, our formulation of
HTN-based knowledge is inspired by the partial-ordering constructs
used in HTN planners, but our use of this type of knowledge during
planning is very different from the workings of HTN planners. The
main difference is that both GOLOG programming and HTN planning
rely on the existence of domain-dependent control knowledge within
the problem specification and cannot perform classical planning in
the absence of this knowledge.  In contrast, our approach, which
is similar to the approach in \cite{bacchus00}, separates the
planner module from the domain knowledge (encoding temporal,
procedural, or HTN-based knowledge), and can plan independent of
the domain knowledge.

\st To achieve our goal of planning using domain-dependent control
knowledge, an important first step is to be able to both reason about actions
and their effects on the world, and represent and reason about
domain-dependent control knowledge. This leads to the question of choosing an
appropriate language for both reasoning and representation tasks.
For this we choose the action language $\calb$ from
\cite{gl98} and the language of logic programming
with answer set semantics (AnsProlog*) \cite{chitta-book}, also
referred to as A-Prolog \cite{gelfond-aij02}. We
discuss our choice on $\calb$ in Section \ref{s2}. We selected
AnsProlog* over other action languages for a number of important
reasons, many of which are listed below.  These points
are elaborated upon in \cite{chitta-book}.

\begin{itemize}

\item AnsProlog* is a non-monotonic language that is suitable for
knowledge representation.  It is especially well-suited to
reasoning in the presence of incomplete knowledge.

\item The non-classical constructs give a structure to
AnsProlog* programs and statements, such as a head and a body,
which allows us to define various subclasses of the language,
each with different
complexity and expressivity properties \cite{dantsin99}. The
subclass of AnsProlog* programs in which no classical negation
is allowed has the same complexity as propositional logic, but
with added expressivity.  The most general class of AnsProlog*
programs, which allows ``{\bf or}'' in the head, has the
complexity and expressivity of the seemingly more complicated
default logic \cite{rei80}.  In general,
AnsProlog* is syntactically simpler than other non-monotonic logics
while equally as expressive as many.

\item There exists a sizable body of ``building block'' results
about AnsProlog* which we may leverage both in knowledge
representation tasks and in the analysis of the correctness of the
representations.
This includes result about composition of several AnsProlog*
programs so that certain original conclusions are preserved
(referred to as `restricted monotonicity'), a transformation of a
program so that it can deal with incomplete information, abductive
assimilation of new knowledge, language independence and
tolerance, splitting  an AnsProlog* program to smaller components
for computing its answer sets, and proving properties about the
original program.

\item There exist several efficient AnsProlog* interpreters
\cite{eiter98a,sim02} and AnsProlog* has been shown to be useful
in several application domains other than knowledge representation
and planning. This includes policy description, product
configuration, cryptography and encryption, wire routing, decision
support in a space shuttle and its `if'--`then' structure has been
found to be intuitive for knowledge encoding from a human expert
point of view.

\item Finally, AnsProlog* has already been used in planning
\cite{sub95,dimo97,lif99}, albeit in the absence of domain-dependent
control
knowledge. In this regard AnsProlog* is suitable for concisely
expressing the effects of actions and static causal relations
between fluents. Note that concise expression of the effects of actions
requires addressing the `frame problem' which was one of
the original motivation behind the development of non-monotonic
logics. Together with its ability to enumerate possible action
occurrences AnsProlog* is a suitable candidate for model-based
planning, and falls under category (ii) (above) of successful
approaches to planning.

\end{itemize}

\noindent As evident from our choice of language, our main focus
in this paper is the knowledge representation aspects of planning
using domain-dependent control knowledge. In particular, our concerns
includes:

\begin{itemize}

\item the ease of expressing effects of actions on the world, and
reasoning about them,

\item the ease of expressing and reasoning about various kinds of
domain constraints,

\item the ease of adding new kinds of domain constraints, and

\item correctness results for the task of planning
using an AnsProlog* representation that includes domain
constraints.

\end{itemize}

\noindent We also perform a limited number efficiency experiments, but
leave more detailed experimentation to future work.

\st With the above focus, the contributions of the paper and the
outline of the paper is as follows:

\begin{enumerate}

\item In Section~\ref{asp-sec} we encode planning (without domain
constraints) using AnsProlog* in the presence of both dynamic effects
of actions and static causal laws, and with goals expressed as a conjunction
of fluent literals. We then formally prove the relationship between
valid trajectories of the action theory and answer sets of the
encoded program. Our approach is similar to \cite{lif99e,eit00}
but differs from \cite{sub95,dimo97,lif99}. The main difference
between our formulation and earlier AnsProlog* encodings in
\cite{sub95,dimo97,lif99} is in our use of static causal laws, and
our consideration of trajectories instead of plans. Our
trajectories are similar to histories in \cite{lif99e} and to
optimistic plans in \cite{eit00}. The reason we relate answer sets
to trajectories rather than relating them to plans is because in
the presence of static
causal laws the effects of actions may be non-deterministic.

\item In Section~\ref{s31} we show how to incorporate
temporal constraints for planning into our formulation of
the planning problem.
Incorporating temporal constraints simply requires
the addition of a few more rules, illustrating
the declarative nature and elaboration tolerance of our approach.
We define formulas for representing temporal
constraints and specify when a trajectory satisfies a temporal
constraint. We then formally prove the relationship
between valid trajectories of the action theory satisfying the
temporal constraints, and answer sets of the updated program. Our
approach differs from \cite{bacchus00,doherty99:time} in that we
use AnsProlog* for both the basic encoding of planning and the
temporal constraints, while the planners in
\cite{bacchus00,doherty99:time} are written in procedural
languages. Preliminary experiments show that our approach is less 
efficient than TLPlan and TALPlan. Nevertheless, both these systems are
highly optimized, so the poorer performance may simply reflect
the lack of optimizations in our implementation. On the other 
hand, our use of AnsProlog* facilitates the provision of correctness
proofs, which is one of our major concerns. Neither
of \cite{bacchus00,doherty99:time} provide correctness proofs of
their planners.

\item In Section~\ref{s32} we consider the use of procedural domain
knowledge in planning.  An example of a procedural domain
knowledge is a program written as {\tt a$_1$; a$_2$; (a$_3$ $|$
a$_4$ $|$ a$_5$);f?}. This program tells the planner that it
should make a plan where $a_1$ is the first action, $a_2$ is the
second action and then it should choose one of $a_3$, $a_4$ or
$a_5$ such that after the plan's execution $f$ will be true.

\st We define programs representing procedural domain knowledge
and specify when a trajectory is a trace of such a program. We then show
how to incorporate the use of procedural domain knowledge in
planning to the initial planning formulation described in item
(1.). As in (2.) the incorporation involves only the addition of a
few more rules. We then formally prove the relation between valid
trajectories of the action theory satisfying the procedural domain
knowledge, and answer sets of the updated program. We also present
experimental results (Section~\ref{s34}) showing the improvement
in planning time due to using such knowledge over planning in the
absence of such knowledge.

\item In Section~\ref{s33} we motivate the need for additional
constructs from HTN planning to express domain knowledge and
integrate features of HTNs with procedural constructs to develop a
more general language for domain knowledge. We then define trace
of such general programs and show how to incorporate them in
planning. We then formally prove the relation between valid
trajectories of the action theory satisfying the general programs
containing both procedural and HTN constructs, and answer sets of
the updated program. To the best of our knowledge this is the
first time an integration of HTN and procedural constructs has
been proposed for use in planning.

\item As noted above, a significant contribution of our work is the
suite of correctness proofs for our AnsProlog* formulations. All the proofs
appear in Appendix A.  For completeness Appendix B presents results
concerning AnsProlog* that we use in the Appendix A proofs.

\end{enumerate}

\noindent In regards to closely related work, although
satisfiability planning (see e.g., \cite{kau92,kau96b,kau96,kau98b})
has been studied quite a bit, those papers do not have correctness
proofs and do not use the varied domain constraints that we use in
this paper.

\st We now start with some preliminaries and background material
about reasoning about actions and AnsProlog*, which will be used
in the rest of the paper.

\section{Preliminaries and Background}
\label{s2}

In this section, we review the basics of the action
description language $\calb$, the answer set semantics of logic
programs (AnsProlog), and key features of problem solving using
AnsProlog.

\subsection{Reasoning about actions: the action description
language $\calb$}

Recall that planning involves finding a sequence of actions that
takes a world from a given initial state to a state that satisfies
certain goal conditions. To do planning, we must first be able to
reason about the impact of a single action on a world. This is
also the first step in `reasoning about actions'. In general,
reasoning about action involves defining a transition function
from states (of the world) and actions to sets of states where the
world might be after executing the action. Since explicit
representation of this function would require exponential space in
the size of the number of fluents (i.e., properties of the world),
actions and their effects on the world are described using an
action description language, and the above mentioned transition
function is implicitly defined in terms of that description.
In this paper, we adopt the language $\calb$ \cite{gl98},
which is a subset of the language proposed in \cite{turner97},
for its simple syntax and its capability to represent relationships
between fluents, an important feature lacking in many variants
of the action description language $\cala$ \cite{GL93a}. We note
that the main results of this paper can be used in
answer set planning systems which use other languages for representing
and reasoning about the effects of actions.

\st We now present the basics of the action description language
$\calb$. An action theory in $\calb$ is defined over two disjoint
sets, a set of actions {\bf A} and a set of fluents {\bf F}, which
are defined over a signature $\sigma = \langle \mathbf{O},
\mathbf{AN}, \mathbf{FN} \rangle$ where

\begin{itemize}
\item $\mathbf{O}$ is a finite set of object constants;

\item {\bf AN} is a finite set of {\em action names},
each action name is associated with a number $n$, $n \ge 0$,  which
denotes its arity;  and

\item {\bf FN} is a finite set of {\em fluent names},
each fluent name is associated with a number $n$, $n \ge 0$,  which
denotes its arity.
\end{itemize}

\noindent
An action in {\bf A} is of the form $a(c_1,\ldots,c_n)$
where $a \in \mathbf{AN}$ is a n-ary action name and
$c_i$ is a constant in {\bf O}.
A fluent in {\bf F} is of the form $f(c_1,\ldots,c_m)$
where $f \in \mathbf{FN}$ is an m-ary fluent name and
$c_i$ is a constant in {\bf O}.
For simplicity, we often write
$a$ and $f$ to represent an action or a fluent whenever
it is unambiguous. Furthermore, we
will omit the specification of $\sigma$ when it is clear from
the context.

\st
A {\em fluent literal} is either a fluent $f \in \mathbf{F}$
or its negation $\neg f$.
A {\em fluent formula} is a propositional formula constructed
from fluent literals.
An action theory is a  set of propositions of the following form:
\begin{eqnarray}
 & \caused(\{p_1,\dots,p_n\}, f) \label{static}\\
 & \causes(a,f,\{p_1,\dots,p_n\}) \label{dynamic}\\
 & \executable(a, \{p_1,\dots,p_n\}) \label{exec}\\
 &  \initially(f) \label{init}
\end{eqnarray}
where $f$ and $p_i$'s are fluent literals and $a$ is an
action. (\ref{static}) represents a {\em static causal law}, i.e., a
relationship between fluents\footnote{
   A constraint between fluents can also be represented
   using static causal laws.
   For example, to represent the fact that a door cannot be {\em opened}
   and {\em closed} at the same time, i.e. the fluents {\em opened}
   and {\em closed} cannot be true at the same time, we introduce a
   new fluent, say {\em inconsistent}, and
   represent the constraint by two static causal laws
   $\caused(\{opened,closed\}, inconsistent)$ and
   $\caused(\{opened,closed\}, \neg inconsistent)$.
}. It conveys that whenever the fluent literals
$p_1,\dots,p_n$ hold in a state, that causes $f$ to also hold in
that state. (\ref{dynamic}), referred to as a {\em dynamic causal
law}, represents the (conditional) effect of $a$ while
(\ref{exec}) encodes an executability condition of $a$.
Intuitively, an executability condition of the form (\ref{exec})
states that $a$ can only be executed if $p_i$'s holds. A dynamic
law of the form (\ref{dynamic}) states that $f$ is caused to be
true after the execution of $a$ in any state of the world where
$p_1,\dots,p_n$ are true. Propositions of the form (\ref{init})
are used to describe the initial state. They state that $f$ holds
in the initial state.

\st An {\em action theory} is a pair $(D,\Gamma)$ where $\Gamma$,
called the {\em initial state}, consists of
propositions of the form (\ref{init}) and $D$, called {\em the
domain description}, consists of propositions of the form
(\ref{static})-(\ref{exec}). For convenience, we sometimes denote
the set of propositions of the form (\ref{static}),
(\ref{dynamic}), and (\ref{exec}) by $D_C$, $D_D$, and $D_E$,
respectively.

\begin{example}
\label{ex1}
{\rm
Let us consider a modified version of the suitcase $s$
with two latches from \cite{lin95a}.
We have a suitcase with two latches $l_1$ and $l_2$.
$l_1$ is up and $l_2$ is down.
To open a latch ($l_1$ or $l_2$) we need a corresponding
key ($k_1$ or $k_2$, respectively).
When the two latches are in the up position, the
suitcase is unlocked. When one of the latches is
down, the suitcase is locked.
The signature of this domain consists of
\begin{itemize}
\item $\mathbf{O} = \{l_1,l_2,s,k_1,k_2\}$;
\item $\mathbf{AN} = \{open, close\}$, both action names
are associated with the number 1, and
\item $\mathbf{FN} = \{up, locked, holding\}$,
all fluent names are associated with the number 1.
\end{itemize}
In this domain, we have that
\[
\mathbf{A} = \{open(l_1), open(l_2), close(l_1), close(l_2)\}
\]
and
\[
\mathbf{F} = \{locked(s), up(l_2), up(l_1), holding(k_1),
holding(k_2)\}.
\]

\ni
We now present the propositions describing the domain.

\st
Opening a latch puts it into the up
position. This is represented by the dynamic laws:
\[
\causes(open(l_1), up(l_1), \{\}) \textnormal{ and }
\causes(open(l_2), up(l_2), \{\}).
\]
Closing a latch puts it into the down position.
This can be written as:
\[
\causes(close(l_1), \neg up(l_1), \{\}) \textnormal{ and }
\causes(close(l_2), \neg up(l_2), \{\}).
\]
We can open the latch only when we have the key.
This is expressed by:
\[
\begin{array}{l}
\executable(open(l_1), \{holding(k_1)\})
\textnormal{ and }
\executable(open(l_2), \{holding(k_2)\}).
\end{array}
\]
No condition is required for closing a latch. This is expressed by
the two propositions:
\[
\begin{array}{l}
\executable(close(l_1), \{\})
\textnormal{ and }
\executable(close(l_2), \{\}).
\end{array}
\]
The fact that the suitcase will be unlocked when the
two latches are in the up position
is represented by the static causal law:
\[
\begin{array}{l}
\caused(\{up(l_1), up(l_2)\}, \neg locked(s)).
\end{array}
\]
Finally, to represent the fact that the suitcase will be locked
when either of the two latches is in the down position,
we use the following static laws:
\[
\begin{array}{l}
\caused(\{\neg up(l_1)\}, locked(s)) \textnormal{ and }
\caused(\{\neg up(l_2)\}, locked(s))
\end{array}
\]
The initial state of this domain is given by
\[
\Gamma = \left \{
\begin{array}{lllll}
\initially(up(l_1)) \\
\initially(\neg up(l_2))  \\
\initially(locked(s))  \\
\initially(\neg holding(k_1)) \\
\initially(holding(k_2)) \\
\end{array}
\right \}
\]
\qed
}
\end{example}

\st A domain description given in $\calb$  defines a transition
function from pairs of actions and states to sets of states whose
precise definition is given below. Intuitively, given an action
$a$ and a state $s$, the transition function $\Phi$ defines the
set of states $\Phi(a,s)$ that may be reached after executing the
action $a$ in state $s$. If $\Phi(a,s)$ is an empty set it means
that the execution of $a$ in $s$ results in an error.
We now formally define $\Phi$.

\st Let $D$ be a domain description in $\calb$.
A set of fluent literals is said to be {\em consistent} if it
does not contain $f$ and $\neg f$ for some fluent $f$. An
{\em interpretation } $I$ of the fluents in $D$ is a maximal
consistent set of fluent literals of $D$. A fluent
$f$ is said to be true (resp. false) in $I$ iff $f \in I$ (resp.
$\neg f \in I$).
The truth value of a fluent formula in $I$ is
defined recursively over the propositional connectives in the
usual way. For example, $f \wedge g$ is true in $I$ iff $f$ is
true in $I$ and $g$ is true in $I$. We say that a formula
$\varphi$ holds in $I$ (or $I$ satisfies $\varphi$), denoted by
$I \models \varphi$, if $\varphi$ is true in $I$.

\st Let $u$ be a consistent set of fluent literals and $K$ a
set of static causal laws. We say that $u$ is closed under
$K$ if for every static causal law
$$\caused(\{p_1,\ldots,p_n\},f)$$ in $K$, if
$u \models p_1 \wedge \ldots \wedge p_n$ then $u \models f$. By
$Cl_K(u)$ we denote the least consistent set of literals from $D$
that contains $u$ and is also closed under $K$. It is worth
noting that $Cl_K(u)$ might be undefined. For instance, if $u$
contains both $f$ and $\neg f$ for some fluent $f$, then $Cl_K(u)$
cannot contain $u$ and be consistent; another example is that if
$u = \{f,g\}$ and $K$ contains $$\caused(\{f\}, h)$$ and
$$\caused(\{f, g\}, \neg h),$$ then $Cl_K(u)$ does not exist
because it has to contain both $h$ and $\neg h$, which means that
it is inconsistent.

\st Formally, a {\em state} of $D$ is an interpretation of the
fluents in {\bf F} that is closed under the set of static causal
laws $D_C$ of $D$.

\st An action $a$ is {\em executable} in a state $s$ if there
exists an executability proposition
$$\executable(a,\{f_1,\ldots,f_n\})$$ in $D$ such that $s \models
f_1 \wedge \ldots \wedge f_n$. Clearly, if $$\executable(a,\{\})$$
belongs to $D$, then $a$ is executable in every state of $D$.

\st The {\em direct effect of an action a} in a state $s$ is the
set
$$E(a,s) = \{f \mid
\causes(a, f, \{f_1,\ldots,f_n\}) \in D, s \models f_1 \wedge
\ldots \wedge f_n\}.
$$

\st For a domain description $D$,
$\Phi(a,s)$, the set of states that may be reached by executing
$a$ in $s$, is defined as follows.

\begin{enumerate}
\item If $a$ is executable in $s$, then
\[\Phi(a,s) = \{s' \; \mid \;  \; s' \mbox{ is a state and }s' =
Cl_{D_C}(E(a,s) \cup (s
\cap s'))\}; \]

\item If $a$ is not executable in $s$, then $\Phi(a,s) = \emptyset$.

\end{enumerate}

\ni The intuition behind the above formulation is as follows.  The
direct effects of an action $a$ in a state $s$ are determined by
the dynamic causal laws and are given by $E(a,s)$.  All fluent
literals in $E(a,s)$ must hold in any resulting state. The set $s
\cap s'$ contains the fluent literals of $s$ which continue to
hold by inertia, i.e they hold in $s'$ because they were not
changed by an action. In addition, the resulting state must be
closed under the set of static causal laws $D_C$. These three
aspects are captured by the definition above. Observe that when
$D_C$ is empty and $a$ is executable in state $s$, $\Phi(a,s)$ is
equivalent to the set of states that satisfy $E(a,s)$ and are
closest to $s$ using symmetric difference\footnote{We say $s_1$ is
strictly closer to $s$ than $s_2$ if $s_1 \setminus s \cup s
\setminus s_1 \subset s_2 \setminus s \cup s \setminus s_2$.} as
the measure of closeness~\cite{mcc95a}. Additional explanations
and motivations behind the above definition can be found in
\cite{bar95b1,mcc95a,turner97}.

\st Every domain description $D$ in $\calb$ has a unique
transition function $\Phi$, and we say $\Phi$ is the transition
function of $D$. We illustrate the definition of the transition
function in the next example.

\begin{example}
\label{ex2}
{\rm
For the suitcase domain in Example \ref{ex1},
the initial state given by the set of propositions
\[
\Gamma = \left \{
\begin{array}{lllll}
\initially(up(l_1)) \\
\initially(\neg up(l_2))  \\
\initially(locked(s))  \\
\initially(\neg holding(k_1)) \\
\initially(holding(k_2)) \\
\end{array}
\right.
\]
is
\[
s_0 = \{up(l_1), \neg up(l_2),
    locked(s), \neg holding(k_1), holding(k_2)\}.
\]
\ni In state $s_0$, the three actions $open(l_2)$, $close(l_1)$,
and $close(l_2)$ are executable. $open(l_2)$ is executable since
$holding(k_2)$ is true in $s_0$ while $close(l_1)$ and
$close(l_2)$ are executable since the theory (implicitly) contains
the propositions:
\[
\executable(close(l_1),\{\}) \hspace{1cm} \textnormal{ and }
\hspace{1cm} \executable(close(l_2),\{\})
\]
which indicate that these two actions are always executable. The
following transitions are possible from state $s_0$:
\begin{eqnarray*}
   \{\: up(l_1), up(l_2), \neg locked(s),\neg holding(k_1),
    holding(k_2) \:\}
 & \in & \Phi(open(l_2), s_0).\nonumber\\
   \{\: up(l_1), \neg up(l_2), locked(s),\neg holding(k_1),
  holding(k_2) \:\}
 & \in & \Phi(close(l_2), s_0).\nonumber\\
   \{\: \neg up(l_1), \neg up(l_2), locked(s),\neg 
holding(k_1),holding(k_2)
\:\}
 & \in & \Phi(close(l_1), s_0).\nonumber\\
\end{eqnarray*}
\qed
}
\end{example}

\ni
For a domain description $D$ with transition function $\Phi$,
a sequence $s_0a_0s_1\ldots a_{n-1}s_n$ where $s_i$'s are states
and $a_i$'s are actions is called a {\em trajectory} in $D$ if
$s_{i+1} \in \Phi(s_i, a_{i+1})$ for $i \in \{0,\ldots,n-1\}$.
A trajectory $s_0a_0s_1\ldots a_{n-1}s_n$ achieves a fluent
formula $\Delta$ if $s_n \models \Delta$.

\st A domain description $D$ is \textit{consistent} iff for every
action $a$ and state $s$, if $a$ is executable in $s$, then
$\Phi(a,s) \neq \emptyset$. An action theory $(D,\Gamma)$ is
consistent if $D$ is consistent and $s_0 = \{f \mid
\initially(f)\in \Gamma\}$ is a state of $D$. In what follows, we
will consider only\footnote{We thank one of the anonymous referee
for pointing out that without this assumption, finding a plan
would be $\Sigma_3 {\bf P}$-complete even with respect to a
complete initial state.} consistent action theories.

\st A {\em planning problem} with respect to $\calb$ is specified
by a triple $\lan D, \Gamma, \Delta \ran$ where $(D,\Gamma)$ is an
action theory in $\calb$ and $\Delta$ is a fluent formula (or {\em
goal}), which a goal state must satisfy. A sequence of actions
$a_0,\ldots,a_{m-1}$ is then called a {\em possible plan for
$\Delta$} if there exists a trajectory $s_0 a_0 s_1 \ldots a_{m-1}
s_m$ in $D$ such that $s_0$ and $s_m$ satisfies $\Gamma$ and
$\Delta$, respectively. Note that we define a `possible plan'
instead of a `plan'. This is because the presence of static causal
laws in $D$ allows the possibility that the effects of actions may
be non-deterministic, and planning with non-deterministic actions
has the complexity of $\Sigma_2 {\bf P}$-complete \cite{turner02}
and hence is beyond the expressiveness of AnsProlog. However, if
$D$ is deterministic, i.e., $|\Phi(a,s)| \le 1$ for every pair
$(a,s)$ of actions and states,
then the notions of `possible plan' and `plan' coincide.

\subsection{Logic Programming with answer set semantics (AnsProlog) and its application}

In this section we review AnsProlog (a sub-class of AnsProlog*)
and its applicability to problem solving.

\subsubsection{AnsProlog}

Although the programming language Prolog and the field of logic
programming have been around for several decades, the answer set
semantics of logic programs -- initially referred to as the stable
model semantics, was only proposed by Gelfond and
Lifschitz in 1988 \cite{GL88}. Unlike earlier characterizations of
logic programs where the goal was to find a unique appropriate
`model' of a logic program, the answer set semantics allows the
possibility that a logic program may have multiple appropriate
models, or no appropriate models at all. Initially, some
considered the existence of multiple or no stable models to be
a drawback of stable model semantics, while others
considered it to be a reflection of the poor quality of the program
in question.  Nevertheless, it is this feature of the answer
set semantics \cite{mar99,nie99,lif99} that is key to
the use of AnsProlog for problem solving. We now present the
syntax and semantics of AnsProlog, which we will simply refer to
as a logic program.

\st A logic program $\Pi$ is a set of rules of the form
\begin{equation}
\label{lprule1}
 a_0 \leftarrow a_1,\ldots,a_m,\naf a_{m+1},\ldots,\naf a_n
\end{equation}
or
\begin{equation}
\label{lprule2}
 \bot \leftarrow a_1,\ldots,a_m,\naf a_{m+1},\ldots,\naf a_n
\end{equation}
where $0 \le m \le n$, each $a_i$ is an atom of a first-order
language $\call{\cal P}$, $\bot$ is a special symbol denoting the
truth value \emph{false}, and $\naf$ is a connective called {\em
negation-as-failure}. A negation as failure literal (or
naf-literal) is of the form $\naf a$ where $a$ is an atom. For a
rule of the form (\ref{lprule1})-(\ref{lprule2}), the left and
right hand side of the rule are called the \emph{head} and the
\emph{body}, respectively. A rule of the form (\ref{lprule2}) is
also called a constraint.

\st Given a logic program $\Pi$, we will assume that each rule in
$\Pi$ is replaced by the set of its ground instances so that all
atoms in $\Pi$ are ground. Consider a set of ground atoms $X$. The
body of a rule of the form (\ref{lprule1}) or (\ref{lprule2}) is
satisfied by $X$ if $\{a_{m+1},\ldots,a_n\} \cap X = \emptyset$
and $\{a_{1},\ldots,a_m\} \subseteq X$. A rule of the form
(\ref{lprule1}) is satisfied by $X$ if either its body is not
satisfied by $X$ or $a_0 \in X$. A rule of the form
(\ref{lprule2}) is satisfied by $X$ if its body is not satisfied
by $X$. An atom $a$ is supported by $X$ if $a$ is the head of some
rule of the form (\ref{lprule1}) whose body is satisfied by $X$.

\st For a set of ground atoms $S$ and a program $\Pi$, the reduct
of $\Pi$ with respect to $S$, denoted by $\Pi^S$, is the program
obtained from the set of all ground instances of $\Pi$ by deleting

\begin{enumerate}
\item each rule that has a naf-literal $\naf a$ in its body with
$a \in S$, and

\item all naf-literals in the bodies of the remaining clauses.
\end{enumerate}

\ni $S$ is an \emph{answer set (or a stable model)} of $\Pi$ if it
satisfies the following conditions.

\begin{enumerate}
\item  If $\Pi$ does not contain any naf-literal
(i.e. $m=n$ in every rule of $\Pi$) then $S$ is the smallest set
of atoms that satisfies all the rules in $\Pi$.

\item If the program $\Pi$ does contain some naf-literal
($m < n$ in some rule of $\Pi$), then $S$ is an answer set of
$\Pi$ if $S$ is the answer set of $\Pi^S$. (Note that $\Pi^S$ does
not contain naf-literals, its answer set is defined in the first
item.)
\end{enumerate}

\ni A program $\Pi$ is said to be \emph{consistent} if it has an
answer set. Otherwise, it is inconsistent.

\st Many robust and efficient systems that can compute answer sets
of propositional logic programs have been developed. Two of the
frequently used systems are {\bf dlv} \cite{eiter98a} and {\bf
smodels} \cite{sim02}. Recently, two new systems
{\bf cmodels} \cite{cmodel} and {\bf ASSAT} \cite{lin02a},
which compute answer sets by using SAT solvers, have been developed.
{\bf XSB} \cite{sagonas94}, a
system developed for computing the well-founded model of logic
programs, has been extended to compute stable models of logic
programs as well.

\subsubsection{Answer set programming: problem solving using AnsProlog}

Prolog and other early logic programming systems were geared
towards answering yes/no queries with respect to a program, and if
the queries had variables they returned instantiations of the variables
together
with a `yes' answer. The possibility of multiple answer sets and
no answer sets has given rise to an alternative way to solve
problems using AnsProlog. In this approach, referred to as answer
set programming (also known as stable model programming)
\cite{mar99,nie99,lif99}, possible solutions of a problem are
enumerated as answer set candidates and non-solutions are
eliminated through rules with $\bot$ in the head, resulting in a
program whose answer sets have one-to-one correspondence with the
solutions of the problem.

\st We illustrate the concepts of answer set programming by
showing how the 3-coloring problem of a bi-directed graph $G$ can
be solved using AnsProlog.  Let the three colors be red ($r$),
blue ($b$), and green ($g$) and the vertex of $G$ be
$0,1,\ldots,n$. Let $P(G)$ be the program consisting of

\begin{itemize}
\item the set of atoms $edge(u,v)$ for every edge $(u,v)$ of $G$,

\item for each vertex  $u$ of $G$, three rules stating that $u$ must
be assigned one of the colors red, blue, or green:
\begin{eqnarray*}
color(u, g) \la \naf color(u, b), \naf color(u,r) \\
color(u, r) \la \naf color(u, b), \naf color(u,g) \\
color(u, b) \la \naf color(u, r), \naf color(u,g)
\end{eqnarray*}
and

\item for each edge $(u,v)$ of $G$, three rules representing the
constraint that $u$ and $v$ must have different color:
\begin{eqnarray*}
\bot \la color(u, r), color(v, r), edge(u,v) \\
\bot \la color(u, b), color(v, b), edge(u,v) \\
\bot \la color(u, g), color(v, g), edge(u,v)
\end{eqnarray*}
\end{itemize}
It can be shown that for each graph $G$, (i) $P(G)$ is
inconsistent iff the 3-coloring problem of $G$ does not have a
solution; and (ii) if $P(G)$ is consistent then each answer set of
$P(G)$ corresponds to a solution of the 3-coloring problem of $G$
and vice versa.

\st To make answer set style programming easier, Niemel{\"{a}} et
al. \cite{nie99b} introduce a new type of rules, called {\em
cardinality constraint rule} (a special form of the {\em weight
constraint rule}) of the following form:
\begin{equation}
\label{weight}
  l \{b_1,\ldots,b_k\} u \leftarrow a_1,\ldots,a_m,\naf a_{m+1},\ldots,\naf a_n
\end{equation}
where $a_i$ and $b_j$ are atoms and $l$ and $u$ are two integers,
$l \le u$. The intuitive meaning of this rule is that whenever its
body is satisfied then at least $l$ and at most $u$ atoms of the
set $\{b_1,\ldots,b_k\}$ must be true. Using rules of this type,
one can greatly reduce the number of rules of programs in answer
set programming. For instance, in the above example, the three
rules representing the  constraint that every node $u$ needs to be
assigned one of the three colors can be packed into one
cardinality constraint rule:
$$ 1 \{color(u,g), color(u,r), color(u,b) \} 1 \leftarrow $$
The semantics of logic programs with such rules is given in
\cite{nie99b} where a program with weight constraint rules is
translated into a normal logic program whose answer sets define
the answer sets of the original program. For our purpose in this
paper we only need to consider rules with $l\le1$, $u=1$, and
restrict that if we have rules of the form (\ref{weight}) in our
program then there are no other rules with any of $b_1, \ldots,
b_k$ in their head.

\section{Answer Set Planning: Using AnsProlog for planning}
\label{asp-sec}

The idea of using logic programming with answer set semantics for
planning was first introduced in \cite{sub95}. It has become more
feasible since the development of fast and efficient answer set
solvers such as {\bf smodels} \cite{sim02} and  {\bf dlv}
\cite{eiter98a}. The term ``{\em answer set planning}'' was coined
by Lifschitz in \cite{lif99} referring to approaches to
planning using logic programming with answer set semantics, where
the planning problem is expressed as a logic program and the
answer sets encode plans. In that paper answer set planning is
illustrated with respect to some specific examples.

\st We now present the main ideas of answer set
planning\footnote{Note that while \cite{lif99,lif99f,lif02a}
illustrated answer set planning through specific examples, the
papers \cite{lif99e,lif99f} mapped reasoning (not planning) in
the action description language ${\cal C}$ to logic programming.} when
the effects of actions on the world and the relationships between
fluents in the world are expressed in the action description
language $\calb$. Given a planning problem $\lan D,\Gamma, \Delta
\ran$, answer set planning solves it by translating it into a
logic program $\Pi(D, \Gamma, \Delta)$ (or $\Pi$, for short)
consisting of {\em domain-dependent} rules that describe $D$,
$\Gamma$, and $\Delta$ and {\em domain-independent} rules that
generate action occurrences and represent the inertial law.
We assume that actions and fluents in {\bf A} and {\bf F} are
specified by facts of the form $action(.)$ and $fluent(.)$,
respectively. The rules of $\Pi$ are adapted mainly from
\cite{dimo97,lif99,lif99e,lif99f} and based on conversations with
Michael Gelfond in 1998-99. As customary in the encoding of
planning problems, we assume that the length of plans we are
looking for is given. We denote the length by the constant
$length$ and use a sort \emph{time}, whose domain is the set of
integers from 0 to $length$, to represent the time moment the
system is in.  We begin with the set of domain-dependent rules.

\subsection{Domain-dependent rules} \label{subact}

For an action theory $(D,\Gamma)$, the encoding of $(D,\Gamma)$
uses the following predicates:

\begin{enumerate}
\item $holds(F,T)$ -- the fluent literal $F$ holds at the time
moment $T$; \item $occ(A,T)$ -- the action $A$ occurs at the time
moment $T$; and \item $possible(A,T)$ -- the action $A$ is
executable at the time moment $T$.
\end{enumerate}

\ni
The translation is as follows\footnote{
  A {\bf Sicstus}-program that translates $\calb$ planning
  problems into their corresponding {\bf smodels} encodings
  is available at
  \url{http://www.cs.nmsu.edu/~tson/ASPlan/Knowledge}.
(An earlier version of this translator was posted to the TAG
discussion web site
\url{http://www.cs.utexas.edu/users/vl/tag/discussions.html}).
}.

\begin{enumerate}

\item For each proposition $$\initially(l)$$ in $\Gamma$, the fact
\begin{equation} \label{ir_init}
holds(l, 0)
\end{equation}
belongs to $\Pi$. This says that at the time moment $0$, the fluent
literal $l$ holds.

\item For each executability condition
$$\executable(a,\{p_1,\dots,p_n\})$$ in $D$,
$\Pi$ contains the rule:
\begin{equation} \label{ir_pos}
possible(a, T) \leftarrow time(T), holds(p_1,T),\ldots,holds(p_n,T).
\end{equation}
This rules states that it is possible to execute the action $a$ at
the time moment $T$ if $\{p_1,\dots,p_n\}$ holds at $T$.

\item For each dynamic causal law
$$\causes(a,f,\{p_1,\dots,p_n\})$$ in $D$, $\Pi$ contains the
rule:
\begin{equation} \label{ir_dyn}
\begin{array}{lll}
holds(f, T+1) & \leftarrow & time(T), occ(a, T), \\
& & possible(a,T), holds(p_1,T),\ldots,holds(p_n,T).
\end{array}
\end{equation}
This rule says that if
$a$ occurs at the time moment $T$ and $a$ is executable at $T$
then the fluent literal $f$ becomes true at $T+1$ if there
exists a dynamic law
$$\causes(a,f,\{p_1,\dots,p_n\})$$ in $D$ and the $p_i$'s
hold at $T$.

\item For each static causal law $$\caused(\{p_1,\dots,p_n\}, f)$$
in $D$, $\Pi$ contains the following rule:
\begin{equation} \label{ir_sta}
holds(f, T) \leftarrow time(T), holds(p_1,T),\ldots,holds(p_n,T).
\end{equation}
This rule is a straightforward translation of the static causal
law into logic programming rule.

\end{enumerate}

\ni
We demonstrate the above translation
by encoding the blocks world domain from Example \ref{ex1}.

\begin{example}
\label{ex4}
{\rm
The rules encoding the fluents and actions of
the suitcase domain in Example \ref{ex1} are:
\[
\begin{array}{llllll}
 action(open(l_1)) & \leftarrow \hspace*{0.5in}&
    fluent(up(l_1))    & \leftarrow \\
 action(open(l_2)) & \leftarrow &
    fluent(up(l_2))    & \leftarrow \\
 action(close(l_1))   & \leftarrow &
    fluent(locked(s)) & \leftarrow \\
  action(close(l_2))   & \leftarrow &
    fluent(holding(k_1)))  & \leftarrow \\
  & &
    fluent(holding(k_2)))  & \leftarrow \\
\end{array}
\]
The first group of rules (left column) define the set {\bf A} and
the second group of rules (right column) define {\bf F}.
The dynamic law $\causes(open(l_1), up(l_1), \{\})$
is translated into the rule:
\[
holds(up(l_1), T+1) \leftarrow time(T), occ(open(l_1), T).
\]
The dynamic law $\causes(close(l_1), \neg up(l_1), \{\})$
is translated into the rule:
\[
holds(\neg up(l_1), T+1) \leftarrow time(T), occ(close(l_1), T).
\]
The executability condition $\executable(open(l_1), \{holding(k_1)\})$
is translated into the rule:
\[
possible(open(l_1), T) \leftarrow time(T), holds(holding(k_1), T).
\]
The static causal law $\caused(\{up(l_1), up(l_2)\}, \neg locked(s))$
is encoded by the rule:
\[
holds(\neg locked(s), T) \leftarrow time(T),
                holds(up(l_1), T), holds(up(l_2), T).
\]
The static causal law $\caused(\{\neg up(l_1)\}, locked(s))$
is encoded by the rule:
\[
holds(locked(s), T) \leftarrow time(T),
                holds(\neg up(l_1), T).
\]
The encoding of other propositions of the domain is similar.
\qed
}
\end{example}

\subsection{Domain independent rules} \label{subind}

The set of domain independent rules of $\Pi$ consists of rules
for generating action occurrences and rules for defining auxiliary
predicates. First, we present the rules for
the generation of action occurrences.
\begin{eqnarray}
occ(A,T) & \la &  action(A), time(T), possible(A,T),
\naf nocc(A,T). \label{ir_occ}\\
nocc(A,T) & \la & action(A), action(B), time(T), A \neq B,
occ(B,T). \label{ir_nocc}
\end{eqnarray}

\ni In the above rules, $A$ and $B$ are variables representing
actions. These rules generate action occurrences, one at a time
\footnote{
 These two rules can be replaced by the {\bf smodels} cardinality
 constraint rule
 $$0 \{occ(A,T) : action(A) \} 1 \la time(T)$$
 and a set of constraints that requires that actions
 can occur only when they are executable and when some actions are
executable
 then one must occur. In many of our experiments, programs with these
 rules yield better performance.
}.
The rules of inertia (or the frame axioms) and rules defining literals
are encoded using the following
rules:
\begin{eqnarray}
literal(F) & \la & fluent(F). \label{fol_lit1}\\
literal(\neg F) & \la & fluent(F). \label{fol_lit2} \\
contrary(F, \neg F) & \la & fluent(F). \label{ir_contra1}\\
contrary(\neg F, F) & \la & fluent(F). \label{ir_contra2}\\
holds(L, T{+}1)& \la & literal(L), literal(G), time(T),
\label{ir_inert}
\\
 & & contrary(L, G),  holds(L, T), \naf
holds(G, T{+}1). \nonumber
\end{eqnarray}
The first two rules define what is considered to be a literal. The
next two rules say that $\neg F$ and $F$ are contrary literals.
The last rule says that if $L$ holds at $T$ and its contrary does
not hold at $T+1$, then $L$ continues to hold at $T+1$. Finally,
to represent the fact that $\neg F$ and $F$ cannot be true at the
same time, the following constraint is added to $\Pi$.
\begin{eqnarray}
\bot  & \la & fluent(F),
holds(F, T), holds(\neg F, T).  \label{constraint}
\end{eqnarray}
\subsection{Goal representation} \label{subgoal}

The goal $\Delta$ is encoded by two sets of rules. The first
set of rules defines $\Delta$ as a formula over fluent
literals and the second set of rules evaluates the truth value
of $\Delta$ at different time moments. In a later section, we
show how fluent formulas can be represented and evaluated.
In this section, we will assume that $\Delta$
is simply a conjunction of literals, i.e.,
\[
\Delta = p_1 \wedge \ldots \wedge p_k
\]
where $p_i$ are literals. Then, $\Delta$ is represented
by the following rules:
\begin{equation} \label{goal}
goal  \la  holds(p_1, n), \ldots, holds(p_k, n).
\end{equation}
(Recall that the constant $n$ denotes the maximal length of
trajectories that we are looking for.)

\subsection{Correctness of $\Pi$}
\label{correctness-pi}

Let $\Pi_n(D,\Gamma,\Delta)$ be the logic
program consisting of
\begin{itemize}
\item the set of rules encoding $D$ and $\Gamma$
(rules (\ref{ir_init})-(\ref{ir_sta}))
in which the domain of $T$ is $\{0,\ldots,n\}$
and the rules define actions and fluents of $(D,\Gamma)$,
\item the set of domain-independent rules
(rules (\ref{ir_occ})-(\ref{constraint}))
in which the domain of $T$ is $\{0,\ldots,n\}$,
\item the rule in (\ref{goal}) and the constraint
$\bot \la \naf goal(n)$ that
encodes the requirement that $\Delta$ holds at $n$.
\end{itemize}
In what follows, we will write $\Pi_n$ instead of
$\Pi_n(D,\Gamma,\Delta)$ when it is clear from the context  what
$D$, $\Gamma$, and $\Delta$ are. The following result (similar to
the correspondence between histories and answer sets in
\cite{lif99e}) shows the equivalence between trajectories
achieving $\Delta$ and answer sets of $\Pi_n$. Before stating the
theorem, we introduce the following notation: for an answer set
$M$ of $\Pi_n$, we define
\[
s_i(M) = \{f \  \mid f \textnormal{ is a fluent literal and }
holds(f,i) \in M\}.
\]
\begin{theorem}
\label{th1}
{\rm
For a planning problem  $\lan D,\Gamma,\Delta \ran$ with a consistent
action theory $(D,\Gamma)$,
\begin{itemize}
\item[(i)]
if $s_0a_0\ldots a_{n-1}s_n$ is a trajectory achieving $\Delta$, then
there exists an answer set $M$ of $\Pi_n$ such that
\begin{enumerate}
\item $occ(a_i,i) \in M$ for  $i \in \{0,\ldots,n-1\}$ and
\item
$s_i = s_i(M)$ for $i \in \{0,\ldots,n\}$.
\end{enumerate}
and
\item[(ii)]
if $M$ is an answer set of $\Pi_n$, then there exists an integer
$0 \le k \le n$ such that $s_0(M)a_{0}\ldots a_{k-1}s_k(M)$ is a
trajectory achieving $\Delta$ where $occ(a_{i},i) \in M$ for $0
\le i < k$ and if $k < n$ then no action is executable in
the state $s_k(M)$.
\end{itemize}
}
\end{theorem}
\proof See Appendix A.1 \qed

\st Note that the second item of the theorem
implies that the trajectory achieving $\Delta$ corresponds to an
answer set $M$ of $\Pi_n$ that could be shorter than the predefined
length $n$. This happens when the goal is reached with a shorter
sequence of actions and no action is executable in the resulting
state.

\st Recall that the sequence of actions $a_0,a_1,\ldots,a_{n-1}$,
where $s_0a_0s_1 \ldots a_{n-1}s_n$ is a trajectory achieving $\Delta$,
is {\em not necessarily a plan achieving the goal} $\Delta$ because
the action theory $(D,\Gamma)$ may be non-deterministic. It is easy
to see that whenever $(D,\Gamma)$ is deterministic, if
$s_0a_0s_1 \ldots a_{n-1}s_n$ is a trajectory achieving $\Delta$
then $a_0,a_1,\ldots,a_{n-1}$ is indeed a plan achieving $\Delta$.
The next corollary follows directly from Theorem \ref{th1}.

\begin{corollary}
\label{cth1}
{\rm
For a planning problem  $\lan D,\Gamma,\Delta \ran$ with a consistent
and deterministic action theory $(D,\Gamma)$,
\begin{enumerate}
\item for each plan $a_0,\ldots,a_{n-1}$ achieving $\Delta$
from $\Gamma$, there exists an answer set $M$ of $\Pi_n$ such
that $occ(a_i,i) \in M$ for  $i \in \{0,\ldots,n-1\}$; and
\item
for each  answer set $M$ of $\Pi_n$, there exists an integer $0
\le k \le n$ such that $a_{0}, \ldots, a_{k-1}$ is a plan
achieving $\Delta$ from $\Gamma$ where $occ(a_{i},i) \in M$ for $0
\le i < k$ and if $k < n$ then no action is executable in
the state reached after executing $a_{0}, \ldots, a_{k-1}$ in the
initial state.
\end{enumerate}
}
\end{corollary}
The first item of the corollary follows from Item (i) of Theorem
\ref{th1}. Nevertheless, we do not need to include the condition on the
states $s_i(M)$ because $s_i(M)$ is uniquely determined by the
initial state $s_0$ and the sequence of actions
$a_0,\ldots,a_{i-1}$. The second item of the corollary follows
from Item (ii) of Theorem \ref{th1}. Again, because of the
determinism of $(D,\Gamma)$, we do not need to include the
conditions on the states $s_i(M)$.

\section{Control Knowledge as Constraints}

In this section, we present the main contribution of this paper:
augmenting the answer set planning program $\Pi$, introduced in
the previous section, with different kinds of domain knowledge,
namely temporal knowledge, procedural knowledge, and HTN-based
knowledge. The domain knowledge acts as constraints on the answer
sets of $\Pi$. For each kind of domain knowledge (also referred to
as constraints) we introduce new constructs for its encoding and
present a set of rules that check when a constraint is satisfied.
We now proceed to introduce the different types of control
knowledge. We start with temporal knowledge.

\subsection{Temporal Knowledge}
\label{s31}

Use of temporal domain knowledge in planning was first proposed by
Bacchus and Kabanza in \cite{bacchus00}. In their formulation,
temporal knowledge is used to prune the search space while
planning using forward search. In their paper, temporal
constraints are specified using a future linear temporal logic
with a precisely defined semantics. Since their representation is
separate from the action and goal representation, it is easy to
add them to (or remove them from) a planning problem. Planners
exploiting temporal knowledge to control the search process have
proven to be highly efficient and to scale up well \cite{aips00}.
In this paper, we represent temporal knowledge using temporal
constraints. Temporal constraints are built from fluent formulae
using the {\em temporal operators} {\bf always}, {\bf until}, {\bf
next}, and {\bf eventually}, and a special {\em goal operator}
{\bf goal}. For simplicity of the presentation, we will write
fluent formulae in prefix notation and use the {\em propositional
connectives} {\bf and, or}, and {\bf negation}. Given a signature
$\langle \mathbf{O}, \mathbf{AN}, \mathbf{BN} \rangle$ we define
term, formula, and closed formula, as follows.

\begin{definition}
{\rm
A {\em term} is a variable or a constant in {\bf O}.
}
\end{definition}

\begin{definition}
{\rm
A formula is either

\begin{itemize}

\item an expression of the form $f(\sigma_1,\ldots,\sigma_n)$ where
$f$ is a n-ary fluent name and each $\sigma_i$ is a term;

\item an expression of the form $\mathbf{and}(\phi,\psi)$,  where
$\phi$ and $\psi$ are formulae;

\item an expression of the form $\mathbf{or}(\phi,\psi)$,  where
$\phi$ and $\psi$ are formulae;

\item an expression of the form $\mathbf{negation}(\phi)$, where
$\phi$ is a formula; or

\item an expression of the form $(\exists X.\{c_1,\ldots,c_n\} \phi)$
or $(\forall X.\{c_1,\ldots,c_n\}  \phi)$ where $X$ is a variable,
$\{c_1,\ldots,c_n\}$ is a set of constants in
$\mathbf{O}$, and $\phi$ is a formula.
\end{itemize}
}
\end{definition}

\ni
We next define the notion of a {\em closed formula}.

\begin{definition}
{\rm The formula over which a quantifier applies is called the
{\em scope} of the quantifier. The scope of $\forall
X.\{c_1,\ldots,c_n\}$ (resp. $\exists X.\{c_1,\ldots,c_n\}$) in
$(\forall X.\{c_1,\ldots,c_n\} \phi)$ (resp. $(\exists
X.\{c_1,\ldots,c_n\} \phi)$) is $\phi$. An occurrence of a
variable in a formula is a bounded occurrence iff the occurrence
is within the scope of a quantifier which has the same variable
immediately after the quantifier or is the occurrence of that
quantifier. An occurrence of a variable in a formula is a free
occurrence iff the occurrence is not bound. A variable is free in
a formula if at least one of its occurrences is a free occurrence.
}
\end{definition}

\begin{definition}
\label{clsformula}
{\rm
A formula without free variables is called a {\em closed formula}.
}
\end{definition}

\begin{remark}
{\rm The truth or falsity of a formula is evaluated with
respect to state in the
standard way. It is easy to see that formulae with quantifiers can
be translated into equivalent formulae without quantifier as
follows: $\forall X.\{c_1,\ldots,c_n\} \phi$ is equivalent to
$\bigwedge_{i=1}^n \phi(c_i)$ and $\exists X.\{c_1,\ldots,c_n\}
\phi$ is equivalent to $\bigvee_{i=1}^n \phi(c_i)$ where
$\phi(c_i)$ is the formula obtained from $\phi$ by replacing every
free occurrence of $X$ in $\phi$ with $c_i$. For this reason, we
will be dealing with formulae without quantifiers hereafter. }
\end{remark}

\ni
We are now ready to define the notion of a temporal constraint.

\begin{definition}
\label{d-formula}
{\rm
A {\em temporal constraint} is either

\begin{itemize}

\item a closed formula (Definition \ref{clsformula})

\item an expression of the form $\mathbf{goal}(\phi)$
where $\phi$ is a closed formula; or

\item an expression of the form
$\aand(\phi,\psi)$, $\oor(\phi,\psi)$, $\negation(\phi)$,
$\until(\phi,\psi)$, $\always(\phi)$,
$\eventually(\phi)$, or $\next(\phi)$
where $\phi$ and $\psi$ are temporal constraints.
\end{itemize}
}
\end{definition}

\ni A temporal constraint is an {\em atomic} constraint if
it is a fluent literal. Otherwise, it is called {\em non-atomic}.
In what follows, a constraint
$\phi$ will be referred as a {\em sub-constraint} of a
constraint $\psi$ if $\phi$ occurs in $\psi$. We will
write $sub(\phi)$ to denote the set of constraints consisting of
$\phi$ and its sub-formulae. It is easy to see that
constraints without temporal operators or the goal operator are
indeed {\em fluent formulae}. Temporal operators are understood
with their standard meaning while the goal operator
$\mathbf{goal}$ provides a convenient way
for expressing the control knowledge which depends on goal
information. A temporal constraint is said to be goal-independent if
no goal formula occurs in it. Otherwise, it is goal-dependent.
Bacchus and Kabanza \cite{bacchus00} observed that useful
temporal knowledge in planning is often goal-dependent. In the
blocks world domain, the following goal-dependent constraint\footnote{
Because material implication (denoted by $\supset$) can be replaced by
$\vee$ and $\neg$, we omit it in the
definition but use it in writing the constraints, to simplify reading.
As before, we use the convention that a formula with variables
represents the set of its ground instantiations.
}:
\beq
\always(\aand(\mathbf{goal}(on(X,tbl)),on(X,tbl)) \supset
     \next(on(X,tbl)))
\eeq{temp-exp}
can be used to express that if the goal is to have a block
on the table and it is already on the table then it should be still on
the table in the next moment of time. This has the effect of
preventing the agent from superfluously picking up a block from
the table if it is supposed to be on the table in the goal state.

\st Notice that under this definition, temporal
operators can be nested many times but the goal operator
$\mathbf{goal}$ cannot be nested. For instance, if $\varphi$ is a
fluent formula, $\always(\next(\varphi))$ is a temporal formula,
but $\mathbf{goal}(\mathbf{goal}(\varphi))$ is not.

\st Goal-independent formulae will be interpreted over an infinite
sequence of states of $D$, denoted by $I = \langle s_0,s_1,\ldots,
\rangle$. On the other hand, goal-dependent formulae will be
evaluated with respect to a pair $\langle I,\varphi \rangle$ where
$I$ is a sequence of states and $\varphi$ is a fluent formula. In
the next two definitions, we formally define when a constraint is
satisfied. Definition \ref{tpl-semantics} deals with
goal-independent constraints while Definition \ref{tpl-semantics1}
is concerned with general constraints.

\begin{definition}  \label{goal-ind-def}
\label{tpl-semantics} {\rm (See \cite{bacchus00}) Let $I = \langle
s_0, s_1, \ldots, s_n, \ldots \rangle$ be a sequence of states of
$D$. Let $f_1$ and $f_2$ be goal-independent temporal constraints,
$t$ be a non-negative integer, and $f_3$ be a fluent formula. Let
$I_t = \langle s_t, s_{t+1}, \ldots, \rangle$ denote the
subsequence of $I$ starting from $s_t$. We say that $I$ satisfies
$f$ ($f$ is either $f_1, \; f_2$, or $f_3$), denoted by $I \models
f$, iff $I_0 \models f$ where

\begin{itemize}
\item  $I_t \models f_3$ iff $s_t \models f_3$.
\item $I_t\models \until(f_1,f_2)$ iff there exists $t \leq t_2 $ such
that $I_{t_2} \models f_2$ and for all $t \leq t_1 < t_2$ we have
$I_{t_1} \models f_1$.
\item $I_t\models \next(f_1)$ iff $I_{t+1} \models f_1$.
\item $I_t\models \eventually(f_1)$ iff there exists $t \leq t_1$ such
that $I_{t_1}\models f_1$.
\item $I_t \models \always(f_1)$ iff for all $t \leq t_1$ we have
$I_{t_1} \models f_1$.
\end{itemize}

\ni For a finite sequence of states $I = \lan s_0,\ldots,s_n \ran
$ and a goal-independent temporal constraint $f$,  we say that $I$
satisfies $f$, denoted by $I \models f$, if $I' \models f$ where
$I' = \lan s_0,\ldots,s_n,s_n, \ldots \ran $. \hfill$\Box$ }
\end{definition}

\ni Next we define when goal-dependent temporal constraints are
satisfied by a sequence of states and a goal. Intuitively, this
should be a straightforward extension of the previous definition
in which formulas of the form $\mathbf{goal}(\varphi)$ need to be
accounted for.   Obviously, such a constraint can only be
evaluated with respect to a sequence of states and a formula
encoding the goal. Furthermore, the intuition behind the formula
$\mathbf{goal}(\psi)$ is that $\psi$ is true whenever the goal is
true, i.e., $\psi$ is entailed by the goal. This is detailed in
the second item of the following definition.

\begin{definition}
\label{tpl-semantics1} {\rm Let $I = \langle s_0, s_1, \ldots,
s_n, \ldots \rangle$ be a sequence of states of $D$ and $\varphi$
be a fluent formula denoting the goal. Let $f_1$ and $f_2$ be
temporal constraints (possibly goal dependent), $t$ be a
non-negative integer, and $f_3$ be a fluent formula. Let $I_t =
\langle s_t, s_{t+1}, \ldots, \rangle$. We say that $I$ satisfies
$f$ ($f$ is either $f_1, \; f_2$, or $f_3$) with respect to
$\varphi$, denoted by $\langle I,\varphi \rangle \models f$, iff
$\langle I_0, \varphi \rangle \models f$ where
\begin{itemize}
\item  $\langle I_t, \varphi
\rangle \models f_3$ iff $s_t \models f_3$.
\item $\langle I_t, \varphi \rangle \models \mathbf{goal}(f_3)$
iff $\varphi \models f_3$ \footnote{
    Here, by $\varphi \models f_3$ we mean
    that $\varphi$ entails $f_3$.
}.
\item $\langle I_t, \varphi \rangle \models \until(f_1,f_2)$
iff there exists $t \leq t_2 $ such that $\langle I_{t_2}, \varphi
\rangle \models f_2$ and for all $t \leq t_1 < t_2$ we have
$\langle I_{t_1}, \varphi \rangle \models f_1$.
\item $\langle I_t, \varphi \rangle \models \next(f_1)$ iff
$\langle I_{t+1}, \varphi \rangle \models f_1$.
\item $\langle I_t, \varphi \rangle \models \eventually(f_1)$
iff there exists $t \leq t_1$ such
that $\langle I_{t_1}, \varphi \rangle \models f_1$.
\item $\langle I_t, \varphi \rangle  \models \always(f_1)$ iff for all
$t \leq t_1$ we have $\langle I_{t_1}, \varphi \rangle \models f_1$.
\end{itemize}

\ni For a finite sequence of states $I = \lan s_0,\ldots,s_n \ran
$, a temporal constraint $f$, and a fluent formula $\varphi$ we
say that $I$ satisfies $f$ with respect to $\varphi$, denoted by
$\langle I,\varphi\rangle \models f$, if $\langle
I',\varphi\rangle \models f$ where $I' = \lan s_0,\ldots,s_n,s_n,
\ldots \ran $. \hfill$\Box$ }
\end{definition}

\ni
To complete the encoding of temporal constraints, we now
provide the rules that check the satisfiability of
a temporal constraint given a trajectory.
We define the predicate $hf(F,T)$ whose truth value
determines whether $F$ is satisfied by $\langle s_T, s_{T+1}, \ldots
s_n \rangle$, where $s_T$ refers to the state corresponding to time
point $T$.
It is easy to see that rules for checking the satisfiability of
temporal constraints can be straightforwardly developed in logic
programming with function symbols. For example,
the rules
\[
\begin{array}{rcl}
hf(L, T) & \leftarrow & holds(L, T), literal(L) \\
hf(and(F_1,F_2), T) & \leftarrow & hf(F_1,T), hf(F_2, T) \\
\end{array}
\]
can be used to determine whether or not the constraint $\aand(F_1,F_2)$
is true at the time moment $T$. The first rule is for atomic
constraints and the second rule is for non-atomic ones.
Although these rules are intuitive and correct, we will
need to modify them for use with the currently available
answer set solvers  such as {\bf dlv} and {\bf smodels}.
This is because {\bf dlv}
does not allow function symbols and {\bf lparse} -- the
parser of the {\bf smodels} system -- requires that variables
occurring in the head of a rule are domain variables, i.e.,
in the second rule, we have to specify the domain of
$F_1$ and $F_2$.

\st
We will now present two possible ways to deal with the
answer set solver's restriction\footnote{
   Another alternative for dealing with temporal constraints
   such as fluent formulae
   is to convert them into disjunctive normal
   form and to develop, for each conjunction
   $\aand(f_1,\aand(f_2,\ldots,\aand(f_{n-1},f_n)))$,
   a rule
   $$hf(\aand(f_1,\aand(f_2,\ldots,\aand(f_{n-1},f_n))),T)
   \la holds(f_1,T), \ldots, holds(f_n,T).$$
   This method, however, cannot be easily extended for
   temporal constraints
   with temporal operators or the goal operator.
}.
The first way is to represent a constraint by a set of rules
that determine its truth value. In other words, we specify
the domains of $F_1$ and $F_2$ in the above rules by grounding
them. For example, for the conjunction $\aand(f,g)$,
the rules
\[
\begin{array}{rcl}
hf(L, T) & \leftarrow & literal(L), holds(L, T) \\
hf(and(f,g), T) & \leftarrow & hf(f,T), hf(g, T)
\end{array}
\]
can be used.
For the disjunction, $\oor(f, \aand(g,h))$, the rules
\[
\begin{array}{rcl}
hf(L, T) & \leftarrow & literal(L), holds(L, T) \\
hf(or(f,and(g,h)), T) & \leftarrow & hf(f,T). \\
hf(or(f,and(g,h)), T) & \leftarrow & hf(and(g,h),T). \\
hf(and(g,h), T) & \leftarrow & hf(g,T), hf(h,T). \\
\end{array}
\]
can be used.
The encodings of other constraints are similar.
Observe that the number of rules for encoding a formula depends
on the number of its sub-constraints.

\st
An alternative to the above
encoding is to assign names to non-atomic constraints, to
define a new type, called {\em formula}, and to provide
the constraint-independent rules for checking the truth value of
constraints. Atomic constraints are defined by the rule:
\[
formula(L) \leftarrow literal(L).
\]
For each non-atomic formula $\phi$, we associate with it a unique
name $n_\phi$ and encode it by a set of facts, denoted by
$r(\phi)$. This set is defined inductively over the structure of
$\phi$ as follows.

\begin{itemize}
\item If $\phi$ is a fluent literal $l$ then $r(\phi) = \{l\}$;

\item If $\phi = \phi_1 \wedge \phi_2$
      then $r(\phi) = r({\phi_1}) \cup r({\phi_2}) \cup
    \{formula(n_\phi), and(n_\phi, n_{\phi_1},n_{\phi_2})\}$;

\item If $\phi = \phi_1 \vee \phi_2$
      then $r(\phi) = r({\phi_1}) \cup r({\phi_2}) \cup
    \{formula(n_\phi), or(n_\phi, n_{\phi_1},n_{\phi_2})\}$;

\item If $\phi = \neg \phi_1$
      then $r(\phi) = r(\phi_1) \cup
    \{formula(n_\phi), negation(n_\phi, n_{\phi_1})\}$;

\item If $\phi = \next(\phi_1)$
      then $r(\phi) = r(\phi) \cup
     \{formula(n_\phi), next(n_\phi, n_{\phi_1})\}$;

\item If $\phi = \until(\phi_1,\phi_2)$
      then $$r(\phi) = r(\phi_1) \cup r(\phi_2) \cup
     \{formula(n_\phi), until(n_\phi, n_{\phi_1},n_{\phi_2})\};$$

\item If $\phi = \always(\phi_1)$
      then $r(\phi) = r(\phi_1) \cup
     \{formula(n_\phi), always(n_\phi, n_{\phi_1})\}$;

\item If $\phi = \eventually(\phi_1)$
      then $r(\phi) = r(\phi_1) \cup
     \{formula(n_\phi), eventually(n_\phi, n_{\phi_1})\}$.
\end{itemize}
For simplicity, the names assigned to a constraint can be used in
encoding other constraints. For example, the constraints $\phi =
\aand(f, \aand(g,h))$ is encoded by the atoms
\begin{eqnarray*}
& formula(n_\psi). \\
& and(n_\psi,g,h). \\
& formula(n_\phi). \\
& and(n_\phi,f, n_\psi).
\end{eqnarray*}
\ni We note that the above encodings can be generated automatically
using a program
front-end to {\bf smodels} that is available on the web-site
containing the experimental results presented in this paper.
Note that during the grounding phase of {\bf smodels} (by
{\bf lparse}), atoms of the form $formula(.,.)$ will be removed.
For this reason, we use the second encoding in our experiments
because it is easier to deal with changes in the constraints used
for encoding the control knowledge.

We now present the
formula-independent rules for evaluating temporal constraints. As
with defining the satisfaction of temporal constraints, we first
consider goal-independent temporal constraints. The rules needed
for evaluating temporal constraints whose first level operator is
different than the {\bf goal} operator are as follows:

{\tt
\begin{eqnarray}
hf(L,T) & \la &
    literal(L),  holds(L,T). \label{fol_lit_tran}
\\
hf(N,T) & \la &
    formula(N), and(N,N_1,N_2), \label{fol_and} \\
    & &  hf(N_1,T), hf(N_2,T). \\
hf(N,T) & \la &
    formula(N), or(N,N_1,N_2), hf(N_1,T).  \label{fol_or1} \\
hf(N,T) & \la &
    formula(N), or(N,N_1,N_2), hf(N_2,T). \label{fol_or2} \\
hf(N,T) & \la &
    formula(N), negation(N,N_1)),
    \naf hf(N_1,T). \label{fol_neg} \\
hf(N, T) & \la & formula(N), until(N, N_1, N_2), \label{lt_until} \\
 & &    T \le T', hf\_during(N_1, T, T'),
      hf(N_2, T'). \nonumber
\\
hf(N, T) & \la & formula(N), always(N,N_1), \label{lt_always} \\
    & & hf\_during(N_1, T, n). \nonumber \\
hf(N, T) & \la & formula(N), eventually(N,N_1), \label{lt_event} \\
& & hf(N_1, T'),T \le T'. \nonumber \\
hf(N, T) & \la & formula(N), next(N,N_1),
hf(N_1, T+1).
\label{lt_next} \\
hf\_during(N,T,T) & \la & hf(N,T).
\label{lt_during0} \\
hf\_during(N,T,T') & \la & hf(N,T), T<T', hf\_during(N,T+1,T').
\label{lt_during}
\end{eqnarray}
}

\st The meaning of these rules is straightforward. The first rule
defines the truth value of an atomic formula (a literal). Rule
(\ref{fol_and}) says that a conjunction holds if its conjuncts
hold. Rules (\ref{fol_or1})-(\ref{fol_or2}) say that a disjunction
holds if one of its disjuncts holds. The rule (\ref{fol_neg})
states that the negation of a formula holds if its negation does
not hold. Its correctness is due to the assumption that initial
states are complete. Rules (\ref{lt_until})-(\ref{lt_during}) deal
with formulae containing temporal operators. The constant $n$
denotes the maximal length of trajectories that we are looking
for. In the following, we refer to this group of rules by
$\Pi_{formula}$.

\st
The next theorem shows that rules (\ref{fol_lit_tran})-(\ref{lt_during})
correctly implement the semantics of goal-independent temporal formulae.

\begin{theorem}\label{the-lt-sem}{
\rm Let $S$ be a finite set of goal-independent temporal formulae,
$I = \langle s_0, s_1\ldots s_n \rangle$ be a sequence of states,
and
$$\Pi_{formula}(S,I) = \Pi_{formula}
\cup r(I) \cup r(S)$$ where
\begin{itemize}
\item $r(S)$ is the set of atoms used in encoding $S$, and
\item
$r(I) = \cup_{t=0}^{n} \{holds(l, t) \mid l$ is a fluent literal
and $l \in s_t\}$.
\end{itemize}

\ni
Then,
\begin{itemize}
\item[(i)] The program $\Pi_{formula}(S,I)$ has a unique answer
set, $X$.
\item[(ii)] For every temporal formula $\phi$
such that $formula(n_\phi) \in r(S)$, $\phi$ is true in $I_t$,
i.e., $I_t \models \phi$, if and only if $hf(n_\phi,t)$ belongs to
$X$ where $I_t = \langle s_t, \ldots s_n \rangle$.
\end{itemize}
}
\end{theorem}
\ni
\proof See Appendix A.2 \qed

\st Having defined temporal constraints and specified when they
are satisfied, adding temporal knowledge to a planning problem in
answer set planning is easy. We must encode the knowledge as a
temporal formula\footnote{
  A set of temporal formulae can be viewed as a conjunction of
  temporal formulae.
}
and then add the set of rules representing this formula
and the rules  (\ref{fol_lit_tran})-(\ref{lt_during}) to $\Pi$.
Finally, we need to
add the constraint that requires that the goal is true at the
final state and the temporal formula is satisfied. More
precisely, for a planning problem
$\langle D,\Gamma,\Delta \rangle$ and a goal-independent
temporal formula $\phi$,
let $\Pi^{TLP}_n$ be the program consisting of
\begin{itemize}
\item the program $\Pi_n$ (Defined as in Sub-section
\ref{correctness-pi}),

\item the rules (\ref{fol_lit_tran})-(\ref{lt_during})

\item the rules encoding $\phi$ and the
constraint $\bot \la \naf hf(n_\phi,0)$.
\end{itemize}

\ni The next theorem is about the correctness of $\Pi^{TLP}_n$.

\begin{theorem}
\label{thtlp} {\rm For a planning problem  $\lan D,\Gamma,\Delta
\ran$ with a consistent action theory $(D,\Gamma)$ and a
goal-independent temporal formula $\phi$,
\begin{itemize}
\item[(i)]
if $s_0a_0\ldots a_{n-1}s_n$ is a trajectory achieving $\Delta$
and $I  \models \phi$ where $ I = \lan s_0,\ldots,s_n \ran$, then
there exists an answer set $M$ of $\Pi^{TLP}_n$ such that
\begin{enumerate}
\item $occ(a_i,i) \in M$ for  $i \in \{0,\ldots,n-1\}$,
\item
$s_i = s_i(M)$ for $i \in \{0,\ldots,n\}$,  and
\item $hf(n_\phi, 0) \in M$.
\end{enumerate}
and
\item[(ii)]
if $M$ is an answer set of $\Pi^{TLP}_n$, then there exists an
integer $0 \le k \le n$ such that
\begin{enumerate}
\item
$s_0(M)a_{0}\ldots a_{k-1}s_k(M)$ is a trajectory achieving
$\Delta$ where $occ(a_{i},i) \in M$ for $0 \le i < k$ and
\item
$I  \models \phi$ where $I = \lan s_0(M),\ldots,s_n(M) \ran$.
\end{enumerate}
\end{itemize}
}
\end{theorem}

\proof Observe that the set of literals of the program $\Pi_n$,
$lit(\Pi_n)$, is a splitting set of the program $\Pi_n^{TPL}$ and
$\Pi_n = b_{lit(\Pi_n)}(\Pi_n^{TPL})$. Thus, $M$ is an answer set
of $\Pi_n^{TPL}$ iff $M = X \cup Y$ where $X$ is an answer set of
$\Pi_n$ and $Y$ is an answer set of $e_{lit(\Pi_n)}(\Pi_n^{TPL}
\setminus \Pi_n, X)$ which consists of the set of rules
(\ref{fol_and})-(\ref{lt_during}), the set of atoms $\{hf(l,t)
\mid holds(l,t) \in X\}$, the rules encoding $\phi$, and the
constraint $\bot \leftarrow \naf hf(n_\phi,0)$. This constraint
implies that $hf(n_\phi,0)$ must belong to every answer set $M$ of
$\Pi_n^{TPL}$.

\st We now prove (i). It follows from Theorem \ref{th1} that there
exists an answer set $X$ of $\Pi_n$ such that the first two
conditions are satisfied. Because $I \models \phi$, we can apply
Theorem \ref{the-lt-sem} to show that any answer set $Y$ of
$e_{lit(\Pi_n)}(\Pi_n^{TPL} \setminus \Pi_n, X)$ contains
$hf(n_\phi,0)$. Thus, $X \cup Y$ is an answer set satisfying (i).

\st To prove (ii), it is enough to notice that the answer set $X$ of
$\Pi_n$, constructed in the proof of Lemma \ref{smodel-traject2-add},
can be used to construct an answer set $M$ of $\Pi_n^{TPL}$ such that
$M$ satisfies (ii).
\qed

\st The above theorem shows how control knowledge represented as
goal-independent temporal formulae can be exploited in answer set
planning. We will now extend this result to allow control
knowledge expressed using goal-dependent temporal formulae. Based
on Definition~\ref{tpl-semantics1}, where satisfaction of
goal-dependent temporal formulae is defined, we will need to
encode $\Delta \models \psi$ where $\Delta$ is the goal and
$\mathbf{goal}(\psi)$ is a formula occurring in a control
knowledge that we wish to use. To simplify this encoding we make
the same assumption that is  made in most classical planning
literature including \cite{bacchus00}: the goal $\Delta$ in a
planning problem $\langle D,\Gamma,\Delta\rangle$ is a set of
literals and each goal formula occurring in a temporal formula
representing our control knowledge is of the form
$\mathbf{goal}(F)$ where $F$ is a fluent literal. In the rest of
this section, whenever we refer to a planning problem or a
goal-dependent temporal formula we assume that they satisfy this
assumption. Let $\langle D,\Gamma,\Delta\rangle$ be a planning
problem and $\phi$ be a temporal formula. $\Pi^{TLP{+}Goal}_n$ be
the program consisting of $\Pi^{TLP}_n$, the set of atoms
$\{fomula(n_{goal_l}) \mid \mathbf{goal}(l)$ is a goal formula
occurring in $\phi$, and the set of rules
\begin{eqnarray}
hf(n_{goal_f}, T) & \la & time(T) \label{lt_e1}
\end{eqnarray}
for each $f \in \Delta$. Intuitively, these rules assert that $f$
is a part of the goal $\Delta$. The next theorem is about the
correctness of $\Pi^{TLP{+}Goal}_n$.

\begin{theorem}
\label{thtlp-goal}
{\rm
For a planning problem  $\lan D,\Gamma,\Delta \ran$ with a consistent
action theory $(D,\Gamma)$ and a temporal formula $\phi$,
\begin{itemize}
\item[(i)]
if $s_0a_0\ldots a_{n-1}s_n$ is a trajectory achieving $\Delta$
and $\langle I,\Delta\rangle \models \phi$ where $ I = \lan
s_0,\ldots,s_n \ran$, then there exists an answer set $M$ of
$\Pi^{TLP{+}Goal}_n$ such that
\begin{enumerate}
\item $occ(a_i,i) \in M$ for  $i \in \{0,\ldots,n-1\}$,
\item
$s_i = s_i(M)$ for $i \in \{0,\ldots,n\}$,  and
\item $hf(n_\phi, 0) \in M$.
\end{enumerate}
and
\item[(ii)]
if $M$ is an answer set of $\Pi^{TLP{+}Goal}_n$, then there exists an
integer $0 \le k \le n$ such that
\begin{enumerate}
\item
$s_0(M)a_{0}\ldots a_{k-1}s_k(M)$ is a trajectory achieving
$\Delta$ where $occ(a_{i},i) \in M$ for $0 \le i < k$ and
\item
$\langle I,\Delta \rangle \models \phi$ where $I = \lan
s_0(M),\ldots,s_n(M) \ran$.
\end{enumerate}
\end{itemize}
}
\end{theorem}
\proof To prove this theorem, we first need to modify Theorem
\ref{the-lt-sem} by (i) allowing goal-dependent formulae to be in
the set $S$; (ii) adding a goal $\Delta$ and the rule
(\ref{lt_e1}) to the program $\Pi$ of Theorem \ref{the-lt-sem}.
The proof of this modified theorem is very similar to the proof of
Theorem \ref{the-lt-sem}. We omit it here for brevity. This
result, together with Theorem \ref{th1}, proves the conclusion of
this theorem. \qed

\subsection{Procedural Knowledge}
\label{s32}

Procedural knowledge can be thought of as an under-specified sketch
of the plans to be generated.  The language constructs
that we use in this paper to describe procedural knowledge are inspired by
GOLOG, an Algol-like logic programming language for agent programming,
control and execution; and based on a situation calculus theory of
action \cite{rei96}. GOLOG has primarily been used as a
programming language for high-level agent control in dynamical
environments (see e.g. \cite{burgard98}).
Although a planner can itself
be written as a GOLOG program (See Chapter 10 of \cite{rei00}),
in this paper, we view a GOLOG program
as an incompletely specified plan (or as a form of procedural
knowledge) that includes non-deterministic choice points that are
filled in by the planner. For example, the procedural knowledge
(which is very similar to a GOLOG program) $a_1;a_2;(a_3 | a_4 |
a_5); f$ represents plans which have $a_1$ followed by $a_2$,
followed by one of $a_3$, $a_4$, or $a_5$ such that $f$ is true
in the following (terminating) state of the plan. A planner, when given this
procedural knowledge needs only to decide which one of $a_3$,
$a_4$, or $a_5$ it should choose as its third action.

\st
We now formally define the syntax of our procedural knowledge,
which -- keeping with the GOLOG terminology -- we refer to as a
{\em program}. A program is built from {\em
complex actions} and {\em procedures}. Complex actions,
procedures, and programs are constructed using variables,
actions, formulae, and procedural program constructs
such as {\bf sequence}, {\bf if-then-else}, {\bf while-do},
or {\bf choice}, etc.
They are defined as follows.

\begin{definition}
\label{ca}
{\rm
A {\em complex action} $\delta$ with a sequence of
variables $X_1,\ldots,X_n$
is
\begin{itemize}
\item {\em a basic complex action}:

\begin{itemize}
\item
an expression of the form $a(\sigma_1,\ldots,\sigma_m)$ where $a$
is an m-ary action name, $\sigma_i$ is either a variable or a
constant of the type $t_i$, and if $\sigma_i$ is a variable then
it belongs to $\{X_1,\ldots,X_n\}$ or

\item an expression of the form $\phi$ where $\phi$ is a formula
whose free variables are from $\{X_1,\ldots,X_n\}$;
\end{itemize}

\item {\em a sequence}: an expression of the
form $\delta_1; \delta_2$ where $\delta_1$ and
$\delta_2$ are
complex actions whose free variables are from $\{X_1,\ldots,X_n\}$;

\item {\em a choice of actions}: an expression of the form
$\delta_1 \mid \ldots \mid \delta_k$ where $\delta_j$'s are
complex actions whose free variables are from $\{X_1,\ldots,X_n\}$;

\item {\em a if-then-else}: an expression of the form
$\mathbf{if} \;  \phi  \; \mathbf{then} \; \delta_1
\; \mathbf{else} \; \delta_2$ where
$\phi$ is a formula and
$\delta_1$ and $\delta_2$ are
complex actions whose free variables are from $\{X_1,\ldots,X_n\}$;

\item {\em a while-do}: an expression of the form
$\mathbf{while} \;  \phi  \; \mathbf{do} \; \delta_1$
where
$\phi$ is a formula and  $\delta_1$ is a complex action
whose free variables are from $\{X_1,\ldots,X_n\}$;

\item {\em a choice of arguments}:  an expression of the form
$\mathbf{pick}(Y, \{c_1,\ldots,c_n\}, \delta_1)$
where ${Y \not\in \{X_1,\ldots,X_n\}}$,
$\{c_1,\ldots,c_n\}$ is a set of constants,  and
$\delta_1$ is a complex action
whose free variables are from
$\{X_1,\ldots,X_n,Y\}$; and

\item  {\em a procedure call}:
an expression of the form $p(X_1,\ldots,X_n)$ where
$p$ is a procedure name whose variables are
$X_1,\ldots,X_n$.
\end{itemize}
}
\end{definition}

\begin{definition}
\label{pr}
{\rm
A {\em procedure} with the name $p$ and a sequence of variables
$X_1,\ldots,X_n$
is of the form $(p(X_1,\ldots,X_n) : \delta)$ where
$\delta$, called the {\em body}, is a complex action
whose free variables are from $\{X_1,\ldots,X_n\}$.

\st
A procedure $(p(X_1,\ldots,X_n) : \delta)$ is called a
{\em nested procedure} if $\delta$ is a procedure call.
}
\end{definition}

\ni
Intuitively, a complex action $\delta$ represents a sketch of a plan
whose variations are given by its variables and its structure.
The execution of $\delta$ is done recursively over its structure
and starts with the instantiation of $X_1,\ldots,X_n$ with
some constants $c_1,\ldots,c_n$. In
the process, an action might be executed, a formula might be
evaluated, other complex actions or procedures
might be instantiated and executed. In other words,
the execution of $\delta$ might depend on the execution of
other complex actions. Let $\delta$ be a complex action with variables
$(X_1,\ldots,X_n)$ and $c_1,\ldots,c_n$ be constants. In the following,
we define
\begin{itemize}
\item the ground instance of $\delta$
with respect to the substitution $\{X_1/c_1,
\ldots, X_n/c_n\}$, denoted by $\delta(c_1,\ldots,c_n)$, and
\item the set of complex actions that the
execution of $\delta(c_1,\ldots,c_n)$ might depend on,
denoted by $prim(\delta( c_1,\ldots,c_n))$.
\end{itemize}
$\delta(c_1,\ldots,c_n)$ and $prim(\delta( c_1,\ldots,c_n))$
are defined recursively as follows:
\begin{itemize}
\item if $\delta = a(\sigma_1,\ldots,\sigma_m)$, then
$\delta(c_1,\ldots,c_n)$ is the action $a(c'_1,\ldots,c'_m)$ where
$c'_i=c_j$ if $\sigma_i$ is the variable $ X_j$ and
$c'_i=\sigma_i$ if $\sigma_i$ is a constant and $prim(\delta(
c_1,\ldots,c_n)) = \{a(c'_1,\ldots,c'_m)\}$,

\item if $\delta = \phi$
then $\delta(c_1,\ldots,c_n) = \phi(c_1,\ldots,c_n)$ where
$\phi(c_1,\ldots,c_n)$ is obtained from $\phi$ by simultaneously
replacing every free occurrence of $X_i$ in $\phi$ by $c_i$ and
$prim(\delta( c_1,\ldots,c_n)) = \{\phi(c_1,\ldots,c_n)\}$,

\item if ${\delta = \delta_1; \delta_2}$
then ${\delta(c_1,\ldots,c_n) = \delta_1(c_1,\ldots,c_n);
\delta_2(c_1,\ldots,c_n)}$
 and \\
${prim(\delta( c_1,\ldots,c_n)) = prim(\delta_1(c_1,\ldots,c_n))
\cup prim(\delta_2(c_1,\ldots,c_n))}$,

\item if $\delta = \delta_1 \mid \ldots\mid \delta_k$
then $\delta(c_1,\ldots,c_n) = \delta_1(c_1,\ldots,c_n) \mid
\ldots \mid
\delta_k(c_1,\ldots,c_n)$  and \\
$prim(\delta( c_1,\ldots,c_n)) = \bigcup_{i=1}^k
prim(\delta_i(c_1,\ldots,c_n)) $,

\item if ${\delta = \iif \phi \; \tthen \; \delta_1 \; \eelse \;
\delta_2}$ then \\
$\delta(c_1,\ldots,c_n) = \iif \phi(c_1,\ldots,c_n) \tthen
\delta_1(c_1,\ldots,c_n)
\; \eelse \; \delta_2(c_1,\ldots,c_n)$  and \\
$prim(\delta( c_1,\ldots,c_n)) = \{\phi(c_1,\ldots,c_n)\} \cup
prim(\delta_1(c_1,\ldots,c_n)) \cup
prim(\delta_2(c_1,\ldots,c_n))$,

\item if $\delta = \wwhile \phi \; \ddo \; \delta_1$
then $\delta(c_1,\ldots,c_n) = \wwhile \phi(c_1,\ldots,c_n) \ddo
\delta_1(c_1,\ldots,c_n)$  and \\
$prim(\delta( c_1,\ldots,c_n)) = \{\phi(c_1,\ldots,c_n)\} \cup
prim(\delta_2(c_1,\ldots,c_n))$,

\item
if ${\delta = \mathbf{pick}(Y, \{y_1,\ldots,y_m\}, \delta_1)}$
then $\delta(c_1,\ldots,c_n) = \delta_1(c_1,\ldots,c_n,y_j)$
for some $j$, $1 \le j \le m$, and \\
$prim(\delta( c_1,\ldots,c_n)) = \{
prim(\delta_1(c_1,\ldots,c_n,y_j)) \mid j=1,\ldots,m\}$; and

\item
If $\delta = p(X_1,\ldots,X_n)$ where $(p(X_1,\ldots,X_n) :
\delta_1)$ is a procedure then $\delta(c_1,\ldots,c_n) =
\delta_1(c_1,\ldots,c_n)$ and $prim(\delta( c_1,\ldots,c_n)) =
\{p(c_1,\ldots,c_n)\} \cup prim(\delta( c_1,\ldots,c_n))$.
\end{itemize}

\ni
A ground
instance of a procedure $(p(X_1,\ldots,X_n) : \delta)$ is
of the form $(p(c_1,\ldots,c_n) : \delta(c_1,\ldots,c_n))$
where $c_1,\ldots,c_n$ are constants and
$\delta(c_1,\ldots,c_n)$ is a ground instance of
$\delta$.

\st In what follows, $\delta(c_1,\ldots,c_n)$ (resp.
$(p(c_1,\ldots,c_n) : \delta(c_1,\ldots,c_n))$) will be referred
to as a ground complex action (resp. ground procedure). As with
complex actions, for a procedure $(p(X_1,\ldots,X_n) : \delta)$
and the constants $c_1,\ldots,c_n$, we define the set of actions
that the execution of $p(c_1,\ldots,c_n)$ might depend on by
$prim(p(c_1,\ldots,c_n)) = prim(\delta(c_1,\ldots,c_n))$. It is
easy to see that under the above definitions, a procedure $p$ may
depend on itself. For example, for two procedures ``$(p : \wwhile
\phi_1 \; \ddo q)$'' and ``$(q : \wwhile \phi_2 \; \ddo p)$'', we
have that $prim(p) = \{p,q,\phi_1,\phi_2\}$ and $prim(q) =
\{p,q,\phi_1,\phi_2\}$. Intuitively, this will mean that the
execution of $p$ (and $q$) might be infinite. Since our goal is to
use programs, represented as a set of procedures and a ground
complex action, to construct plans of finite length, procedures
that depend on themselves will not be helpful. For this reason, we
define a notion called {\em well-defined} procedures and limit
ourselves to this type of procedure hereafter. We say that a
procedure $p$ with variables $X_1,\ldots, X_n$ is {\em
well-defined} if there exists no sequence of constants
$c_1,\ldots,c_n$ such that $p(c_1,\ldots,c_n) \in
prim(p(c_1,\ldots,c_n))$. We will limit ourselves to sets of
procedures in which no two procedures have the same name and every
procedure is well-defined and is not a nested procedure. We call
such a set of procedures {\em coherent} and define programs as
follows.

\begin{definition}
[Program]
\label{prog1}
{\rm
A {\em program} is a pair $(S,\delta)$
where $S$ is a coherent set of procedures and
$\delta$ is a ground instantiation of a complex action.
}
\end{definition}

\ni
We illustrate the above definition with the following
example.

\begin{example}
\label{ex5}
{\rm
In this example, we introduce the elevator domain from
\cite{rei96} which we use in our initial experiments
(Section~\ref{s34}). The set of constants in this domain
consists of integers between $0$ and $k$
representing the floor numbers controlled by the elevator.
The fluents in this domain and their
intuitive meaning are as follows:

\begin{itemize}

\item $on(N)$ - the request service light of the floor $N$ is on,
indicating a service is requested at the floor $N$,

\item $opened$ - the door of the elevator is open,  and

\item $currentFloor(N)$ - the elevator is currently at the floor $N$.
\end{itemize}

\ni The actions in this domain and their intuitive meaning are as
follows:
\begin{itemize}
\item $up(N)$ - move up to floor $N$,
\item $down(N)$ - move down to floor $N$,
\item $turnoff(N)$ - turn off the indicator light of the floor $N$,
\item $open$ - open the elevator door, and
\item $close$ - close the elevator door.
\end{itemize}

\ni The domain description is as follows:
\[
D_{elevator} = \left \{
\begin{array}{ll}
& \causes(up(N), currentFloor(N), \{\}) \\
& \causes(down(N), currentFloor(N), \{\}) \\
& \causes(turnoff(N), \neg on(N), \{\}) \\
& \causes(open, opened, \{\})  \\
& \causes(close, \neg opened,\{\}) \\
& \caused(\{currentFloor(M)\}, \neg currentFloor(N))  \mbox{ for all } M
\neq  N\\
& \executable(up(N), \{currentFloor(M), \neg opened\}) \mbox{ for
all } M <
N\\
& \executable(down(N), \{currentFloor(M), \neg opened\})  \mbox{
for all } M
> N \\
& \executable(turnoff(N), \{currentFloor(N)\}) \\
& \executable(open, \{\}) \\
& \executable(close, \{\}) \\
& \executable(null, \{\}) \\
\end{array}
\right.
\]

\ni We consider arbitrary initial states where $opened$ is false,
$currentFloor(N)$ is true for a particular $N$ and a set of
$on(N)$ is true; and our goal is to have $\neg on(N)$ for all $N$.
In planning to achieve such a goal, we can use the following set
of procedural domain knowledge. Alternatively, in
the terminology of GOLOG, we can say that the following set of
procedures, together with the ground
complex action {\em control},
can be used to control the elevator, so as to satisfy
service requests -- indicated by the light being on -- at
different floors. That is, the program for controlling
the elevator is $(S,control)$ where
\[
S = \left \{
\begin{array}{rcl}
(go\_floor(N) & : &  currentFloor(N) | up(N) | down(N))  \\
(serve(N) & : &  go\_floor(N); turnoff(N) ; open; close) \\
(serve\_a\_floor & : &  \pick(N,  \{0,\ldots,k\}, (on(N); serve(N))). \\
(park  & : &  \iif currentFloor(0) \tthen open \eelse [down(0);open]) \\
(control  & : &  [\wwhile \exists N.\{0,\ldots,k\} \;
[on(N)] \ddo serve\_a\_floor]; park). \\
\end{array}
\right.
\]

\st Observe that the formula $\exists N.d(N) \; [on(N)]$, as
discussed before, is the shorthand of the disjunction
\[
\oor(on(0), \oor(on(1), \ldots, \oor(on(k-1), on(k))))
\]
where $0,\ldots,k$ are the floor constants of the domain.
\qed
}
\end{example}

\ni The operational semantics of programs specifies when a
trajectory $s_0a_0s_1 \ldots a_{n-1}s_n$, denoted by $\alpha$, is
{\em a trace of a program $(S,\delta)$}. Intuitively, if $\alpha$ is
a trace of a program $(S,\delta)$ then that means $a_0, \ldots,
a_{n-1}$ is a sequence of actions (and $\alpha$ is a corresponding
trajectory) that is consistent with the sketch provided by the
complex action $\delta$ of the program $(S,\delta)$ starting from
the initial state $s_0$. Alternatively, it can be thought of as
the program $(S,\delta)$ {\em unfolding} to the sequence of actions
$a_0, \ldots, a_{n-1}$ in state $s_0$. We now formally define the
notion of a $trace$.
\begin{definition}
[Trace] \label{deftrace1} {\rm Let $p = (S,\delta)$ be a program.
We say that a trajectory $s_0a_0s_1 \ldots a_{n-1}s_n$ is a trace
of $p$ if one of the following conditions is satisfied:
\begin{itemize}
  \item $\delta = a$ and $a$ is an action, $n=1$ and $a_0 = a$;
  \item $\delta = \phi$, $n = 0$ and $\phi$ holds in $s_0$;
  \item $\delta = \delta_1;\delta_2$, and there exists an $i$ such
that
  $s_0a_0\ldots s_i$ is a trace of $(S,\delta_1)$ and
  $s_ia_i\ldots s_n$ is a trace of $(S,\delta_2)$;
  \item ${\delta = \delta_1 \mid \ldots \mid \delta_n}$, and
$s_0a_0\ldots a_{n-1}s_n$
  is a trace of $(S,\delta_i)$ for some $i$;
  \item $\delta = \iif \phi \tthen \delta_1 \eelse \delta_2$, and
$s_0a_0\ldots a_{n-1}s_n$
  is a trace of $(S,\delta_1)$ if $\phi$ holds in $s_0$ or
$s_0a_0\ldots a_{n-1}s_n$ is a trace of
$(S,\delta_2)$ if $\neg \phi$ holds in $s_0$;
  \item ${\delta = \wwhile \phi \ddo \delta_1}$, $n =0$ and
  $\neg \phi$ holds in $s_0$, or\\
  $\phi$ holds in $s_0$ and there
  exists some $i > 0$ such that
  $s_0a_0\ldots s_i$ is a trace of
  $(S,\delta_1)$ and $s_ia_i\ldots s_n$ is a trace of $(S,\delta)$; or
  \item ${\delta = \pick(Y, \{y_1,\ldots,y_m\}, \delta_1)}$ and
  $s_0a_0s_1 \ldots a_{n-1}s_n$ is a trace of
  $(S,\delta_1(y_j))$ for some $j$, $1 \le j \le m$.

  \item ${\delta = p(c_1,\ldots,c_n)}$ where
${(p(X_1,\ldots,X_n):\delta_1)}$ is a procedure, and \\
  $s_0a_0s_1 \ldots a_{n-1}s_n$ is a trace of
  $(S,\delta_1(c_1,\ldots,c_n))$.
\end{itemize}
\qed
}
\end{definition}
\ni
The above definition allows us to determine whether a trajectory
$\alpha$ constitutes a trace of a program $(S,\delta)$. This process
is done recursively over the structure of $\delta$.
More precisely, if $\delta$ is not an action
or a formula, checking whether $\alpha$ is
a trace of $(S,\delta)$ amounts to determining whether $\alpha$
is a trace of $(S,\delta')$ for some component $\delta'$ of $\delta$.
We note that because $\delta$ is grounded,
$\delta'$ is also a ground complex action; thus, guaranteeing
that $(S,\delta')$ is a program and hence the applicability of
the definition. It is easy to see that $\delta'$ belongs to
$prim(\delta)$. Because of the coherency of $S$ and the finiteness
of the domains, this process will eventually stop.

\st We will now present the {\bf smodels} encoding for programs.
The encoding of a program $(S,\delta)$ will include the encoding
of all procedures in $S$ and the encoding of $\delta$. The
encoding of a complex action or a procedure consists of the
encoding of all of its ground instances. Similar to the encoding
of formulae, each complex action $\delta$ will be assigned a
distinguished name, denoted by $n_\delta$, whenever it is
necessary. Because procedure names are unique in a program, we
assign the name $p(c_1,\ldots,c_n)$ to the complex action
$\delta(c_1,\ldots,c_n)$ where $(p(X_1,\ldots,X_n) : \delta)$ is a
procedure and $c_1,\ldots,c_n$ is a sequence of constants. In
other words, $n_{\delta(c_1,\ldots,c_n)} = p(c_1,\ldots,c_n)$. We
note that since the body of a procedure is not a procedure call,
this will not cause any inconsistency in the naming of complex
actions. We now describe the set of rules encoding a complex
action $\delta$, denoted by $r(\delta)$, which is defined
inductively over the structure of $\delta$ as as follows:

\begin{itemize}
\item For $\delta = a$ or $\delta = \phi$, $r(\delta)$ is the action
$a$ or the rules encoding $\phi$, respectively.

\item For $\delta = \delta_1;\delta_2$,
  $r(\delta) = \{sequence(n_\delta,n_{\delta_1},n_{\delta_2})\} \cup
r(\delta_1) \cup r(\delta_2)$.
  \item For $\delta =
        \delta_1 \mid \delta_2 \ldots \mid \delta_n$,
  $r(\delta) =  \bigcup_{i=1,\ldots,n} r(\delta_i) \cup
\{in(n_{\delta_i}, n_\delta) | i=1,\ldots,n \}
  \cup \{choiceAction(n_\delta)\}.$

  \item For $\delta = \iif \phi \tthen \delta_1 \eelse \delta_2$,
$ r(\delta) = r(\phi) \cup r(\delta_1) \cup r(\delta_2) \cup
\{if(n_\delta, n_\phi,
       n_{\delta_1}, n_{\delta_2})\}.
$
  \item For $\delta = \wwhile \phi \ddo \delta_1$,
$ r(\delta) = r(\phi) \cup r(\delta_1) \cup \{while(n_\delta,
n_\phi,
       n_{\delta_1})\}.
$
  \item For ${\delta = \pick(Y, \{y_1,\ldots,y_m\}, \delta_1)}$,
\[
r(\delta) = \bigcup_{j=1}^{j=m}
 r(\delta_1(y_j))
    \cup R
 \]
where $R = \{choiceArgs(n_\delta, n_{\delta_1(y_j)}) \mid
j=1,\ldots,m)\}$.

\item For ${\delta = p(c_1,\ldots,c_n)}$ where
 $(p(X_1,\ldots,X_n) : \delta_1)$ is a procedure,
$r(\delta) = \{\delta\}$.
\end{itemize}

\ni
A procedure $(p(X_1,\ldots,X_n) : \delta_1)$ is encoded by
the set of rules encoding the collection of its ground
instances. The encoding of a program $(S,\delta)$ consists of
$r(\delta)$ and the rules encoding the procedures in $S$.
Observe that because of $S$'s coherence, the set
of rules encoding a program $(S,\delta)$ is uniquely
determined.

\begin{example}
\label{ex6}
{\rm
In this example we present the encoding of the
program $(S,control)$
from Example~\ref{ex5}.

\st
We start with the set of rules encoding the ground procedure
$(go\_floor(i) : currentFloor(i) \mid up(i) \mid down(i))$
where $i$ is a floor constant. First, we assign the name
$go\_floor(i)$ to the complex action
$currentFloor(i) \mid up(i) \mid down(i)$ and encode this
complex action by the set $r(go\_floor(i))$. This set
consists of the following
facts:
\[
\begin{array}{c}
choiceAction(go\_floor(i)). \\
in(currentFloor(i), go\_floor(i)). \\
in(up(i), go\_floor(i)). \\
in(down(i), go\_floor(i)). \\
\end{array}
\]

\ni
Similar atoms are needed to encode other instances
of the procedure $go\_floor(N)$.

\st For each floor $i$,
the following facts encode the complex action
$go\_floor(i),turnoff(i),open,close$, which is the body
of a ground instance
of the procedure
$(serve(N) : go\_floor(N),turnoff(N),open,close)$:
\[
\begin{array}{c}
sequence(serve(i),go\_floor(i),serve\_tail\_1(i)). \\
sequence(serve\_tail\_1(i),turnoff(i),open\_close). \\
sequence(open\_close,open,close).
\end{array}
\]

\ni To encode the procedure $(serve\_a\_floor :
\pick(N,\{0,1,\ldots,k\}, (on(N);serve(N)))$, we need the set of
rules which encode $serve(i)$, $0 \le i \le k$, (above) and the
rules encode the complex action $on(i); serve(i)$ for every $i$.
These rules are:
\[
sequence(body\_serve\_a\_floor(i), on(i), serve(i)),
\]
where $body\_serve\_a\_floor(i)$ is the name assigned to
the complex action $on(i); serve(i)$, and the
following rule:
\[
choiceArgs(serve\_a\_floor, body\_serve\_a\_floor(i)).
\]
\ni
The following facts encode the procedure $(park :
\mathbf{if} \; currentFloor(0) \; \mathbf{then} \; open \; \mathbf{else} \; [down; park])$:
\[
\begin{array}{cl}
 if(park, currentFloor(0), open, park\_1). \\
 sequence(park\_1,down(0),open).
\end{array}
\]

\ni
Finally, the encoding of the procedure $control$  consists of the
rules encoding the formula
\[
\oor(on(0), \oor(on(1), \ldots, \oor(on(k-1), on(k))))
\]

which is assigned the name '$existOn$' and the following rules:
\[
\begin{array}{cl}
 sequence(control,while\_service\_needed,park). \\
 while(while\_service\_needed, existOn, serve\_a\_floor).\\
\end{array}
\]

\qed
}
\end{example}
\ni We now present the AnsProlog rules that realize the
operational semantics of programs. We define a predicate
$trans(P,T_1,T_2)$ where $P$ is a program and $T_1$ and $T_2$ are
two time points, $T_1 \le T_2$. Intuitively, we would like to have
$trans(p,t_1,t_2)$ be true in an answer set $M$ iff $s_{t_1}(M)
a_{t_1} \ldots a_{t_2-1} s_{t_2}(M)$ is a trace of the program
$p$\footnote{
  Recall that we define $s_i(M) = \{holds(f,i) \in M\  \mid  \ f$
  is a fluent literal$\}$.
}.
\begin{eqnarray}
trans(A,T,T+1)& \la &
    action(A), occ(A,T).  \label{tr_act} \\
trans(F,T_1,T_1)& \la &
     formula(F), hf(F,T_1). \label{tr_form}\\
trans(P,T_1,T_2) & \la &
    sequence(P,P_1,P_2), T1 \le T' \le T_2, \label{tr_proc}\\
&&  trans(P_1,T_1,T'),trans(P_2,T',T_2).\\
trans(N,T_1,T_2) & \la &
     choiceAction(N),  \label{tr_choice}\\
 &&    in(P_1,N),trans(P_1,T_1,T_2). \nonumber \\
trans(I,T_1,T_2)& \la &
    if(I,F,P_1,P_2), hf(F,T_1), trans(P_1,T_1,T_2). \label{tr_if_true}\\
trans(I,T_1,T_2)& \la &
    if(I,F,P_1,P_2), \naf hf(F,T_1),trans(P_2,T_1,T_2).
\label{tr_if_false}\\
trans(W,T_1,T_2)& \la &
    while(W,F,P), hf(F,T_1), T_1 < T' \le T_2,  \label{tr_while_true}\\
 &&    trans(P,T_1,T'), trans(W,T',T_2).\nonumber \\
trans(W,T,T)& \la &
    while(W,F,P), \naf hf(F,T). \label{tr_while_false}\\
trans(S, T_1, T_2)& \la &
    choiceArgs(S, P), trans(P, T_1, T_2). \label{tr_pick}\\
trans(\Null,T,T) & \la &   \label{tr_null}
\end{eqnarray}
\ni Here $\Null$\  denotes a dummy program that performs no
action. This action is added to allow programs of the form $\iif
\varphi \tthen p$ to be considered (this will be represented as
$\iif \varphi \tthen p \eelse \Null$). The rules are used for
determining whether a trajectory -- encoded by answer sets of the
program $\Pi_n$ -- is a trace of a program or not. As with
temporal constraints, this is done inductively over the structure
of programs. The rules (\ref{tr_act}) and  (\ref{tr_form}) are for
programs consisting of an action and a fluent formula
respectively. The other rules are for the remaining cases. For
instance, the rule (\ref{tr_while_true}) states that the
trajectory from $T_1$ to $T_2$ is a trace of a while loop
``$\wwhile F \ddo P$'', named $W$ and
encoded by the atom $while(W,F,P)$, if the formula $F$ holds at
$T_1$ and there exists some $T'$, $T_1 < T' \le T_2$ such that the
trajectory from $T_1$ to $T'$ is a trace of $P$ and the trajectory
from $T'$ to $T_2$ is a trace of $W$; and the rule
(\ref{tr_while_false}) states that the trajectory from $T$ to $T$
is a trace of $W$ if the formula $F$ does not holds at
$T$. These two rules effectively determine whether the trajectory
from $T_1$ to $T_2$ is a trace of $while(W,F,P)$. The meanings of
the other rules are similar.

\st Observe that we do not have
specific rules for complex actions which are procedure calls.
This is because of every trace of a procedure call $p(c_1,\ldots,c_n)$,
where $p(X_1,\ldots,X_n) : \delta)$ is a procedure,
is a trace of the complex action $\delta(c_1,\ldots,c_n)$ ---
whose name is $p(c_1,\ldots,c_n)$, as described earlier ---
and vice versa. Furthermore, $\delta$ is not a procedure call ,
traces of $\delta_1$ can be computed using the above rules.
The correctness of the above set of rules (see Theorem \ref{th2})
means that procedure calls are treated correctly in our implementation.

\st To specify that a plan of length $n$ starting from an initial
state must obey the sketch specified by a program $p=(S,\delta)$,
all we need to do is to add the rules encoding $p$ and the
constraint ${\la \naf trans(n_p,0,n)}$ to $\Pi_n$. We now
formulate the correctness of our above encoding of procedural
knowledge given as programs, and relate the traces of  program
with the answer sets of its AnsProlog encoding. Let
$\Pi^{Golog}_n$ be the program obtained from $\Pi_n$ by (i) adding
the rules (\ref{tr_act})-(\ref{tr_null}) and
(\ref{fol_lit_tran})-(\ref{lt_during}), (ii) adding $r(p)$, and
(iii) replacing the goal constraint with $\bot \la \naf
trans(n_p,0,n)$. The following theorem is similar to Theorem
\ref{th1}.
\begin{theorem}
\label{th2} {\rm Let $(D,\Gamma)$ be a consistent action theory
and $p=(S,\delta)$ be a program. Then,
\begin{itemize}
\item[(i)] for every  answer set $M$ of $\;\Pi^{Golog}_n$ with
$occ(a_i,i) \in M$ for $i \in \{0,\ldots,n-1\}$, $\,s_0(M)a_0
\ldots a_{n-1} s_n(M)\,$ is a trace of $p$; and

\item[(ii)] if $\,s_0a_0\ldots a_{n-1}s_n\,$ is a trace
of $p$ then there exists an answer set $M$ of $\;\Pi^{Golog}_n$
such that $\,s_j = s_j(M)$ and $\;occ(a_i,i) \in M$ for $j \in
\{0,\ldots,n\}$ and $\,i \in \{0,\ldots,n-1\}$.
\end{itemize}
}
\end{theorem}
\proof See Appendix A.3 \qed

\st To do planning using procedural constraints all we
need to do is to add the goal constraint to $\Pi^{Golog}_n$, which will
filter out all answer sets where the goal is not satisfied in time
point $n$, and at the same time will use the sketch provided by
the program $p$.

\subsection{HTN-Based Knowledge}
\label{s33}

The programs in the previous subsections are good for representing
procedural knowledge but prove cumbersome for encoding
partial-ordering information. For example, to represent that any
sequence containing the $n$ programs $p_1,\ldots,p_n$, in which
$p_1$ occurs before $p_2$, is a valid plan for a goal $\Delta$,
one would need to list all the possible sequences and then use the
non-determinism construct. For $n = 3$, the program fragment would
be $(p_1;p_2;p_3 | p_1;p_3;p_2 | p_3;p_1;p_2)$. Alternatively, the
use of the \emph{concurrent construct} $\|$ from \cite{deg00},
where $p\|q$ represents the set consisting of two programs $p;q$
and $q;p$, is not very helpful either.
This deficiency of pure procedural constructs of the type
discussed in the previous section prompted us to look at the
constructs in HTN planning \cite{sac74}. The partial-ordering
information allowed in HTN descriptions serves the purpose. Thus
all we need is to add constraints that says  $p_1$ must occur
before $p_2$.

\st The constructs in HTN by themselves are not expressive enough
either as they do not have procedural constructs such as
procedures, conditionals, or loops, and expressing a while loop
using pure HTN constructs is not trivial. Thus we decided to
combine the HTN and procedural constructs and to go further than the
initial attempt in \cite{baral-son99} where complex programs are
not allowed to occur within HTN programs.

\st We now define a more general notion of program that allows
both procedural and HTN constructs. For that we need the following
notion. Let $\Sigma = \{(p_1 : \delta_1),\ldots,(p_k :
\delta_k)\}$ be a set of procedures  with free variables
$\{X_1,\ldots,X_n\}$.

\begin{itemize}

\item
An ordering constraint over $\Sigma$ has the form $p_i \prec p_j$ where
$p_i \ne p_j$.

\item  A truth constraint is of the form $(p_i,\phi)$,
$(\phi,p_i)$, or $(p_i,\phi,p_t)$, where $\phi$ is a formula
whose free variables are from the set $\{X_1,\ldots,X_n\}$.
\end{itemize}

\ni
Given a set of procedures $\Sigma$ and a set of constraints $C$
over $\Sigma$, the execution of $\Sigma$ will begin with the
grounding of $\Sigma$ and $C$, i.e.,
the variables $X_1,\ldots,X_n$ are substituted by some constants
$c_1,\ldots,c_n$. The constraints in $C$ stipulate an order
in which the procedures in $\Sigma$ is executed. The intuition
behind these types of constraints is as follows:

\begin{itemize}

\item An ordering constraint $p_i \prec p_j$
requires that the procedure $p_i$ has to be executed before the
procedure $p_j$.

\item A truth constraint of the form $(p_i,\phi)$ (resp.
$(\phi,p_i)$) requires that immediately after (resp. immediately
before) the execution of $p_i$, $\phi$ must hold.

\item A constraint of the form $(p_i,\phi,p_t)$
indicates that $\phi$ must hold immediately after $p_i$
is executed until $p_t$ begins its execution.

\end{itemize}

\ni
Because a constraint of the form $(p_i,\phi,p_t)$ implicitly requires
that $p_i$ is executed before $p_t$, for convenience, we
will assume hereafter
that whenever $(p_i,\phi,p_t)$ belongs to $C$, so does
$p_i \prec p_t$.

\st
The definition of general complex actions follows.

\begin{definition}
[General Complex Action]
\label{ca1}
{\rm
For an action theory $(D,\Gamma)$, a general complex action
with variables $X_1,\ldots,X_n$ is either
\begin{itemize}
\item a complex action (Definition \ref{ca});
or
\item a pair $(\Sigma,C)$ where $\Sigma$ is a set of procedures
and $C$ is a set of constraints over $\Sigma$ and the variables
of each procedure in $\Sigma$ are from $X_1,\ldots,X_n$.
\end{itemize}
}
\end{definition}

\ni The definition of a procedure or program does not change. The
notion of ground instantiation, dependency, and well-definedness
of a procedure can be extended straightforwardly to general
programs. We will continue to assume that programs in
consideration are well-defined. As in the case of programs, the
operational semantics of general programs is defined using the
notion of trace. In the next definition, we extend the notion of a
{\em trace} to cover the case of general programs.

\begin{definition}
[Trace of general programs] \label{deftrace2} {\rm Let
$p=(S,\delta)$ be a general program. We say that a trajectory
$s_0a_0\ldots a_{n-1}s_n$ is a trace of $p$ if one of the
following conditions is satisfied:
\begin{itemize}
\item $s_0a_0\ldots a_{n-1}s_n$ and $(S,\delta)$ satisfy one of the
condition in Definition \ref{deftrace1}; or

\item
$\delta = (\Sigma,C)$, $\Sigma = \{(p_1 :
\delta_1),\ldots,(p_k,\delta_k)\}$, and there exists $j_0 {=}0 \le
j_1 \le \ldots \le j_k{=}n$ and a permutation $(i_1,\ldots,i_k)$
of $(1,\ldots,k)$ such that the sequence of trajectories $\alpha_1
= s_0 a_0 \ldots s_{j_1}$, $\alpha_2 = s_{j_1} a_{j_1} \ldots
s_{j_2}$, $\ldots$, $\alpha_k = s_{j_{k-1}} a_{j_{k-1}} \ldots
s_{n}$ satisfies the following conditions:
\begin{enumerate}
\item for each $l$, $1 \le l \le k$, $\alpha_l$ is a trace of
$(S, \delta_{i_l})$,
\item if $n_t \prec n_l \in C$ then $i_t < i_l$,
\item if $(\phi,n_l) \in C$ (or $(n_l,\phi) \in C$)  then $\phi$ holds
in the state $s_{j_{l-1}}$ (or $s_{j_l}$), and
\item if $(n_t,\phi,n_l) \in C$ then $\phi$ holds in
$s_{j_t},\ldots,s_{j_{l-1}}$.
\end{enumerate}
\end{itemize}
\qed
}
\end{definition}
\ni The last item of the above definition can be visualized by the
following illustration:
\begin{eqnarray*}
\underbrace{s_0a_0s_1 \ldots a_{j_1-1}s_{j_1}}_{
\stackrel{\alpha_1}{
\stackrel{\uparrow}
{\stackrel{\downarrow}{\mathrm{trace \ of \ (S,\delta_{i_1})}}}}
}
\underbrace{s_{j_1} a_{j_1} \ldots a_{j_2-1}s_{j_2}}_{
\stackrel{\alpha_2}{
\stackrel{\uparrow}
{\stackrel{\downarrow}{\mathrm{trace \ of \ (S,\delta_{i_2})}}}}
}
\ldots
\underbrace{s_{j_{l-1}} a_{j_{l-1}+1} \ldots a_{j_l-1}s_{j_l}}_{
\stackrel{\alpha_l}{
\stackrel{\uparrow}
{\stackrel{\downarrow}{\mathrm{trace \ of \ (S,\delta_{i_l})}}}}
}
\ldots
\underbrace{s_{j_{k-1}} a_{j_{k-1}+1} \ldots a_{j_k-1}s_{j_k}}_{
\stackrel{\alpha_k}{
\stackrel{\uparrow}
{\stackrel{\downarrow}{\mathrm{trace \ of \ (S,\delta_{i_k})}}}}
}
\end{eqnarray*}

\ni Next we show how to represent general programs. Similar to
programs in the previous section, we will assign names to general
programs and their elements. A general program $p = (S,C)$ is
encoded by the set of atoms and rules
\[ r(p) = \{htn(n_{p}, n_S, n_C)\} \cup r(S) \cup r(C) \]
where $r(S)$ and $r(C)$ is the set of atoms and rules encoding $S$
and $C$ and is described below. Recall that $S$ is a set of
programs and $C$ is a set of constraints. Both $S$ and $C$ are
assigned unique names, $n_S$ and $n_C$. The atoms $set(n_S)$ and
$set(n_C)$ are added to $r(S)$ and $r(C)$ respectively. Each
element of $S$ and $C$ is encoded by a set of rules which are
added to $r(S)$ and $r(C)$, respectively. Finally, the predicate
$in(.,.)$ is used to specify what belongs to $S$ and $C$,
respectively. Elements of $C$ are represented by the predicates
$order({*},{+},{+})$, $postcondition({*},{+},{-})$,
$precondition({*},{-},{+})$, and $maintain({*},{+},{-},{+})$ where
the place holders `*', `+', and `-' denote the name of a
constraint, a general program, and a formula, respectively. For
example, if $n_1 \prec n_2$ belongs to $C$ then the set of atoms
encoding $C$ will contain the atom $in(order(n_0,n_1,n_2),n_C)$
where $n_{0}$ and $n_C$ are the names assigned to the ordering
constraint $n_1 \prec n_2$ and $C$, respectively. Similarly, if
$C$ contains $(n_1,\varphi,n_2)$ then $in(maintain(n_0,n_\varphi,
n_1, n_2), n_C)$ (again, $n_0$ and $n_C$ are the name assigned to
the truth constraint $n_1 \prec n_2$ and $C$, respectively) will
belong to the set of atoms encoding $C$.

\st In the following example, we illustrate the encoding  of a general
program about the blocks world domain.

\begin{example}
{\rm Consider a general program, $(S,C)$, to build a tower from
blocks $a,b,c$ that achieves the goal that $a$ is on top of $b$
and $b$ is on top of $c$, i.e., the goal is to make $on(a,b)
\wedge on(b,c)$ hold. We have $S = \{ move(b,c),  move(a,b)\}$,
and
\[
C = \left\{
\begin{array}{lll}
o & : &   move(b,c) \prec move(a,b), \\
f_1 & : &  (clear(b), move(b,c)), \\
f_2 & : &   (clear(c), move(b,c)), \\
f_3 & : &   (clear(b),move(a,b)), \\
f_4 & : &   (clear(a), move(a,b))
\end{array}
\right\}
\]
The constant preceding the semicolon is the name assigned to the formula
of $C$. The encoding of $p = (S,C)$ is as follows:

\begin{itemize}

\item $r(p) = \{htn(p, n_S, n_C) \} \cup r(S) \cup r(C)$;

\item $r(S) = \{set(n_S), in(move(a,b), n_S), in(move(b,c), n_S)\}$;
and

\item $r(C)$ consists of

\begin{itemize}

\item the facts defining $n_C$ and declaring its elements
\[
\{set(n_C), in(o, n_C), in(f_1,n_C), in(f_2,n_C), in(f_3,n_C),
in(f_4,n_C)\}
\]
\item the facts defining each of the constraints in $C$:
\begin{itemize}

\item the order constraint $o$: $order(o, move(b,c), move(a,b))$,

\item the precondition constraints $f_1,\ldots,f_4$:
\begin{itemize}
\item $precondition(f_1, clear(b), move(b,c))$,
\item $precondition(f_2, clear(c), move(b,c))$,
\item $precondition(f_3, clear(b), move(a,b))$, and
\item $precondition(f_4, clear(a), move(a,b))$.
\end{itemize}
\end{itemize}
\end{itemize}
\end{itemize}
\qed
}
\end{example}

\ni We now present the AnsProlog rules that realize the
operational semantics of general programs. For this purpose we
need the rules (\ref{tr_act})-(\ref{tr_null}) and the rules for
checking the satisfiability of formulae that were presented
earlier. These rules are for general programs whose top level
structure is not an HTN. For general programs whose top level
feature is an HTN we add the following rule:
\begin{eqnarray}
trans(N,T_1,T_2)& \la &
    htn(N,S,C),   \naf nok(N,T_1,T_2). \label{thtn1}
\end{eqnarray}

\ni Intuitively, the above rule states that the general program
$N$ can be unfolded between time points $T_1$ and $T_2$ (or
alternatively: the trajectory from $T_1$ and $T_2$ is a trace of
$N$) if $N$ is an HTN construct $(S,C)$, and it is not the case
that  the trajectory from $T_1$ and $T_2$ is not a trace of $N$.
The last phrase is encoded by $nok(N,T_1,T_2)$ and is true when
the trajectory from $T_1$ and $T_2$ violates one of the many
constraints dictated by $(S, C)$. The main task that now remains
is to present AnsProlog rules that define $nok(N,T_1,T_2)$. To do
that,  as suggested by the definition of a trace of a program
$(S,C)$, we will need to enumerate the permutations
$(i_1,\ldots,i_k)$ of $(1,\ldots,k)$ and check whether particular
permutations satisfy the conditions in $C$. We now introduce some
necessary auxiliary predicates and their intuitive meaning.

\begin{itemize}
\item $begin(N, I, T_3, T_1, T_2)$ -- This means that
$I$, a general program belonging to $N$, starts its execution at
time $T_3$, and $N$ starts and ends its execution at $T_1$ and
$T_2$ respectively.

\item $end(N, I, T_4, T_1, T_2)$ -- This means that
$I$, a general program belonging to $N$, ends its execution at
time $T_4$, and $N$ starts and ends its execution at $T_1$ and
$T_2$, respectively.

\item $between(T_3,T_1,T_2)$ -- This is an auxiliary predicate
indicating that the inequalities $T_1 \le T_3 \le T_2$ hold.

\item $not\_used(N, T, T_1, T_2)$ -- This means that there exists
no sub-program $I$ of $N$ whose execution covers the time moment
$T$, i.e., $T < B$ or $T > E$ where $B$ and $E$ are the start and
finish time of $I$, respectively.

\item $overlap(N, T, T_1, T_2)$ -- This indicates that there
exists at least two general programs $I_1$ and $I_2$ in $N$ whose
intervals contain $T$, i.e., $B_1 < T \le E_1$ and $B_2 < T \le
E_2$ where $B_i$ and $E_i$ ($i=1,2$) is the start- and
finish-time of $I_i$, respectively.

\end{itemize}
\ni We will now give the rules that define the above predicates.
First, to specify that each general program $I$ belonging to the
general program $(S,C)$, i.e., $I \in S$, must start and end its
execution exactly once during the time $(S,C)$ is executed,
we use the following rules:
\begin{eqnarray}
1\{begin(N, I, T_3, T_1, T_2):between(T_3,T_1,T_2)\}1 &\la  &
htn(N,S,C), \label{thtn2}\\
&&             in(I, S), \nonumber \\
& &  trans(N,T_1,T_2). \nonumber \\
1\{end(N, I, T_3, T_1, T_2):between(T_3,T_1,T_2)\}1   &\la  &
    htn(N,S,C), \label{thtn3} \\
& & in(I, S), \nonumber \\
& & trans(N,T_1,T_2).  \nonumber
\end{eqnarray}
\ni The first (resp. second) rule says that $I$ -- a program
belonging to $S$ -- must start (resp. end) its execution exactly
once between $T_1$ and $T_2$. Here, we use cardinality constraints
with variables \cite{nie99b} in expressing these constraints. Such
constraints with variables are short hand for a set of
instantiated rules of the form (\ref{weight}). For example, the
first rule is shorthand for the set of rules corresponding to the
following cardinality constraint:
$$
\begin{array}{lll}
1\{begin(N, I, T_1, T_1, T_2),
\ldots,begin(N, I, T_2, T_1, T_2)\}1 & \la &  htn(N,S,C), \\
 & &            in(I, S),  \\
 & & trans(N,T_1,T_2).
\end{array}
$$
We now give the rules defining $not\_used(., ., ., .)$ and
$overlap(., ., ., .)$.
\begin{eqnarray}
used(N,T,T_1,T_2)& \la &
    htn(N,S,C), in(I,S), \label{thtn11} \\
 & & begin(N, I, B, T_1, T_2), \nonumber \\
 & & end(N, I, E, T_1, T_2), \nonumber \\
&&   B \le T \le E.  \nonumber \\
not\_used(N,T,T_1,T_2)& \la & \naf used(N,T,T_1,T_2).
\label{thtn12} \\
overlap(N, T, T_1, T_2) & \la &
    htn(N,S,C), in(I_1,S),\label{thtn14} \\
& & begin(N, I_1, B_1, T_1, T_2), \nonumber \\
& &    end(N, I_1, E_1, T_1, T_2),  \nonumber \\
&&   in(I_2,S), begin(N, I_2, B_2, T_1, T_2), \nonumber \\
& & end(N, I_2, E_2, T_1, T_2), \nonumber \\
&&   B_1 <  T \le E_1,  B_2 < T \le E_2, I_1 \ne I_2.  \nonumber
\end{eqnarray}

\ni The rule (\ref{thtn11}) states that if a general program $I$
in $N$ starts its execution at $B$ and ends its execution at $E$
then its execution spans over the interval $[B,E]$, i.e., every
time moment between $B$ and $E$ is \emph{used} by some general
program in $N$. The rule (\ref{thtn12}) states that if a time
moment between $T_1$ and $T_2$ is not used by some general program
in $N$ then it is \emph{not\_used}. The last rule in this group
specifies the situation when two general programs belonging to $N$
overlap.

\st We are now ready to define $nok(.,.,.)$. There are several
conditions whose violation make $nok$ true. The first condition
is that the time point when a program starts must occur before its
finish time. Next, each general program belonging to the set $S$ of
$(S,C)$ must have a single start and finish time. The violation
of these two conditions is encoded by the following rules which
are added to $\Pi$.

\begin{eqnarray}
nok(N,T_1,T_2) & \la &
    htn(N,S,C),
     in(I, S), T_3 > T_4,  \label{thtn4}\\
& &     begin(N, I, T_3, T_1, T_2),\nonumber \\
& &     end(N, I, T_4, T_1, T_2). \nonumber \\
nok(N,T_1,T_2)& \la &
    htn(N,S,C),in(I, S),
     T_3 \leq T_4, \label{thtn5}\\
&&     begin(N, I, T_3, T_1, T_2),\nonumber \\
&&     end(N, I, T_4, T_1, T_2), \nonumber \\
&&      \naf trans(I,T_3,T_4). \nonumber \\
nok(N,T_1,T_2) & \la &
    htn(N,S,C), T_1 \le T \le T_2, \label{thtn13} \\
& & not\_used(N,T, T_1,T_2).  \nonumber \\
nok(N,T_1,T_2) & \la &
    htn(N,S,C), T_1 \le T \le T_2,     \label{thtn15}\\
& & overlap(N,T, T_1,T_2).  \nonumber
\end{eqnarray}

\ni Together the rules (\ref{thtn2})-(\ref{thtn15}) define when
the permutation determined by the set of atoms of the form
$begin(N,I,B,T_1,T_2)$ and $end(N,I,E,T_1,T_2)$ violates the
initial part of condition 8 of Definition \ref{deftrace2}. The
rules (\ref{thtn2})-(\ref{thtn3}) require each general program in
$N$ to have a unique start and finish time and the rule
(\ref{thtn4}) encodes the violation when the finish time is
earlier than the start time. The rule (\ref{thtn5}) encodes the
violation when the trace of a general program in $N$  does not
correspond to its start and finish times. The rule (\ref{thtn13})
encodes the violation when some time point on the trajectory of
$N$ is not covered by the trace of a general program in $N$; and
the rule (\ref{thtn15}) encodes the violation when the trace of
two general programs in $N$ overlap.

\st The next group of rules encode the violation of conditions
8(b) -- 8(d) of Definition \ref{deftrace2}.

\begin{eqnarray}
nok(N,T_1,T_2) & \la &
    htn(N, S, C),
    in(I_1, S),
    begin(N, I_1, B_1, T_1, T_2),
\label{thtn7}\\
 && in(I_2, S),
    begin(N, I_2, B_2, T_1, T_2),
\nonumber \\
 &&  in(O, C), order(O, I_1, I_2),
     B_1 > B_2.\nonumber \\
nok(N,T_1,T_2) & \la &
        htn(N,S,C),
        in(I_1, S),
        end(N, I_1, E_1, T_1, T_2),\label{thtn8} \\
 &&     in(I_2, S),
        begin(N, I_2, B_2, T_1, T_2),
        E_1 < T_3 < B_2,
        \nonumber \\
 &&     in(O, C), maintain(O, F, I_1, I_2),
        \naf hf(F,T3).\nonumber \\
nok(N,T_1,T_2) & \la &
        htn(N,S,C),
        in(I, S),
        begin(N, I, B, T_1, T_2),
\label{thtn9}\\
 &&     in(O,C),
        precondition(O, F, I),
        \naf hf(F, B).\nonumber \\
nok(N,T_1,T_2) & \la &
        htn(N,S,C), in(I, S),
        end(N, I, E, T_1, T_2), \label{thtn10} \\
 &&     in(O, C),
        postcondition(O, F, I),
        \naf hf(F, E).\nonumber
\end{eqnarray}
\ni The rule (\ref{thtn7}) encodes the violation when the
constraint $C$ of the general program $N = (S,C)$ contains $I_1
\prec I_2$, but $I_2$ starts earlier than $I_1$. The rule
(\ref{thtn8}) encodes the violation when $C$ contains $(I_1, F,
I_2)$ but the formula $F$ does not hold in some point between the
end of $I_1$ and start of $I_2$. The rules (\ref{thtn9}) and
(\ref{thtn10}) encode the violation when $C$ contains the
constraint $(F, I)$ or $(I, F)$ and $F$ does not hold immediately
before or after respectively, the execution of $I$.

\st  We now formulate the correctness of our above encoding of
procedural and HTN knowledge given as general programs, and relate
the traces of a general program with the answer sets of its
AnsProlog encoding. For an action theory $(D,\Gamma)$ and a
general program $p$, let $\Pi^{HTN}_n$ be the AnsProlog program
obtained from $\Pi_n$ by (i) adding the rules
(\ref{tr_act})-(\ref{tr_null}) and (\ref{thtn1})-(\ref{thtn10}),
(ii) adding $r(p)$, and (iii) replacing the goal constraint with
$\bot \la \naf trans(n_P,0,n)$. The following theorem extends Theorem
\ref{th2}.
\begin{theorem}
\label{th3}
{\rm
Let $(D,\Gamma)$ be a consistent action theory and $p$ be a
general program. Then,
\begin{itemize}
\item[(i)] for every  answer set $M$ of $\;\Pi^{HTN}_n$ with
$occ(a_i,i) \in M$ for $i \in \{0,\ldots,n-1\}$, $\,s_0(M)a_0
\ldots a_{n-1} s_n(M)\,$ is a trace of $p$; and

\item[(ii)] if $\,s_0a_0\ldots a_{n-1}s_n\,$ is a trace
of $p$ then there exists an answer set $M$ of $\;\Pi^{HTN}_n$ such
that $\,s_j = s_j(M)$ and $\;occ(a_i,i) \in M$ for $j \in
\{0,\ldots,n\}$ and $\,i \in \{0,\ldots,n-1\}$ and $trans(n_p,0,n)
\in M$.
\end{itemize}
}
\end{theorem}
\proof See Appendix A.4 \qed

\st As before, to do planning using procedural and HTN
constraints all we
need to do is to add the goal constraint to $\Pi^{HTN}_n$, which will
filter out all answer sets where the goal is not satisfied in time
point $n$, and at the same time will use the sketch provided by
the general program $p$.

\subsection{Demonstration Experiments}
\label{s34}

We tested our implementation with some domains from the general
planning literature and from the AIPS planning competition
\cite{aips00}. In particular, we tested our program with the
Miconic-10 elevator domain. We also tested our program with the
Block domain. In both domains, we conducted tests with procedural
control knowledge. Our motivation was: (i) it has already been
established that well-chosen temporal and hierarchical constraints
will improve a planner's efficiency; (ii) we have previously
experimented with the use of temporal knowledge in the answer set
planning framework \cite{baral01:sp1}; and (iii) we are not aware
of any empirical results indicating the utility of procedural
knowledge in planning, especially in answer set planning. (Note
that \cite{rei00} concentrates on using GOLOG to do planning in
domains with incomplete information, not on exploiting procedural
knowledge in planning.)

We report the results obtained from our experiment with the
elevator example from \cite{rei96} (elp1-elp3) and the Miconic-10
elevator domain (s1-0,\ldots,s5-0), proposed by Schindler Lifts
Ltd. for the AIPS 2000 competition \cite{aips00}. Note that some
of the planners, that competed in AIPS 2000, were unable to solve
this problem. The domain description for this example is described
earlier in Example \ref{ex5} and the {\bf smodels} code can be
downloaded from
\url{http://www.cs.nmsu.edu/~tson/ASPlan/Knowledge}. We use a
direct encoding of procedural knowledge in the Block domain to
avoid the grounding problem. For this reason, we do not include the
results of our experiments on the Block domain in this paper. The
results and the encodings of this domain are available on the above
mentioned web site. We note that the direct encoding
of procedural knowledge in the Block domain yields significantly
better results, both in terms of the time needed to find a
trajectory as well as the number of problems that can be solved.
Whether or not the methodology used in the Block domain can be
generalized and applied to other domains is an interesting
question that we would like to investigate in the near future.

\st The initial state for the elevator planning problem encodes a
set of floors where the light is on and the current position of
the elevator. For instance, $\Gamma = \{on(1), on(3), on(7),
currentFloor(4)\}$. The goal formula is represented by the
conjunction $\wedge_{f \mbox{ is a floor}} \neg on(f)$. Sometimes,
the final position of the elevator is added to the goal. The
planning problem is to find a sequence of actions that will serve
all the floors where the light is on and thus make the $on$
predicate false for all floors, and if required take the elevator
to its destination floor.

\st  Since there are a lot of plans that can achieve the desired
goal, we can use procedural constraints to guide us to
preferable plans. In particular, we can use the procedural knowledge
encoded by the following
set of simple GOLOG programs from \cite{rei96}, which we
earlier discussed in Example~\ref{ex5}.
\begin{eqnarray*}
(go\_floor(N) & : & currentFloor(N) | up(N) | down(N)) \\
(serve(N) & : & go\_floor(N); turnoff(N) ; open; close) \\
(serve\_a\_floor & : & \pick(N, d(N), (on(N); serve(N)))) \\
(park & : & \iif currentFloor(0) \tthen open \eelse [down(0);open])\\
(control & : & [\wwhile \exists N.\{0,1,\ldots,k\} \; [on(N)] \ddo
serve\_a\_floor(N)]; park)
\end{eqnarray*}

\st
We ran experiments  on
a Sony VAIO laptop with 256 MB Ram and an Intel Pentium
4 1.59 GHz processor, using {\bf lparse} version 1.00.13
(Windows, build Aug 12, 2003) and
{\bf smodels} version 2.27. for planning in this
example with and without the procedural control knowledge.
The timings obtained are given in the following table.

\begin{center}
\begin{tabular}{|c|c|c|c|c|c|}
\hline Problem & Plan & \# Person & \# Floors & With Control
& Without Control \\
& Length & & & Knowledge & Knowledge \\ \hline\hline
 elp1 & 8 & 2 & 6 & 0.380 & {\bf 0.140} \\
 elp2 & 12 & 3 & 6 & 0.901 & {\bf 0.230} \\
 elp3 & 16 & 4 & 6 & 2.183 & {\bf 1.381} \\
 elp4 & 20 & 5 & 6 & 4.566 & 79.995 \\
 \hline
 s1-0 & 4 & 1 & 2 & 0.120 & {\bf 0.020} \\
 s2-0 & 7 & 2 & 4 & 1.201 & {\bf 0.080} \\
 s3-0 & 9 & 3 & 6 & 5.537 & {\bf 0.310} \\
 s4-0 & 15 & 4 & 8 & 64.271 & {\bf 15.211} \\
 s5-0 & 19 & 5 & 10 & 260.183 & 1181.158 \\ \hline
\end{tabular}
\end{center}

\st As can be seen, the encoding with control knowledge yields
substantially better performance in situations where the plan
length is longer. In some instances with small plan lengths, as
indicated through boldface in column 6, the speed up due to the use
of procedural knowledge does not make up for the overhead needed
in grounding the knowledge. The output of  {\bf smodels} for each
run is given in the file \emph{result} at the above mentioned URL.
For larger instances of the elevator domain \cite{aips00} (5
persons or more and 10 floors or more), our implementation
terminated prematurely with either a stack overflow error or a
segmentation fault error.

\section{Conclusion}
\label{s4}

In this paper we considered three different kinds of
domain-dependent control knowledge (temporal, procedural and HTN-based)
that are useful in planning. Our approach is declarative and
relies on the language of logic programming with answer set
semantics. We showed that the addition of these three kinds of
control knowledge only involves adding a few more rules to a
planner written in AnsProlog that can plan without any control
knowledge. We formally proved the correctness of our planner, both
in the absence and presence of the control knowledge. Finally, we
did some initial experimentation that illustrates the reduction in
planning time when procedural domain knowledge is used and the
plan length is big.

\st In the past, temporal domain knowledge has been used in planning in
\cite{bacchus00,doherty99:time}. In both cases, the planners are
written in a procedural language, and there is no correctness
proof of the planners. On the other hand the performance of these
planners is much better\footnote{This provides a challenge to the
community developing AnsProlog* systems to develop AnsProlog*
systems that can match or come close to (if not surpass) the
performance of planners with procedural knowledge.}
than our implementation using
AnsProlog. In comparison, our focus in this paper is on the
`knowledge representation' aspects of planning with domain
dependent control knowledge and demonstration of relative
performance gains when such control knowledge is used. Thus, we
present correctness proofs of our planners and stress the ease of
adding the control knowledge to the planner. In this regard, an
interesting observation is that it is straightforward to add
control knowledge from multiple sources or angles. Thus say two
different general programs can be added to the planner, and any
resulting plan must then satisfy the two sketches dictated by the
two general programs.

\st As mentioned earlier our use of HTN-based constraints in
planning is very different from HTN-planning and the recent
HTN-based planner \cite{nau99}. Unlike our approach in this paper,
these planners cannot be separated into two parts: one doing
planning that can plan even in the absence of the knowledge
encoded as HTN and the other encoding the knowledge as an HTN. In
other words, these planners are not extended classical planners
that allow the use of domain knowledge in the form of HTN on top
of a classical planner. The timings of the planner \cite{nau99} on
AIPS 2000 planning domains are very good though. To convince
ourselves of the usefulness of procedural constraints we used
their methodology with respect to procedural domain knowledge and
wrote general programs for planning with blocks world and the
package delivery domain and as in  \cite{nau99} we wrote planners
in a procedural language (the language C to be specific) for these
domains and also observed similar performance. We plan to report
this result in a future work. With our focus on the knowledge
representation aspects we do not further discuss these experiments
here.

\st Although we explored the use of each of the different kinds of domain
knowledge separately, the declarative nature of our approach allows
us to use the different kinds of domain knowledge for the same
planning problem. For example, for a particular planning problem
we may have both temporal domain knowledge and a mixture of
procedural and hierarchical domain knowledge given as a general
program. In such a case, planning will involve finding an action
sequence that follows the sketch dictated by the general program
and at the same time obeys the temporal domain knowledge.  This
distinguishes our work from other related work
\cite{hua99,kau98,baral-son99,mci00} where the domain knowledge
allowed was much more restricted.

\st A byproduct of the way we deal with procedural knowledge is
that, in a propositional environment, our approach to planning
with procedural knowledge can be viewed as an off-line interpreter
for a GOLOG program. Because of the declarative nature of
AnsProlog the correctness of this interpreter is easier to prove
than the earlier interpreters which were mostly written in Prolog.

\appendixhead{URLend}


\begin{acks}
We wish to thank the anonymous reviewers for their
detailed comments and suggestions that have helped us to improve the
paper in several ways. The first two authors would like
to acknowledge the support of the NASA grant NCC2-1232. The fourth
author would like to acknowledge the support of NASA grant
NAG2-1337.  The work of Chitta Baral was also supported in part by
the NSF grant 0070463. The work of Tran Cao Son was also supported
in part by NSF grant 0220590.  The work of Sheila McIlraith was
also supported by NSERC.
A preliminary version of this paper appeared in \cite{sbm00b}.
\end{acks}

\bibliographystyle{acmtrans}

\elecappendix

\section*{Appendix A - Proofs}

We apply the Splitting Theorem and Splitting Sequence Theorem
\cite{lif94a} several times in our proof\footnote{
   To be more precise, we use a modified version of these theorems
   since programs in this paper contain constraints of the form
   (\ref{lprule2}) which were not discussed in \cite{lif94a}.
   The modification is discussed in Appendix B.
}. For ease of reading, the
basic notation and the splitting theorem are included in Appendix
B. Since we assume a propositional language any rule in this paper
can be considered as a collection of its ground instances.
Therefore, throughout the proof, we often say a rule $r$ whenever
we refer to a ground rule $r$. By $lit(\pi)$, we denote the set of
literals of a program $\pi$.

\section*{Appendix A.1 - Proof of Theorem \ref{th1}}

For a planning problem $\langle D,\Gamma,\Delta \rangle$, let
$$\pi = \Pi_n(D,\Gamma,\Delta) \setminus \{(\ref{goal}), \; \bot
\la \naf goal\},$$ i.e., $\pi$ is obtained from
$\Pi_n(D,\Gamma,\Delta)$ by removing the rules encoding $\Delta$
and the constraint $\bot \la \naf goal$. Let $S$ be a set of
literals of the form $holds(l,t)$. Abusing the notation, we say
that $S$ is {\em consistent with respect to {\bf F}} (or
consistent, for short) if for every pair of a fluent $f \in
\mathbf{F}$ and a time moment $t$, $S$ does not contain both
$holds(f,t)$ and $holds(\neg f,t)$. For a set of causal laws $K$
and a set of fluent literals $Y$, let

\begin{equation} \label{def-mky}
\begin{array}{lll}
M_K(Y) = Y \cup \{l &  \mid  & \exists \;
\caused(\{p_1,\ldots,p_k\},l) \in K \;\; s.t. \;\;  Y \models p_1
\wedge \ldots \wedge p_k\}.
\end{array}
\end{equation}

\st It is easy to see that $M_K(Y)$ is a monotonic function
and bounded above by {\bf F}. Thus,
the sequence  $\langle  M_K^i(Y)\rangle_{i < \infty}$, where
$M_K^0(Y) = Y$ and $M_K^{i+1}(Y) = M_K(M_K^i(Y))$ for $i \ge 0$,
is a convergent sequence. In other words, there exists some $i \ge
0$ such that $M_K^{j+1}(Y) = M_K^j(Y)$ for $j \ge i$. Let $C(K,Y)
= \bigcup_{i < \infty} M_K^i(Y)$. We have that $C(K,Y)$ is closed
under $K$. We can prove the following lemma.

\begin{lemma}
\label{ladd00}
For a set of causal laws $K$ and a set of fluent literals $Y$,
\begin{enumerate}
\item $C(K,Y)$ is consistent and $C(K,Y)  = Cl_K(Y)$ iff
$Cl_K(Y)$ is defined; and
\item $C(K,Y)$ is inconsistent iff $Cl_K(Y)$ is undefined.
\end{enumerate}
\end{lemma}

\proof The lemma is trivial for inconsistent set of literals $Y$.
We need to prove it for the case where $Y$ is consistent.

\begin{enumerate}
\item Consider the case where $C(K,Y)$ is consistent. From
the definition of the function $M_K$ in (\ref{def-mky}), we have
that $M_K(S) \subseteq Cl_K(S)$ for every set of literals $S$.
Therefore, $M_K^i(Y) \subseteq Cl_K(Y)$ for every $i$. Thus,
$C(K,Y) \subseteq Cl_K(Y)$. Because $C(K,Y)$ is closed under $K$
and $Cl_K(Y)$ is the least set of literals closed under $K$, we
can conclude that $Cl_K(Y) \subseteq  C(K,Y)$.

\st To complete the proof of the first item, we need to show that
if $Cl_K(Y)$ is defined, then $C(K,Y)$ is consistent and $C(K,Y) =
Cl_K(Y)$. Again, it follows immediately from Equation
(\ref{def-mky}) that $M_K(S) \subseteq Cl_K(S)$ for every
consistent set $S$. This implies that $M_K^i(Y) \subseteq Cl_K(Y)$
for every $i$. Thus, we have that $C(K,Y) \subseteq Cl_K(Y)$. This
implies the consistency of $C(K,Y)$. The equality $C(K,Y) =
Cl_K(Y)$ follows from the closeness of $C(K,Y)$ with respect to
$K$ and the definition of $Cl_K(Y)$. This concludes the first item
of the lemma.

\item
Consider the case where $C(K,Y)$ is inconsistent.
Assume that $Cl_K(Y)$ is defined.
By definition of $M^i_K(Y)$, we know that if $M^i_K(Y)$ is inconsistent
then $M_K^{i+1}(Y)$ is also inconsistent. Thus,
there exists an integer $k$ such that $M_K^k(Y)$ is consistent and
$M_K^{k+1}(Y)$ is inconsistent.
Because of the consistency of $Cl_K(Y)$ and
$M_K^k(Y) \subseteq Cl_K(Y)$,
we conclude that $M_K^{k+1}(Y) \setminus Cl_K(Y) \ne \emptyset$.
Consider $l \in M_K^{k+1}(Y) \setminus Cl_K(Y)$. By the definition
of $M_K$, there exists some static causal law
$$\caused(\{p_1,\ldots,p_n\},l)$$ in $K$ such that
$\{p_1,\ldots,p_n\} \subseteq M^k_K(Y)$. This implies that $Cl_K(Y)$
is not closed under $K$, which contradicts the definition
of $Cl_K(Y)$. This shows that $Cl_K(Y)$ is undefined.

\st
Now, consider the case where $Cl_K(Y)$ is undefined. Using contradiction
and the result of the first item, we can also show that $C(K,Y)$ is
inconsistent. This concludes the proof of the second item of the lemma.
\end{enumerate}
\qed

\begin{lemma}
\label{ladd0} For a set of causal laws $K$ and a set of
fluent literals $Y$, for every integer $k$, the program
consisting of the following rules:
\[
\begin{array}{rcll}
holds(l,k) & \la & holds(l_1,k),\ldots,holds(l_m,k) & (\mbox{if }
\caused(\{l_1,\ldots,l_m\},l) \in K) \\
holds(l,k) & \la & & (\mbox{if } l \in Y) \\
\bot & \la & fluent(f), holds(f, k), holds(\neg f, k) &
(\mbox{if } f \in \mathbf{F})\\
\end{array}
\]

\begin{enumerate}
\item has a unique  answer set $\{holds(l,k) \mid l \in
Cl_K(Y)\}$ iff $Cl_K(Y)$ is defined; and
\item does not have an answer set iff $Cl_K(Y)$ is undefined.
\end{enumerate}
\end{lemma}
\proof Let us denote the given program by $Q$ and $P$ be the
program consisting of the rules of $Q$ with the head different
than $\bot$. Since $P$ is a positive program, we know that it has
a unique answer set, says $X$. It is easy to see that if $X$ is
consistent with respect to {\bf F}, then $X$ is the unique answer
set of $Q$; otherwise, $Q$ does not have an answer set.

\st It is easy to see that $X = \{holds(l,k) \mid l \in C(K,Y)\}$
is the unique answer set of $P$. Consider the two cases:

\begin{enumerate}
\item $Cl_K(Y)$ is defined. Lemma \ref{ladd00} implies that
$C(K,Y) = Cl_K(Y)$, and hence, $X$ is consistent with respect to
{\bf F}. This implies that $X$ is the unique answer set of $Q$.

\item $Cl_K(Y)$ is undefined. Lemma \ref{ladd00} implies that
$C(K,Y)$ is inconsistent, which implies that $X$ is inconsistent
with respect to {\bf F}, and
hence, $Q$ does not have an answer set.
\end{enumerate}
\qed

\st We now prove some useful properties of $\pi$. We will prove
that if $(D,\Gamma)$ is consistent then $\pi$ is consistent (i.e.,
$\pi$ has an answer set) and that $\pi$ correctly implements the
transition function $\Phi$ of $D$. First, we simplify $\pi$ by
using the splitting theorem \cite{lif94a} (Theorem \ref{spl1},
Appendix B). Let $V$ be the set of literals in the language of
$\pi$ whose parameter list does not contain the time parameter,
i.e., $V$ consists of auxiliary atoms of the form $literal(l)$,
$fluent(f)$, $action(a)$, $contrary(l_1,l_2)$.

\st It is easy to see that $V$ is a splitting set of $\pi$.
Furthermore, it is easy to see that the bottom program $b_V(\pi)$
consists of the rules that define actions, fluents, and the rules
(\ref{fol_lit1})--(\ref{ir_contra2}). Obviously, $b_V(\pi)$ is a
positive program, and hence, it has a unique answer set. Let us
denote the unique  answer set of $b_V(\pi)$ by $A_0$. The partial
evaluation of $\pi$ with respect to $(V,A_0)$, $\pi_1 = e_V(\pi
\setminus b_V(\pi), A_0)$, is the collection of the following
rules:
\begin{eqnarray}
holds(l, t{+}1) & \la &
    occ(a, t), holds(l_1, t),\ldots,holds(l_k,t).
    \label{pi2_dyn} \\
&&   \mbox{(if } \causes(a, l,
\{l_1,\ldots,l_k\})
    \in D) \nonumber\\
holds(l, t) & \la &
    holds(l_1, t),\ldots,holds(l_m,t).   \label{pi2_cau}\\
&& \mbox{(if } \caused(\{l_1,\ldots,l_m\}, l) \in D)
\nonumber \\
possible(a,t) & \la &
    holds(l_1, t),\ldots,holds(l_t,t).     \label{pi2_pos}\\
&&  \mbox{(if } \executable(a, \{l_1,\ldots,l_t\}) \in D)
\nonumber \\
holds(l, 0) & \la &        \label{pi2_init}\\
&& \mbox{(if } \initially(l) \in \Gamma) \nonumber \\
occ(a,t) & \la &  possible(a,t), \naf nocc(a,t).     \label{pi2_occ}\\
&&
(\mbox{if } a \mbox{ is an action}) \nonumber \\
nocc(a,t) & \la & occ(b,t).       \label{pi2_nocc} \\
&& (\mbox{for every pair of actions  } a \ne b) \nonumber \\
holds(f, t{+}1)& \la & holds(f, t), \naf holds(\neg f, t{+}1).
    \label{pi2_in1}\\
&& (\mbox{for every fluent } f)  \nonumber \\
holds(\neg f, t{+}1)& \la & holds(\neg f, t), \naf holds(f,
t{+}1).       \label{pi2_in2} \\
&&  (\mbox{for every fluent } f)   \nonumber \\
\bot & \la & holds(f, t), holds(\neg f, t). \label{pi2_constraint} \\
&&    (\mbox{for every fluent } f)   \nonumber
\end{eqnarray}

\ni
It follows from the splitting theorem that to prove the
consistency and correctness of $\pi$ it is enough to prove the
consistency of $\pi_1$ and that $\pi_1$
correctly implements the
transition function $\Phi$ of $D$. We prove this in the next
lemmas.

\begin{lemma}\label{smodel-traject1-add}
Let $(D,\Gamma)$ be a consistent action theory.
Let $X$ be an answer set of $\pi_1$. Then,
\begin{enumerate}
\item $s_t(X)$ is a state of $D$ for every $t \in \{0,\ldots,n\}$
\footnote{
Recall that for every set $Y$, $s_t(Y)$ is
the set $\{f \mid holds(f,t) \in Y\}$}
for every $t \in \{0,\ldots,n-1\}$,
\item if $X$ contains $occ(a,t)$ then $a$ is executable in $s_t(X)$ and
$s_{t+1}(X) \in \Phi(a, s_{t}(X))$, and
\item for every $t \in \{0,\ldots,n-1\}$,
if $occ(a,t) \not\in X$ for every action $a$, then  $s_{t+1}(X) =
s_{t}(X)$.
\end{enumerate}
\end{lemma}
\proof It is easy to see that the sequence $\langle U_t \rangle_{t
= 0}^n$, where

\[
\begin{array}{lll}
U_t & = & \{holds(l,k) \mid l \mbox{ is a literal and } k \le t\} \cup \\
    &   & \{occ(a,k) \mid a \mbox{ is an action and } k \le t\} \cup \\
    &   & \{nocc(a,k) \mid a \mbox{ is an action and } k \le t\} \cup \\
    &   & \{possible(a,k) \mid a \mbox{ is an action and } k \le t\}
      \cup \{ \bot \},
\end{array}
\]
is a splitting sequence of $\pi_1$. Since $X$ is an answer set of
$\pi_1$, by the splitting sequence theorem (Theorem \ref{spl2},
Appendix B), there exists a sequence of sets of literals $\langle
X_t \rangle_{t=0}^n$ such that $X_t \subseteq U_t \setminus
U_{t-1}$, and
\begin{itemize}
\item $X = \bigcup_{i=0}^n X_i$,

\item $X_0$ is an answer set of $b_{U_0}(\pi_1)$ and

\item for every $t > 0$, $X_t$ is an answer set of
$e_{U_t}(b_{U_{t}}(\pi_1) \setminus b_{U_{t-1}}(\pi_1),
\bigcup_{i \le t-1}X_i)$.
\end{itemize}
We will prove the lemma by inductively proving that
for  every $t$, $0 \le t \le n$, the following holds:
\begin{itemize}
\item[(i)] $X_t$ is complete and consistent with
respect to $\mathbf{F}$ in the sense that for each fluent $f$,
$X_t$ contains either $holds(f,t)$ or $holds(\neg f,t)$ but not both,

\item[(ii)] $X_t$ contains at most one atom of the form
$occ(a,t)$,

\item[(iii)] $s_{t}(X_t)$ is a state of $D$, and

\item[(iv)] if $occ(a,t-1) \in X_{t-1}$ then $a$ is executable in
$s_{t-1}(X_{t-1})$ and $s_{t}(X_t) \in \Phi(a, s_{t-1}(X_{t-1}))
$; if no atom of the form $occ(a, t-1)$ belongs to $X_{t-1}$ then
$s_{t-1}(X_{t-1}) = s_{t}(X_{t})$.
\end{itemize}
\ni {\bf Base case:} $t=0$. Trivially, $X_0$ satisfies (iv). So,
we only need to show that  $X_0$ satisfies (i)-(iii). Let $P_0 =
b_{U_0}(\pi_1)$. We have that $P_0$ consists of only rules of the
form (\ref{pi2_cau})-(\ref{pi2_nocc}) and (\ref{pi2_constraint})
with $t = 0$. Let  $Z_0 = \{holds(f,0) \mid f$ is a fluent$\} \cup
\{holds(\neg f,0) \mid f$ is a fluent$\}$. We can easily checked
that $Z_0$ is a splitting set of $P_0$. Thus, by the splitting
theorem, $X_0 = M_0 \cup N_0$ where $M_0$ is an answer set of
$b_{Z_0}(P_0)$ and $N_0$ is an answer set  of $e_{Z_0,M_0} =
e_{Z_0}(P_0 \setminus b_{Z_0}(P_0), M_0)$. Because $M_0$ contains
only literals of the form $holds(l,0)$ and $N_0$ contains only
literals of the form $occ(a,0)$, $nocc(a,0)$, and $possible(a,0)$,
we have that $s_0(X_0) = s_0(M_0)$ and $occ(a,0) \in X_0$ iff
$occ(a,0) \in N_0$. Hence, to prove that $X_0$ satisfies
(i)-(iii), we show that $M_0$ satisfies (i) and (iii) and $N_0$
satisfies (ii).

\st We have that the bottom program $b_{Z_0}(P_0)$ consists of
rules of the form (\ref{pi2_init}) and (\ref{pi2_cau}). Because of
the consistency of $(D,\Gamma)$, we have that $s_0 = \{f \mid
\initially(f) \in \Gamma\}$ is consistent and hence $\{holds(f,0)
\mid \initially(f) \in \Gamma\}$ is consistent with respect to
{\bf F}. It follows from Lemma \ref{ladd0} that $M_0 =
\{holds(f,0) \mid f \in s_0\}$ is the  unique  answer set of
$b_{Z_0}(P_0)$ where $s_0$ is the initial state of $(D,\Gamma)$.
Because of the completeness of $\Gamma$ and the consistency of
$(D,\Gamma)$, we can conclude that $M_0$ is complete and
consistent. Thus, $M_0$ satisfies (i). Furthermore, because
$s_0(M_0) = s_0$, we conclude that $M_0$ satisfies (iii).

\st
The partial evaluation of $P_0$ with respect to $(Z_0,M_0)$,
$e_{Z_0,M_0}$, consists of
\[
e_{Z_0,M_0} = \left \{
\begin{array}{lclr}
possible(a,0) &\la & &\hspace*{1.8cm} (a1) \\
 && (\mbox{if } \executable(a,\{l_1,\ldots,l_m\}) \in D  \\
 & & \mbox{ and } holds(l_i,0)
\in M_0)  \\
 occ(a,0) & \la &  possible(a,0), \naf nocc(a,0). &   (a2)\\
nocc(a,0) & \la & occ(b,0). & (a3)\\ && (\mbox{for every pair of
actions } a \ne b)
\\
& \la & holds(f,0), holds(\neg f,0) & (a4) \\ &&  (\mbox{for every
fluent $f$})
\end{array}
\right.
\]
Let $R$ be the set of atoms occurring in the rule (a1) of $e_{Z_0,M_0}$.
There are two cases:
\begin{itemize}
\item {\bf Case 1:} $R = \emptyset$. Obviously, the empty set is the unique
answer set of $e_{Z_0,M_0}$. Thus, $N_0$ does not contain any atom
of the form $occ(a,0)$.

\item {\bf Case 2:} $R \ne \emptyset$. By applying the splitting theorem
one more time with the splitting set $R$ we can conclude that
$N_0$ is an answer set of $e_{Z_0,M_0}$ if and only if there
exists some action $a$, $possible(a,0) \in R$, and $$N_0 = R \cup
\{occ(a,0)\} \cup \{nocc(b,0) \mid b \textrm { is an action in }
D, b \ne a\}.$$ Thus, $N_0$ contains only one atom of the form
$occ(a,0)$.
\end{itemize}

\ni
The above two cases show that $N_0$ contains at most one
atom of the form $occ(a,0)$. This
concludes the proof of the base case.

\st {\bf Inductive step:} Assume that $X_t$, $t < k$, satisfies
(i)-(iv). We will show that $X_k$ also satisfies (i)-(iv). Let
$M_{k-1} = \bigcup_{t=0}^{k-1} X_t$. The splitting sequence
theorem implies that $X_k$ is an answer set of $P_k$ that consists
of the following rules:
\begin{eqnarray}
holds(l, k) & \la &
    \label{pik_dyn} \\
&&  \mbox{(if } occ(a,k-1) \in M_{k-1}, \; \nonumber\\
&&  \causes(a, l, \{l_1,\ldots,l_m\})   \in D, \;   \nonumber\\
&&   holds(l_i,k-1) \in M_{k-1}) \nonumber\\
holds(l, k) & \la &
    holds(l_1, k),\ldots,holds(l_m,k).   \label{pik_cau}\\
&& \mbox{(if } \caused(\{l_1,\ldots,l_m\}, l) \in D) \nonumber\\
possible(a,k) & \la &
    holds(l_1, k),\ldots,holds(l_t,k). \label{pik_pos}\\
&& \mbox{(if } \executable(a, \{l_1,\ldots,l_t\}) \in D)
\nonumber\\
occ(a,k) & \la &  possible(a,k), \naf nocc(a,k).     \label{pik_occ}\\
&& (\mbox{if } a \mbox{ is an action}) \nonumber\\
nocc(a,k) & \la & occ(b,k).   \label{pik_nocc} \\
&& (\mbox{for every
pair of actions  } a \ne b) \nonumber\\
holds(f, k)& \la & \naf holds(\neg f, k).  \label{pik_in1}\\
&& (\mbox{if } holds(f, k-1) \in M_{k-1}) \nonumber\\
holds(\neg f, k)& \la & \naf holds(f, k).      \label{pik_in2} \\
& & (\mbox{if  } holds(\neg f, k-1) \in M_{k-1}) \nonumber\\
\bot & \la & holds(f, k), holds(\neg f, k).
    \label{pik_constr2}
\end{eqnarray}
 From the constraint (\ref{pik_constr2}), we have that
for every fluent $f$, $X_k$ cannot contain both $holds(f,k)$ and
$holds(\neg f,t)$. This means that $X_k$ is consistent.
We now show that $X_k$ is also complete. Assume the contrary, i.e.,
there exists a fluent $f$ such that neither
$holds(f,k)$ nor $holds(\neg f,f)$ belongs to $X_k$. Because of the
completeness of $s_{k-1}(X_{k-1})$ (Item (i), inductive hypothesis),
either $holds(f,k-1) \in s_{k-1}(X_{k-1})$
or $holds(f,k-1) \not\in s_{k-1}(X_{k-1})$.
If the first case happens, rule (\ref{pik_in1})
belongs to $P_k$, and hence, $X_k$ must contain $holds(f,k)$, which
contradicts our assumption that $holds(f,k) \not\in X_k$. Similarly,
if the second case happens, because of rule (\ref{pik_in2}),
we can conclude that  $holds(\neg f,k) \in X_k$ which is also
a contradiction. Thus, our assumption on the incompleteness of
$X_k$ is incorrect. In other words, we have proved that $X_k$ is
indeed complete and consistent, i.e., (i) is proved for $X_k$.
We now prove the other items of the conclusion.
Let
\[
Y_k = \{holds(l,k) \mid  l \textnormal{  is a fluent literal and }
    holds(l,k) \in X_k\}
\]
and
\[
Z_k = \{holds(l,k) \mid l \textnormal { is a fluent literal}\}.
\]
$Z_k$ is a splitting set of $P_k$. Let $\pi_k = b_{Z_k}(P_k)$.
   From the splitting theorem, we know that $Y_k$ must be an answer
set of the program $(\pi_k)^{Y_k}$ that consists of the following
rules:
\[
\begin{array}{rclrl}
holds(l, k) & \la &    & \hspace*{0.7cm}\mbox{(b1)} \\
 & &  \mbox{(if } occ(a,k-1) \in M_{k-1}, \causes(a, l,\{l_1,\ldots,l_m\})
    \in D, \\
& & \; holds(l_i,k-1) \in M_{k-1} \mbox{ for } i=1,\ldots,m) \\
holds(l, k) & \la &
    holds(l_1, k),\ldots,holds(l_n,k). & \mbox{(b2)}  \\
& & \mbox{(if } \caused(\{l_1,\ldots,l_n\}, l) \in D) &   \\
holds(f, k)& \la &  & \mbox{(b3)}\\
& & (\mbox{ if } holds(f, k-1 ) \in M_{k-1}
\; \mbox{and } holds(\neg f, k ) \not\in Y_k) & \\
holds(\neg f, k)& \la & & \mbox{(b4)} \\
& & (\mbox{ if } holds(\neg f, k-1 ) \in M_{k-1}
\; \mbox{and  } holds( f, k ) \not\in Y_k) &  \\
\bot & \la & holds(f, k), holds(\neg f, k). & \mbox{(b5)} \\
     & & (\mbox{for every fluent } f)
\end{array}
\]
Let $Q_1$ and $Q_2$ be the set of atoms occurring in
the rule (b1) and (b3)-(b4), respectively. Let
 $C_1 = \{l \mid holds(l,k) \in Q_1\}$
and $C_2 = \{l \mid holds(l,k) \in Q_2\}$. There are two cases:

\begin{itemize}

\item {\bf Case 1:}
$M_{k-1}$ does not contain an atom of the form $occ(a,k-1)$.
We have that $Q_1 = \emptyset$ and $C_2 \subseteq
s_{k-1}(X_{k-1})$. From Lemma \ref{ladd0}, we know that
$(\pi_k)^{Y_k}$ has a unique answer set $\{holds(f,k) \mid f \in
Cl_{D_C}(C_2)\} $ which is $Y_k$. Because $s_k(X_k) = \{f \mid f
\in Y_k\}$ and the definition of $Cl_{D_C}$, we have that
$s_k(X_k) \subseteq s_{k-1}(X_{k-1})$. The completeness and
consistency of $s_k(X_k)$ and $s_{k-1}(X_{k-1})$ implies that
$s_k(X_k) = s_{k-1}(X_{k-1})$. Because $X_{k-1}$ satisfies
(i)-(iv), $X_k$ also satisfies (i)-(iv).

\item {\bf Case 2:}
There exists an action $a$ such that $occ(a,k-1) \in M_{k-1}$.
Because of the rule (b1) we have that $C_1 =
E(a,s_{k-1}(X_{k-1}))$. The completeness of $s_k(X_k)$ and
$s_{k-1}(X_{k-1})$ and the rules (b3)-(b4) imply that $C_2 =
s_k(X_k) \cap s_{k-1}(X_{k-1})$. Furthermore, Lemma \ref{ladd0}
implies that $(\pi_k)^{Y_k}$ has a unique answer set $\{holds(f,k)
\mid f \in Cl_{D_C}(C_1 \cup C_2)\}$ which is $Y_k$ (because $Y_k$
is an answer set of $\pi_k$). Hence, $s_k(X_k) =
Cl_{D_C}(E(a,s_{k-1}(X_{k-1})) \cup (s_k(X_k) \cap
s_{k-1}(X_{k-1})))$. This implies that $s_k(X_k) \in
\Phi(a,s_{k-1}(X_{k-1}))$. In other words, we have proved that
$X_k$ satisfies (iii)-(iv).
\end{itemize}
The above two cases show that $X_k$ satisfies (iii) and (iv). It
remains to be shown that $X_k$ contains at most one atom of the
form $occ(a,k)$. Again, by the splitting theorem, we can conclude
that $N_k = X_k \setminus Y_k$ must be an answer set of the
following program
\[
e_{Y_k} = \left \{
\begin{array}{lll}
possible(a,k) & \la & \\
& &    \mbox{(if } \executable(a, \{l_1,\ldots,l_m\}) \in D \\
& &    \mbox{ and } holds(l_i,k) \in Y_k)
 \\
occ(a,k) & \la &  possible(a,k), \naf nocc(a,k). \\
& & (\mbox{if } a \mbox{ is an action})
\\
nocc(a,k) & \la & occ(b,k).   \\
& & (\mbox{for every
pair of actions  } a \ne b)
\end{array}
\right.
\]
Let $R_k$ be the set of atoms occurring in the first rule of
$e_{Y_k}$. Similar to the proof of the base case, we can show that
for every  answer set $N_k$ of  $e_{Y_k}$, either $N_k$ does not
contain an atom of the form $occ(a,k)$ or there exists one and
only one action $a$  such that $possible(a,k) \in R_k$ and $N_K =
R_k \cup \{occ(a,k)\} \cup \{nocc(b,a) \mid b$ is an action, $b
\ne a\}$. In either case, we have that $X_k = Y_k \cup N_k$
satisfies the conditions (ii). The inductive step is proved.

\st The conclusion of the lemma follows immediately from the fact
that $s_t(X) = s_t(X_t)$ for every $t$ and $occ(a,t) \in X$ iff
$occ(a,t) \in X_t$ and $X_t$ satisfies the property (i)-(iv). The
lemma is proved. \qed

\begin{lemma}\label{smodel-traject2-add}
For every trajectory $s_0a_0\ldots a_{n-1}s_n$ in $D$ and a
consistent action theory $(D,\Gamma)$, $\pi_1$ has an answer set
$X$ such that
\begin{enumerate}
\item $s_t(X) = s_t$ for every $t$, $0 \le t \le n$, and
\item $occ(a_t,t) \in X$ for every $t$, $0 \le t \le n-1$.
\end{enumerate}
\end{lemma}
\proof We prove the theorem by constructing an answer set $X$ of
$\pi_1$ that satisfies the Items 1 and 2. Again, we apply the
splitting sequence theorem with the splitting sequence $\langle
U_t \rangle_{t=0}^n$, where
\[
\begin{array}{lll}
U_t & = & \{holds(l,k) \mid l \mbox{ is a literal and } k \le t\}
\cup  \\
& & \{occ(a,k) \mid a \mbox{ is an action and } k \le t\} \cup
\\ && \{nocc(a,k) \mid a \mbox{ is an action and } k \le t\} \cup \\
& &  \{possible(a,k) \mid a \mbox{ is an action and } k \le t\} \cup
\{\bot\}.
\end{array}
\]
For every $t$, $0 \le t \le n$, let $R_t = \{possible(a,t) \mid a$
is executable in $s_t\}$. We define a sequence of sets of literals
$\langle X_t \rangle_{t=0}^n$ as follows.
\begin{itemize}
\item For $0 \le t \le n-1$,
$$
\begin{array}{lll}
X_t & = & \{holds(f,t) \mid f \in s_t\} \cup
 \{occ(a_t,t) \} \cup   \\
    &   &  \{nooc(b,t) \mid b \mbox{ is an action in } D, \; b \ne a_t \}
\cup R_t.
\end{array}
$$
\item If $R_n \ne \emptyset$, then let $a_n$ be an arbitrary
action that is executable in $s_n$ and
$$
\begin{array}{lll}
X_n & = & \{holds(f,n) \mid f \in s_n\} \cup
 \{occ(a_n,n) \} \cup \\
    &   &  \{nooc(b,t) \mid b \mbox{ is an action in } D, \; b \ne a_n
\}
\cup R_n.
\end{array}
$$
\item If $R_n = \emptyset$, then
$$
X_n = \{holds(f,n) \mid f \in s_n\} \cup
 \{nooc(b,t) \mid b \mbox{ is an action in } D\}
\cup R_n,
$$
\end{itemize}

\ni We will prove that  $\langle X_t \rangle_{t=0}^n$ is a
solution to $\pi_1$ with respect to $\langle U_t \rangle_{t=0}^n$.
This amounts to prove that
\begin{itemize}
\item $X_0$ is an answer set of $b_{U_0}(\pi_1)$ and

\item for every $t > 0$, $X_t$ is an answer set of
$e_{U_t}(b_{U_{t}}(\pi_1) \setminus b_{U_{t-1}}(\pi_1),
\bigcup_{i \le t-1}X_i)$.
\end{itemize}
We first prove that $X_0$ is an answer set of $P_0 =
b_{U_0}(\pi_1)$. By the construction of $P_0$ and $X_0$, we have
that $(P_0)^{X_0}$ consists of the following rules:
\[
(P_0)^{X_0} = \left \{
\begin{array}{rclr}
holds(f,0) & \la & (\mbox{if } \initially(f) \in \Gamma) &  (a1) \\
holds(l,0) & \la &
    holds(l_1, 0),\ldots,holds(l_m,0).& \mbox{(a2)}  \\
& & \mbox{(if } \caused(\{l_1,\ldots,l_m\}, l) \in D) \\
possible(a,0) &\la & holds(l_1,0),\ldots,holds(l_m,0). &
\hspace*{1.7cm}(a3)
\\ && (\mbox{if }
\executable(a,\{l_1,\ldots,l_m\}) \in D)  \\
 occ(a_0,0) & \la &  possible(a_0,0). &   (a4)\\
nocc(b,0) & \la & occ(a,0). & (a5)\\
&& (\mbox{for every pair of actions } b \ne a)
\\
\bot & \la & holds(f,0), holds(\neg f,0) & (a6) \\ &&  (\mbox{for every
fluent $f$})
\end{array}
\right.
\]
We will show that $X_0$ is a minimal set of literals closed under
the rules (a1)-(a6) and therefore is an answer set of $P_0$. Since
$holds(f,0) \in X_0$ iff $f \in s_0$ (Definition of $X_0$) and $f
\in s_0$ iff $\initially(f) \in \Gamma$ (Definition of $s_0$), we
conclude that $X_0$ is closed under the rule of the form (a1).
Because of $s_0$ is closed under the static causal laws in $D$, we
conclude that $X_0$ is closed under the rule of the form (a2). The
definition of $R_0$ guarantees that $X_0$ is closed under the rule
of the form (a3). Since $s_0a_0\ldots a_{n-1}s_n$ is a trajectory
of $D$, $a_0$ is executable in $S_0$. This implies that
$possible(a_0,0) \in R_0$. This, together with the fact that
$occ(a_0,0) \in X_0$, implies that $X_0$ is closed under the rule
(a4). The construction of $X_0$ also implies that $X_0$ is closed
under the rule (a5). Finally, because of the consistency of
$\Gamma$, we have that $X_0$ does not contain $holds(f,0)$ and
$holds(neg(f),0)$ for any fluent $f$. Thus, $X_0$ is closed under
the rules of $(P_0)^{X_0}$.

\st To complete the proof, we need
to show that $X_0$ is minimal.
Consider an arbitrary set of atoms $X'$ that is closed under the rules
(a1)-(a6).
This implies the following:
\begin{itemize}
\item $holds(f,0) \in X'$ for every $f \in s_0$ (because of the rule
(a1)).

\item $R_0 \subset X'$ (because of the rule (a3) and the definition of
$R_0$).

\item $occ(a_0,0) \in X'$ (because of the rule (a4)).

\item $\{nocc(b,0) \mid b $ is an action, $b \ne a\} \subseteq X'$
(because $occ(a_0,0) \in X'$ and the rule (a5)).
\end{itemize}
The above items imply that $X_0 \subseteq X'$. In other words, we
show that $X_0$ is a minimal set of literals that is closed under
the rules (a1)-(a6). This implies that $X_0$ is an answer set of
$(P_0)^{X_0}$, which implies that $X_0$ is an answer set of $P_0$.

\st To complete the proof of the lemma, we will prove by induction
over $t$, $t > 0$, that $X_t$ is an answer set of $P_t =
e_{U_t}(b_{U_{t-1}}(\pi_1) \setminus b_{U_{t-1}}(\pi_1),
\bigcup_{i \le t-1}X_i)$. Since the proof of the base case ($t=1$)
and the inductive step is similar, we skip the base case and
present only the proof for the inductive step. Now, assuming that
$X_t$, $t < k$, is an answer set of $P_t$. We show that $X_k$ is
an answer set of $P_k$. Let $M_{k-1} = \bigcup_{i \le k-1}X_i$.
The program $P_k$ consists of the following rules:
\begin{eqnarray}
holds(l, k) & \la &
    \label{pik1_dyn} \\
&&  \mbox{(if } \causes(a_{k-1}, l, \{l_1,\ldots,l_m\})   \in D, \;
holds(l_i,k-1) \in M_{k-1}) \nonumber\\
holds(l, k) & \la &
    holds(l_1, k),\ldots,holds(l_m,k).   \label{pik1_cau}\\
&& \mbox{(if } \caused(\{l_1,\ldots,l_m\}, l) \in D) \nonumber\\
possible(a,k) & \la &
    holds(l_1, k),\ldots,holds(l_t,k). \label{pik1_pos}\\
&& \mbox{(if } \executable(a, \{l_1,\ldots,l_t\}) \in D)
\nonumber\\
occ(a,k) & \la &  possible(a,k), \naf nocc(a,k).     \label{pik1_occ}\\
&& (\mbox{if } a \mbox{ is an action}) \nonumber\\
nocc(a,k) & \la & occ(b,k).   \label{pik1_nocc} \\
&& (\mbox{for every
pair of actions  } a \ne b) \nonumber\\
holds(f, k)& \la & \naf holds(\neg f, k).  \label{pik1_in1}\\
&& (\mbox{if } holds(f, k-1) \in M_{k-1}) \nonumber\\
holds(\neg f, k)& \la & \naf holds(f, k).      \label{pik1_in2} \\
& & (\mbox{if  } holds(\neg f, k-1) \in M_{k-1}) \nonumber\\
\bot & \la & holds(f, k), holds(\neg f, k).
    \label{pik1_constr2}
\end{eqnarray}
It is easy to see that $P_k$ can be split by the set of literal
$Z_k = \{holds(f,k) \mid f$ is a fluent literal$\}$ and the bottom
program $\pi_k = b_{Z_k}(P_k)$ consists of the rules
(\ref{pik1_dyn})-(\ref{pik1_cau}) and
(\ref{pik1_in1})-(\ref{pik1_in2}). We will prove first that $Y_k =
\{holds(l,k) \mid holds(l,k) \in X_k\}$ is an answer set of the
program $(\pi_k)^{Y_k}$ that consists of the following rules:
\[
\begin{array}{rclrl}
holds(l, k) & \la &    & \hspace*{1.5cm}\mbox{(b1)} \\
 & &  \mbox{(if } \causes(a_{k-1}, l,\{l_1,\ldots,l_m\})
    \in D, \\
 & & \; holds(l_i,k-1) \in M_{k-1} \mbox{ for } i=1,\ldots,m) \\
holds(l, k) & \la &
    holds(l_1, k),\ldots,holds(l_n,k). & \mbox{(b2)}  \\
& & \mbox{(if } \caused(\{l_1,\ldots,l_n\}, l) \in D) &   \\
holds(f, k)& \la &  & \mbox{(b3)}\\
& & (\mbox{if } holds(f, k-1 ) \in M_{k-1}
\; \mbox{and } holds(\neg f, k ) \not\in Y_k) & \\
holds(\neg f, k)& \la & & \mbox{(b4)} \\
& & (\mbox{if } holds(\neg f, k-1 ) \in M_{k-1}
\; \mbox{and  } holds( f, k ) \not\in Y_k) &  \\
\bot & \la & holds(f, k), holds(\neg f, k). & \mbox{(b5)}
\end{array}
\]

\ni Let $Q_1$ and $Q_2$ be the set of atoms occurring in the rule
(b1) and (b3)-(b4), respectively. Let $C_1 = \{l \mid holds(l,k)
\in Q_1\}$ and $C_2 = \{l \mid holds(l,k) \in Q_2\}$. Rule (b1)
and the fact that $f \in s_{k-1}(X_{k-1})$ iff $holds(f,k-1) \in
M_{k-1}$ imply that $C_1 = E(a_{k-1},s_{k-1}(X_{k-1}))$. Similar
argument allows us to conclude that $C_2 = s_{k}(X_k) \cap
s_{k-1}(X_{k-1})$. Lemma \ref{ladd0} implies that $(\pi_k)^{Y_k}$
has a unique answer set $Y = \{holds(f,k) \mid f \in Cl_{D_C}(C_1
\cup C_2)\}$. Since $occ(a_{k-1},k-1) \in M_{k-1}$ and $s_k(X_k)
\in \Phi(a_{k-1},s_{k-1}(X_{k-1}))$, we have that $s_k(X_k) =
Cl_{D_C}(C_1 \cup C_2)$. It follows from the definition of $Y_k$
that $Y_k = Y$. Thus, $Y_k$ is an answer set of $\pi_k$. It
follows from the splitting theorem that to complete the proof of
the inductive step, we need to show that $N_k = X_k \setminus Y_k$
is an answer set of the partial evaluation of $P_k$ with respect
to $(Z_k,Y_k)$, $e_{Z_k,Y_k} = e_{Z_k}(P_k \setminus b_{Z_k}(P_k),
X_k)$, which is the following program
\[
e_{Z_k,Y_k} = \left \{
\begin{array}{rcl}
possible(a,k) & \la &
    \mbox{(if } \executable(a, \{l_1,\ldots,l_m\}) \in D \\
              &   & \mbox{and } holds(l_i,k) \in Y_k)
 \\
occ(a,k) & \la &  possible(a,k), \naf nocc(a,k). \\
& & (\mbox{if } a \mbox{ is an action})
\\
nocc(a,k) & \la & occ(b,k).   \\
& &  (\mbox{for every pair of actions  } a \ne b)
\end{array}
\right.
\]
It is easy to see that the reduct of $e_{Z_k,Y_k}$ with respect to
$N_k$, $(e_{Z_k,Y_k})^{N_k}$, consists of the following rules
\[
(e_{Z_k,Y_k})^{N_k} = \left \{
\begin{array}{rcl}
possible(a,k) & \la &
    \mbox{(if } \executable(a, \{l_1,\ldots,l_m\}) \in D \\
   & &  \mbox{and } holds(l_i,k) \in Y_k)
 \\
occ(a_k,k) & \la &  possible(a_k,k). \\
nocc(a,k) & \la & occ(b,k).  \\
& & (\mbox{for every pair of actions  } a \ne b)
\end{array}
\right.
\]
Let $R_k$ be the set of atoms occurring in the first rule of
$(e_{Z_k,Y_k})^{N_k}$. Because $s_0a_0\ldots a_ns_n$ is a
trajectory in $D$, $a_k$ is executable in $s_{k}$. Thus,
$possible(a_k,k)$ belongs to $R_k$. It is easy to see that $N_k$
is the unique answer set of $(e_{Z_k,Y_k})^{N_k}$. In other words,
$N_k$ is an answer set of $e_{Z_k,Y_k}$. The inductive step is
proved.

\st The property of $X_t$ implies that the sequence $\langle X_t
\rangle_{t=0}^n$ is a solution to $\pi_1$ with respect to the
sequence $\langle U_t \rangle_{t=0}^n$. By the splitting sequence
theorem, $X = \bigcup_{t=0}^n X_t$ is an answer set of $\pi_1$.
Because of the construction of $X_t$, we have that $s_t(X) =
s_t(X_t) = s_t$ for every $t$ and $occ(a_t,t) \in X$ for every
$t$, $0 \le t \le n$. The lemma is proved. \qed

\st The above lemmas lead to the following corollaries.

\begin{corollary}\label{smodel-traject1}
Let $X$ be an answer set of $\pi$. Then,
\begin{itemize}
\item[(i)] $s_t(X)$ is a state of $D$ for every $t$, $0 \le t
\le n$,
\item[(ii)] if $X$ contains $occ(a,t)$ then $a$ is executable in
$s_t(X)$
and
$s_{t+1}(X) \in \Phi(a, s_{t}(X))$ for every $t$, $0 \le t
\le n-1$, and
\item[(iii)] if $occ(a,t) \not\in X$ for every action $a$,
then  $s_{t+1}(X) = s_{t}(X)$ for every $t$, $0 \le t \le n-1$.
\end{itemize}
\end{corollary}
\proof It follows from the splitting theorem that $Y = X \cap
lit(\pi_1)$ is an answer set of $\pi_1$. Because $s_t(X) = s_t(Y)$
and Lemma \ref{smodel-traject1-add}, we conclude that $X$
satisfies the (i)-(iii). \qed

\begin{corollary}\label{smodel-traject2}
For every trajectory $s_0a_0\ldots a_{n-1}s_n$ in $D$ and a
consistent action theory $(D,\Gamma)$, $\pi$ has an answer set $X$
such that
\begin{itemize}
\item[(i)] $s_t(X) = s_t$  for every $t$, $0 \le t \le n$, and
\item[(ii)] $occ(a_t,t) \in X$  for every $t$, $0 \le t \le n-1$.
\end{itemize}
\end{corollary}
\proof From Lemma \ref{smodel-traject2-add}, there exists an
answer set $Y$ of $\pi_1$ such that $s_t(Y) = s_t$ and $occ(a_t,t)
\in Y$. Again, from the splitting theorem, we can conclude that
there exists an answer set $X$ of $\pi$ such that $Y = X \cap
lit(\pi_1)$. Because $s_t(X) = s_t(Y)$, we conclude that $X$
satisfies (i)-(ii). \qed

\st The next observation is also useful.

\begin{observation}\label{smodel-observation}
For every answer set $X$ of $\pi$, if there exists an $t$ such
that $X$ does not contain an atom of the form $occ(a,t)$, then $X$
does not contain an atom of the form $occ(a,t')$ for $t \le t'$.
\end{observation}

\ni
Using the result of the above
corollaries we can prove Theorem \ref{th1}.

\setcounter{theorem}{0}
\begin{theorem}
{\rm
For a planning problem  $\lan D,\Gamma,\Delta \ran$,
\begin{itemize}
\item[(i)]
if $s_0a_0\ldots a_{n-1}s_n$ is a trajectory achieving $\Delta$, then
there exists an answer set $M$ of $\Pi_n$ such that
\begin{enumerate}
\item $occ(a_i,i) \in M$ for  $i \in \{0,\ldots,n-1\}$ and
\item
$s_i = s_i(M)$ for $i \in \{0,\ldots,n\}$.
\end{enumerate}
and
\item[(ii)]
if $M$ is an answer set of $\Pi_n$, then there exists an integer
$0 \le k \le n$ such that $s_0(M)a_{0}\ldots a_{k-1}s_k(M)$ is a
trajectory achieving $\Delta$ where $occ(a_{i},i) \in M$ for $0 \le i <
k$ and if $k < n$ then no action is executable in
the state $s_k(M)$.
\end{itemize}
}
\end{theorem}
\proof We have that $\Pi_n = \pi \cup \{(\ref{goal}), \leftarrow
\naf goal\}$. Assume that $\Delta = p_1 \wedge \ldots \wedge p_k$.

\st {\bf (i)}. Since $s_0a_0\ldots a_{n-1}s_n$ is a trajectory
achieving $\Delta$,  the existence of $X$ that satisfies the
condition (i) of the theorem follows from Corollary
\ref{smodel-traject2}. Furthermore, because of $s_n \models
\Delta$, we can conclude that $holds(p_i, n) \in X$ for every $i$,
$1 \le i \le k$. Thus, $X \cup \{goal\}$ is an answer set of
$\Pi_n$. This implies the existence of $M$ satisfying (i).

\st {\bf (ii)}. Let $M$ be an answer set of $\Pi_n$. It is easy to
see that this happens iff $goal \in M$ and $X = M \setminus
\{goal\}$ is an answer set of $\pi$ and $holds(p_i, n) \in X$ for
every $i$, $1 \le i \le k$. It follows from Observation
\ref{smodel-observation} that there exists an integer $k \le n$
such that for each $i$, $0 \le i < k$, there exists an action
$a_i$ such that $occ(a_i,i) \in M$ and for $t \ge k$, $occ(a,t)
\not\in M$ for every action $a$. By Corollary
\ref{smodel-traject1}, we know that $a_i$ is executable in
$s_i(M)$ and $s_{i+1}(M) \in \Phi(a_i, s_i(M))$. This means that
$s_0(M) a_0 \ldots a_{k-1} s_k(M)$ is a trajectory and $s_k(M) =
s_n(M)$. Moreover, $\Delta$ holds in $s_n(M)=s_k(M)$. Thus,
$s_0(M) a_0 \ldots a_{k-1} s_k(M)$ is a trajectory achieving
$\Delta$. Furthermore, it follows from Corollary
\ref{smodel-traject1} and the rules (\ref{pi2_occ}) and
(\ref{pi2_nocc}) that if $k < n$ then $M$ does not contain
literals of the form $possible(a,k)$. This implies that no action
is executable in $s_k(M)$ if $k<n$. \qed

\section*{Appendix A.2 - Proofs of Theorem \ref{the-lt-sem}}

\setcounter{theorem}{1}
\begin{theorem}
{ \rm Let $S$ be a finite set of goal-independent temporal
formulae, $I = \langle s_0, s_1\ldots s_n \rangle$ be a sequence
of states, and
$$\Pi_{formula}(S,I) = \Pi_{formula}
\cup r(I) \cup r(S)$$ where
\begin{itemize}
\item $r(S)$ is the set of atoms used in encoding $S$, and
\item
$r(I) = \cup_{t=0}^{n} \{holds(l, t) \mid l$ is a fluent literal
and $l \in s_t\}$.
\end{itemize}

\ni
Then,
\begin{itemize}
\item[(i)] The program $\Pi_{formula}(S,I)$ has a unique answer set, $X$.
\item[(ii)] For every temporal formula $\phi$
such that $formula(n_\phi) \in r(S)$, $\phi$ is true in $I_t$,
i.e., $I_t \models \phi$, if and only if $hf(n_\phi,t)$ belongs to
$X$ where $I_t = \langle s_t, \ldots s_n \rangle$.
\end{itemize}
}

\end{theorem}
\proof First, we prove (i). We know that if a program is locally
stratified then it has a unique answer set. We will show that
$\Pi_{formula}(S,I)$ (more precisely, the set of ground rules of
$\Pi_{formula}(S,I)$) is indeed locally stratified. To accomplish
that we need to find a mapping $\lambda$ from literals of
$\Pi_{formula}(S,I)$ to $\mathbf{N}$ that has the property: if
$A_0 \leftarrow A_1, A_2, \ldots A_n, \naf B_1, \naf B_2, \ldots
\naf B_m$ is a rule in $\Pi_{formula}(S,I)$, then $\lambda(A_0)
\geq \lambda(A_i)$ for all $1\leq i\leq n$ and $\lambda(A_0) >
\lambda(B_j)$ for all $1\leq j\leq m$.

To define $\lambda$, we first associate to each constant $\phi$
that occurs as the first parameter of the predicate $formula(.)$
in $\Pi_{formula}(S,I)$ a non-negative number $\sigma(\phi)$ as
follows.
\begin{itemize}
\item $\sigma(l) = 0$ if $l$ is a literal (recall that if $l$
is a literal then $n_l = l$).

\item $\sigma(n_\phi) = \sigma(n_{\phi_1}) + 1$
if $\phi$ has the form $\negation(\phi_1)$,
$\next(\phi_1)$, $\eventually(\phi_1)$, or $\always(\phi_1)$.

\item $\sigma(n_\phi) = \max \{\sigma(n_{\phi_1}), \sigma(n_{\phi_2})\} + 1$
if $\phi$ has the form $\aand(\phi_1,\phi_2)$,
$\oor(\phi_1,\phi_2)$,  or $\until(\phi_1,\phi_2)$.
\end{itemize}

\ni We define $\lambda$ as follows.
\begin{itemize}
\item $\lambda(hf(n_\phi,t)) = 5 * \sigma(n_\phi) + 2$,
\item $\lambda(hf\_during(n_\phi,t,t')) = 5 * \sigma(n_\phi) + 4$, and
\item $\lambda(l) = 0$ for every other literal of $\Pi_{formula}(S,I)$.
\end{itemize}

\ni
Examining all the rules in $\Pi_{formula}(S,I)$, we can verify that
$\lambda$
has the necessary property.

\st We now prove (ii). Let $X$ be the  answer set of
$\Pi_{formula}(S,I)$. We prove the second conclusion of the lemma
by induction over $\sigma(n_\phi)$.

\st {\bf Base:} Let $\phi$ be a formula with $\sigma(n_\phi) = 0$.
By the definition of $\sigma$, we know that $\phi$ is a literal.
Then $\phi$ is true in $s_t$ iff $\phi$ is in $s_t$, that is, iff
$holds(\phi,t)$ belongs to $X$, which, because of rule
(\ref{fol_lit_tran}), proves the base case.

\st {\bf Step:} Assume that for all $0\leq j\leq k$ and formula
$\phi$ such that $\sigma(n_\phi) = j$, the formula $\phi$ is true
in $s_t$ iff $hf(n_\phi,t)$ is in $X$.

\st Let $\phi$ be such a formula that $\sigma(n_\phi) = k + 1$.
Because of the definition of $\sigma$, $\phi$ is a non-atomic
formula. We have the following cases:
\begin{itemize}
\item {\bf Case 1:} $\phi = \negation(\phi_1)$.
We have that $\sigma(n_{\phi_1}) = \sigma(n_\phi) - 1 = k$.
Because of $formula(n_\phi) \in X$ and
$negation(n_\phi,n_{\phi_1}) \in X$, $hf(n_\phi,t) \in X$ iif the
body of rule (\ref{fol_neg}) is satisfied by $X$ iff
$hf(n_{\phi_1}, t) \notin X$ iff $s_t \not\models \phi_1$ (by
inductive hypothesis) iff $s_t \models \phi$.
\item {\bf Case 2:} $\phi = \aand(\phi_1, \phi_2)$.
Similar to the first case, it follows from the rule
(\ref{fol_and}) and the facts $formula(n_\phi)$ and $and(n_\phi,
n_{\phi_1}, n_{\phi_2})$ that $hf(n_\phi,t) \in X$ iif the body of
rule (\ref{fol_and}) is satisfied by $X$ iff $hf(n_{\phi_1}, t)
\in X$ and $hf(n_{\phi_2}, t) \in X$ iff $s_t \models \phi_1$ and
$s_t \models \phi_2$ (inductive hypothesis) iff $s_t \models
\phi$.
\item {\bf Case 3:} $\phi = \oor(\phi_1, \phi_2)$.
The proof is similar to the above cases, relying on the two rules
(\ref{fol_or1}), (\ref{fol_or2}), and the fact $formula(n_\phi)
\in X$ and $or(n_\phi, n_{\phi_1}, n_{\phi_2}) \in X$.

\item {\bf Case 4:} $\phi = \until(\phi_1,\phi_2)$.
We have that $\sigma(n_{\phi_1})\leq k$ and $\sigma(n_{\phi_2})\leq k$.
Assume that $I_t \models \phi$. By Definition \ref{tpl-semantics},
there exists $t \leq t_2 \leq n$ such that $I_{t_2} \models
\phi_2$ and for all $t \leq t_1 < t_2$, $I_{t_1} \models \phi_1$.
By inductive hypothesis, $hf(n_{\phi_2},t_2) \in X$ and for all
$t_1$, $t \leq t_1 < t_2$, $hf(n_{\phi_1},t_1) \in X$. It follows
that $hf\_during(n_{\phi_1},t,t_2) \in X$. Because of rule
(\ref{lt_until}), we have $hf(n_\phi,t) \in X$.

On the other hand, if $hf(n_\phi,t) \in X$,
because the only rule supporting $hf(n_\phi,t)$ is (\ref{lt_until}),
there exists $t \leq t_2 \leq n$ such that
$hf\_during(n_{\phi_1},t,t_2) \in X$ and $hf(n_{\phi_2},t_2)$. It
follows from $hf\_during(n_{\phi_1},t,t_2) \in X$ that
$hf(n_{\phi_1},t_1) \in X$ for all $t \leq t_1 < t_2$. By
inductive hypothesis, we have $I_{t_1} \models \phi_1$ for all $t
\leq t_1 < t_2$ and $I_{t_2} \models \phi_2$. Thus $I_t \models
\until(\phi_1,\phi_2)$, i.e., $I_t \models \phi$.
\item {\bf Case 5:} $\phi = \next(\phi_1)$. Note that
$\sigma(n_{\phi_1})\leq k$. Rule (\ref{lt_next}) is the only rule supporting
$hf(n_\phi,t)$ where $\phi = \next(\phi_1)$. So $hf(n_\phi,t) \in
X$ iff  $hf(n_{\phi_1},t+1) \in X$ iff $I_{t+1} \models \phi_1$
iff $I_t \models \next(\phi_1)$ iff $I_t \models \phi$.
\item {\bf Case 6:} $\phi = \always(\phi_1)$.
We note that $\sigma(n_{\phi_1}) \leq k$. Observe that $hf(n_\phi,t)$
is supported only by rule (\ref{lt_always}). So $hf(n_\phi,t) \in
X$ iff $hf\_during(n_{\phi_1},t,n) \in X$. The latter happens iff
$hf(n_{\phi_1},t_1) \in X$ for all $t \leq t_1 \leq n$, that is,
iff $I_{t_1} \models \phi_1$ for all $t \leq t_1 \leq n$ which is
equivalent to $I_t \models \always(\phi_1)$, i.e., iff $I_t
\models \phi$.
\item {\bf Case 7:} $\phi = \eventually(\phi_1)$.
We know that $hf(n_\phi,t)\in X$ is supported only by rule
(\ref{lt_event}). So $hf(n_\phi,t) \in X$ iff there exists $t \leq
t_1 \leq n$ such that $hf(n_{\phi_1},t_1) \in X$. Because
$\sigma(n_{\phi_1}) \leq k$, by induction, $hf(n_\phi,t) \in X$ iff
there exists $t \leq t_1 \leq n$ such that $I_{t_1} \models
\phi_1$, that is, iff $I_t \models \eventually(\phi_1)$, i.e., iff
$I_t \models \phi$.
\end{itemize}
\ni
The above cases prove the inductive step, and hence, the theorem.
\qed

\section*{Appendix A.3 - Proof of Theorem \ref{th2}}

We first prove some lemmas that are needed for proving Theorem
\ref{th2}. Abusing the notation, by $\pi_{f}$ we denote the
program consisting of the rules of $\pi$ (Appendix A.1) and the
set of rules $\Pi_{formula}$ where the time constant $T$ takes the
value between $0$ and $n$.

\begin{lemma}\label{lemtrace1} For a consistent
action theory $(D,\Gamma)$, a ground complex action $p$, and an answer
set $M$ of $\Pi_n^{Golog}$ with $occ(a_i,i) \in M$ for $i \in
\{0,\ldots,n-1\}$, $s_{0}(M)a_{0}s_{1}(M)\ldots a_{n-1}s_{n}(M)$
is a trace of $p$.
\end{lemma}
\proof It is easy to see that the union of the set of
literals of $\pi_{f}$
 and the set of rules and atoms encoding
$p$, i.e., $U = lit(\pi_{f}) \cup r(p)$, is a splitting set of
$\Pi_n^{Golog}$. Furthermore, $b_U(\Pi_n^{Golog}) = \pi_{f} \cup
r(p)$. Thus, by the splitting theorem,  $M$ is an answer set of
$\Pi^{Golog}_n$ iff $M = X \cup Y$ where $X$ is an answer set of
$\pi_{f} \cup r(p)$, and $Y$ is an answer set of
$e_U(\Pi_n^{Golog} \setminus \pi_{f},X)$. Because of the
constraint $\bot \la \naf trans(n_p,0,n)$, we know that if $M$ is
an answer set of $\Pi_n^{Golog}$ then every answer set $Y$ of
$e_U(\Pi_n^{Golog} \setminus \pi_{f}, X)$ must contain
$trans(n_p,0,n)$. Furthermore, we have that $s_t(X) = s_t(M)$ for
every $t$. Hence, in what follows we will use $s_t(X)$ and
$s_t(M)$ interchangeably. We prove the conclusion of the lemma by
proving a stronger conclusion\footnote{
  Recall that for simplicity,
  in encoding programs or formulae we use $l$ or
  $a$ as the name associated to $l$ or $a$, respectively.
}:

\begin{itemize}
\item[(*)] for every ground complex action $q$ with the name $n_q$
and two time points $t_1,t_2$ such that $q \neq \Null$ and
$trans(n_q,t_1,t_2) \in M$, $s_{t_1}(M)a_{t_1}s_{t_1+1}(M)\ldots
a_{t_2-1}s_{t_2}(M)$ is a trace of $q$
(the states $s_i(M)$ and actions $a_i$ are given in the Lemma's
statement).
\end{itemize}

\ni Denote $\pi_3=e_U(\Pi_n^{Golog} \setminus \pi_{f},X)$. We have
that $\pi_3$ consists of the following rules:
\begin{eqnarray}
trans(a,t,t+1)& \la &
    (\mbox{if } action(a) \in X, \; occ(a,t) \in X) \label{pi1tr_act}
\\
trans(f,t_1,t_1)& \la &
     (\mbox{if } formula(f,.) \in X, \; hf(f,t_1) \in X)
\label{pi1tr_form} \\
trans(p,t_1,t_2) & \la &
     t1 \le t' \le t_2,
  trans(p_1,t_1,t'),trans(p_2,t',t_2). \label{pi1tr_proc}\\
&&  (\mbox{if } sequence(p,p_1,p_2) \in X))  \nonumber \\
trans(n,t_1,t_2) & \la &  trans(p_1,t_1,t_2).
      \label{pi1tr_choice}\\
 && (\mbox{if } choiceAction(n) \in X, \; in(p_1,n)\in X) \nonumber \\
trans(i,t_1,t_2)& \la &
    trans(p_1,t_1,t_2). \label{pi1tr_if_true}\\
&& (\mbox{if } if(i,f,p_1,p_2) \in X,\; hf(f,t_1) \in X)   \nonumber \\
trans(i,t_1,t_2)& \la &
    trans(p_2,t_1,t_2).  \label{pi1tr_if_false}\\
&& (\mbox{if }  if(i,f,p_1,p_2) \in X,\; hf(f,t_1) \not\in X) \nonumber
\\
trans(w,t_1,t_2)& \la &
    t_1 < t' \le t_2,
    trans(p,t_1,t'), trans(w,t',t_2). \label{pi1tr_while_true}\\
&& (\mbox{if }  while(w,f,p) \in X, \; hf(f,t_1) \in X) \nonumber \\
trans(w,t,t)& \la &
 (\mbox{if }  while(w,f,p) \in X, hf(f,t) \not\in X)
\label{pi1tr_while_false}\\
trans(s, t_1, t_2)& \la & trans(p, t_1, t_2). \label{pi1tr_pick}\\
&& (\mbox{if }  choiceArgs(s, p) \in X)
 \nonumber \\
trans(\Null,t,t) & \la &   \label{pi1tr_null}
\end{eqnarray}

\ni Clearly, $\pi_3$ is a positive program. Thus, the unique
answer set of $\pi_3$, denoted by $Y$, is the fix-point of the
$T_{\pi_3}$ operator, defined by $T_{\pi_3}(X) = \{o \mid $ there
exists a rule $o \la o_1,\ldots,o_n$ in $\pi_3$ such that $o_i \in
X$ for $i=1,\ldots,n\}$.
Let $Y_k = T_{\pi_3}^ k(\emptyset)$. By definition $Y =
\lim_{n\rightarrow \infty} Y_n$.

\st
For every atom $o \in Y$, let
$\rho(o)$ denote the smallest integer $k$ such that
for all $0 \leq t < k$, $o \not\in Y_t$ and for all $t \geq k$,
$o \in Y_t$. (Notice that the existence
of $\rho(o)$ is guaranteed because $T_{\pi_3}$ is a monotonic,
fix-point operator.)

\st
We prove (*) by induction over $\rho(trans(n_q,t_1,t_2))$.

\st {\bf Base:} $\rho(trans(n_q,t_1,t_2)) = 0$. Then $\pi_3$
contains a rule of the form $trans(n_q,t_1,t_2)\leftarrow$.
Because $q \neq \Null$, we know that
$trans(n_q,t_1,t_2)\leftarrow$ comes from a rule $r$ of the form
(\ref{pi1tr_act}), (\ref{pi1tr_form}), or
(\ref{pi1tr_while_false}).
\begin{itemize}
\item $r$ is of the form (\ref{pi1tr_act}). So, $q$ is some action $a$,
i.e., $action(a)$ and $occ(a,t)$ both belong to $X$. Further, $t_2
= t_1+1$. Because of Corollary \ref{smodel-traject1} we know that
$a$ is executable in $s_{t_1}(X)$ and $s_{t_2}(X) \in
\Phi(a,s_{t_1}(X))$. Since $s_t(M) = s_t(X)$ for every $t$, we
have that $s_{t_1}(M) \; a \; s_{t_2}(M)$ is a trace of $q$.
\item $r$ is of the form (\ref{pi1tr_form}).
Then $q = \phi, t_2 = t_1 = t$, where $\phi$ is a formula and
$hf(n_\phi,t)$ is in $X$. By Theorem \ref{the-lt-sem}, $\phi$
holds in $s_t(X)$. Again, because $s_t(M) = s_t(X)$, we have that
$s_t(M)$ is a trace of $q$.
\item $r$ is of the form (\ref{pi1tr_while_false}).
Then, $t_1 = t_2$, $while(n_q,\phi,p_1) \in X$, and
$hf(n_\phi,t_1) \not\in X$. That is, $q$ is the program ``$\wwhile
\phi \ddo p_1$'' and $\phi$ does not holds in $s_{t_1}(M)$. Thus,
$s_{t_1}(M)$ is a trace of $q$.
\end{itemize}

\ni {\bf Step:} Assume that we have proved (*) for
$\rho(trans(n_q,t_1,t_2)) \leq k$. We need to prove it for the
case $\rho(trans(n_q,t_1,t_2)) = k+1$.

\st Because $trans(n_q,t_1,t_2)$ is in $T_{\pi_3}(Y_k)$, there is
some rule $trans(n_q,t_1,t_2) \leftarrow A_1, \ldots A_m$ in
$\pi_3$ such that all $A_1,\ldots A_m$ are in $Y_k$. From the
construction of $\pi_3$, we have the following cases:
\begin{itemize}
\item $r$ is a rule of the form (\ref{pi1tr_proc}). Then,
there exists $q_1, q_2, t'$ such that
$sequence(n_q,n_{q_1},n_{q_2}) \in X$, $trans(n_{q_1},t_1,t') \in
Y_k$, and $trans(n_{q_2},t',t_2)$.  Hence,
$\rho(trans(n_{q_1},t_1,t'))\leq k$ and
$\rho(trans(n_{q_2},t',t_2))\leq k$. By inductive hypothesis,
$s_{t_1}(M)a_{t_1}s_{t_1+1}(M)\ldots a_{t'-1}s_{t'}(M)$ is a trace
of $q_1$ and $s_{t'}(M)a_{t'}s_{t'+1}(M)\ldots
a_{t_2-1}s_{t_2}(M)$ is a trace of $q_2$. Since
$sequence(n_q,n{q_1},n_{q_2}) \in X$ we know that $q = q_1;q_2$.
By Definition \ref{deftrace1},
$s_{t_1}(M)a_{t_1}s_{t_1+1}(M)\ldots a_{t_2-1}s_{t_2}(M)$ is a
trace of $q$.
\item $r$ is a rule of the form (\ref{pi1tr_choice}).
Then, $choiceAction(n_q)$ is in $X$. So, $q$ is a choice program,
say $q = q_1\mid q_2\ldots \mid q_l$. In addition, there exists $1
\leq j \leq l$ such that $in(n_{q_j},n_q) \in X$ and
$trans(n_{q_j},t_1,t_2) \in Y_k$. By the definition of $\rho$,
$\rho(trans(q_j,t_1,t_2)) \leq k$. By inductive hypothesis,
$s_{t_1}(M)a_{t_1}s_{t_1+1}(M)\ldots a_{t_2-1}s_{t_2}(M)$ is a
trace of $q_j$. By Definition \ref{deftrace1}, it is also a trace
of $q$.
\item $r$ is a rule of the form (\ref{pi1tr_if_true}).
Then, by the construction of $\pi_3$, there exists $\phi$, $q_1$,
$q_2$ such that $if(n_q,n_\phi,n_{q_1},n_{q_2}) \in X$,
$hf(n_\phi,t_1) \in X$, and $trans(n_{q_1},t_1,t_2) \in Y_k$.
Thus $q$ is the program ``$\iif \phi \tthen q_1 \eelse q_2$'' and
$\rho(trans(n_{q_1},t_1,t_2)) \leq k$. Again, by inductive
hypothesis, $s_{t_1}(M)a_{t_1}s_{t_1+1}(M)\ldots
a_{t_2-1}s_{t_2}(M)$ is a trace of $q_1$. Because of Theorem
\ref{the-lt-sem}, $\phi$ holds in $s_{t_1}(M)$. Hence,
$s_{t_1}(M)a_{t_1}s_{t_1+1}(M)\ldots a_{t_2-1}s_{t_2}(M)$ is a
trace of $q$.
\item $r$ is a rule of the form (\ref{pi1tr_if_false}).
Similarly to the previous items, we know that there exist $\phi$,
$q$, $q_1$, and $q_2$ such that $if(n_q,n_\phi,n_{q_1},n_{q_2})
\in X$, $hf(n_\phi,t_1) \not\in X$, and $trans(n_{q_2},t_1,t_2)
\in Y_k$. This means that $\rho(trans(n_{q_2},t_1,t_2)) \leq k$.
Hence, by inductive hypothesis and Theorem \ref{the-lt-sem},
$s_{t_1}(M)a_{t_1}s_{t_1+1}(M)\ldots a_{t_2-1}s_{t_2}(M)$ is a
trace of $q_2$ and $\phi$ is false in $s_{t_1}(M)$, which mean
that $s_{t_1}(M)a_{t_1}s_{t_1+1}(M)\ldots a_{t_2-1}s_{t_2}(M)$ is
a trace of ``$\iif \phi \tthen q_1 \eelse q_2$'', i.e.,  a trace
of $q$.
\item $r$ is a rule of the form (\ref{pi1tr_while_true}). This
implies that there exist a
formula $\phi$, a program $q_1$ and a time point $t' > t_1$
such that $while(n_q,n_\phi,n_{q_1}) \in X$ and $hf(n_\phi,t_1) \in X$,
$trans(n_{q_1},t_1,t')$ and $trans(n_q,t',t_2)$ are in $Y_k$. It
follows that $q$ is the program ``$\wwhile \phi \ddo
q_1$''. Furthermore, $\phi$ holds in $s_{t_1}(M)$, and
$s_{t_1}(M)a_{t_1}s_{t_1+1}(M)\ldots a_{t'-1}s_{t'}(M)$ is a
trace of $q_1$ and $s_{t'}(M)a_{t'}s_{t'+1}(M)\ldots
a_{t_2-1}s_{t_2}(M)$ is a trace of $q$. By Definition \ref{deftrace1},
this implies that
$s_{t_1}(M)a_{t_1}s_{t_1+1}(M)\ldots a_{t_2-1}s_{t_2}(M)$ is a
trace of $q$.
\item $r$ a rule of the form is (\ref{pi1tr_pick}).
Then, $q$ has the form $\pick(X, \{c_1,\ldots,c_n\}, q_1)$.
Therefore, $choiceArgs(n_q, n_{q_1}(c_j))$ is in $X$ for
$j=1,\ldots,n$. $trans(n_q,t_1,t_2) \in Y$ implies that there
exists an integer $j$, $1 \le j \le n$, such
$trans(n_{q_1(c_j)},t_1,t_2) \in Y_k$. By the definition of
$\rho$, $\rho(trans(n_{q_1}(c_j),t_1,t_2)) \leq k$. By inductive
hypothesis, $s_{t_1}(M)a_{t_1}s_{t_1+1}(M)\ldots
a_{t_2-1}s_{t_2}(M)$ is a trace of program $q_1(c_j)$. Thus, we
can conclude that $s_{t_1}(M)a_{t_1}s_{t_1+1}(M)\ldots
a_{t_2-1}s_{t_2}(M)$ is a trace of $q$.
\end{itemize}

\ni The above cases prove the inductive step for (*). The lemma
follows immediately since $trans(n_p,0,n)$ belongs to $M$.
\qed

\st To prove the reverse of Lemma \ref{lemtrace1}, we define a
function $\mu$ that maps each ground complex action $q$ into an
integer $\mu(q)$ that reflects the notion of complexity of $q$ (or
the number of nested constructs in $q$). $\mu(q)$ is defined
recursively over the construction of $q$ as follows.

\begin{itemize}
  \item For $q = \phi$ and $\phi$ is a formula, or
$q =  a$ and $a$ is an action, $\mu(q) = 0$.
  \item For $q = q_1;q_2$ or $q = \iif \phi \tthen q_1 \eelse q_2$,
$\mu(q) = 1 + \mu(q_1) + \mu(q_2)$.
  \item For $q = q_1\mid \ldots \mid q_m$,
$\mu(q) = 1+ \max\{\mu(q_i) \mid i=1,\ldots,m\}$.
  \item For $q = \wwhile \phi \ddo q_1$, $ \mu(q) = 1 + \mu(q_1)$.
  \item For $q = \pick(X,\{c_1,\ldots,c_n\},q_1)$,
$\mu(q) = 1 + \max\{\mu(q_1(c_j)) \mid j=1,\ldots,n\}$.

\item For $q = p(c_1,\ldots,c_n)$ where
$(p(X_1,\ldots,X_n): \delta_1)$ is a procedure, $\mu(q) = 1 +
\mu(\delta_1(c_1,\ldots,c_n))$.
\end{itemize}

\ni
It is worth noting that $\mu(q)$ is always defined for
programs considered in this paper.

\begin{lemma}\label{lemtrace2}
Let $(D,\Gamma)$ be a consistent action theory, $p$ be a program,
and $s_0a_0\ldots s_{n-1}a_n$ be a trace of $p$. Then $\Pi_n^{Golog}$
has an answer set $M$ such that
\begin{itemize}
\item $occ(a_i,t) \in M$ for $0 \le i \le n-1$,
\item $s_t = s_t(M)$ for every $0 \leq t \leq n$, and
\item $trans(n_p,0,n) \in M$.
\end{itemize}
\end{lemma}
\proof We prove the lemma by constructing an answer set of
$\Pi_n^{Golog}$ that satisfies the conditions of the lemma.
Similar to the proof of Lemma \ref{lemtrace1}, we split
$\Pi_n^{Golog}$ using $U = lit(\pi_{f}) \cup r(p)$. This implies
that $M$ is an answer set of $\Pi_n^{Golog}$ iff $M = X \cup Y$
where $X$ is an answer set of $b_U(\Pi_n^{Golog})$ and $Y$ is an
answer set of $\pi_3 = e_U(\Pi_n^{Golog} \setminus
b_U(\Pi_n^{Golog}), X)$, which is the program consisting of the
rules (\ref{pi1tr_act})-(\ref{pi1tr_null}) with the corresponding
conditions.

\st Because $s_0a_0\ldots a_{n-1}s_n$ is a trace of $p$, it is a
trajectory in $D$. By Corollary \ref{smodel-traject2}, we know
that $\pi_{f}$ has an answer set $X'$ that satisfies the two
conditions:
\begin{itemize}
\item $occ(a_i,t) \in X'$ for $0 \le i \le n-1$ and
\item $s_t = s_t(X')$ for every $0 \leq t \leq n$.
\end{itemize}

\ni Because $r(p)$ consists of only rules and atoms encoding the
program $p$, it is easy to see that there exists an answer set $X$
of $\pi_{f} \cup r(p)$ such that $X' \subseteq X$. Clearly, $X$ also
satisfies the two conditions:
\begin{itemize}
\item $occ(a_i,t) \in X$ for $0 \le i \le n-1$ and
\item $s_t = s_t(X)$ for every $0 \leq t \leq n$.
\end{itemize}

\ni Since $\pi_3$ is a positive program we know that $\pi_3$ has a
unique  answer set, say $Y$. From the splitting theorem, we have
that $M = X \cup Y$ is an answer set of $\Pi_n^{Golog}$. Because
$s_t(X) = s_t(M)$, $M$ satisfies the first two conditions of the
lemma. It remains to be shown that $M$ also satisfies the third
condition of the lemma. We prove this by proving a stronger
conclusion:
\begin{itemize}
  \item[(*)] If $q$ is a program with the name $n_q$
  and there exists two integers $t_1$ and $t_2$ such that
  $s_{t_1}(M)a_{t_1}\ldots a_{t_2-1}s_{t_2}(M)$ is a trace of
  $q$ then $trans(n_q,t_1,t_2) \in M$.
(the states $s_i(M) = s_i$ -- see above -- and the actions $a_i$
are defined as in the Lemma's statement)
\end{itemize}

\ni
We prove (*) by induction over $\mu(q)$, the complexity of the program
$q$.

\st {\bf Base:} $\mu(q) = 0$. There are only two cases:
\begin{itemize}
\item $q = \phi$ for some formula $\phi$, and
hence, by Definition \ref{deftrace1}, we have that $t_2 = t_1$. It
follows from the assumption that $s_{t_1}(M)$ is a trace of $q$
that $s_{t_1}(M)$ satisfies $\phi$. By Theorem \ref{the-lt-sem},
$hf(n_\phi, t_1) \in X$, and hence, we have that
$trans(n_\phi,t_1,t_1) \in Y$ (because of rule
(\ref{pi1tr_form})).

\item $q = a$ where $a$ is an action. Again,
by Definition \ref{deftrace1}, we have that $t_2 = t_1+1$. From
the assumption that $s_{t_1}(M) a_{t_1} s_{t_2}(M)$ is a trace of
$q$ we have that $a_{t_1} = a$. Thus, $occ(a,t_1) \in M$. By rule
(\ref{pi1tr_act}) of $\pi_3$, we conclude that $trans(a,t_1,t_2)
\in Y$, and thus, $trans(a,t_1,t_2) \in M$.
\end{itemize}

\ni The above two cases prove the base case.

\st {\bf Step:} Assume that we have proved (*) for every program
$q$ with $\mu(q) \le k$. We need to prove it for the case ${\mu(q)
= k+1}$. Because $\mu(q) > 0$, we have the following cases:

\begin{itemize}
\item  $q = q_1;q_2$. By Definition \ref{deftrace1},
there exists $t'$, $ t_1 \le t' \le t_2$, such that
$s_{t_1}a_{t_1}\ldots s_{t'}$ is a trace of $q_1$ and
$s_{t'}a_{t'}\ldots s_{t_2}$ is a trace of $q_2$. Because
$\mu(q_1) < \mu(q)$ and $\mu(q_2) < \mu(q)$, by inductive
hypothesis, we have that $trans(n_{q_1},t_1,t')\in M$ and
$trans(n_{q_2},t',t_2)\in M$. $q = q_1;q_2$ implies
$sequence(n_q,n_{q_1},n_{q_2}) \in M$. By rule (\ref{pi1tr_proc}),
$trans(n_q,t_1,t_2)$ must be in $M$.
\item $q = q_1 \mid \ldots \mid q_i$. Again, by Definition \ref{deftrace1},
$s_{t_1}a_{t_1}\ldots a_{t_2-1}s_{t_2}$ is a trace of some $q_j$. Since
$\mu(q_j) < \mu(q)$, by inductive hypothesis, we have that
$trans(n_{q_j},t_1,t_2) \in M$.
Because of rule (\ref{pi1tr_choice}), $trans(n_q,t_1,t_2)$ is in $M$.
\item $q = \iif \phi \tthen q_1 \eelse q_2$. Consider two cases:
\begin{itemize}

\item $\phi$ holds in $s_{t_1}$.
This implies that $s_{t_1}(M)a_{t_1}\ldots a_{t_2-1}s_{t_2}(M)$
is a trace of $q_1$.
Because of Theorem \ref{the-lt-sem}, $hf(n_\phi,t_1) \in M$.
Since $\mu(q_1) < \mu(q)$, $trans(n_{q_1},t_1,t_2) \in M$
by inductive hypothesis. Thus, according to
rule (\ref{pi1tr_if_true}), $trans(n_q,t_1,t_2)$ must belong to $M$.
\item $\phi$ does not holds in $s_{t_1}$.
This implies that $s_{t_1}(M)a_{t_1}\ldots a_{t_2-1}s_{t_2}(M)$
is a trace of $q_2$.
Because of Theorem \ref{the-lt-sem}, $hf(n_\phi,t_1)$ does not
hold in $M$. Since
$\mu(q_1) < \mu(q)$, $trans(n_{q_2},t_1,t_2)$ is in $M$
by inductive hypothesis. Thus, according to
rule (\ref{pi1tr_if_false}), $trans(n_q,t_1,t_2) \in M$.
\end{itemize}
\item $q= \wwhile \phi \ddo q_1$. We prove this case by induction over
the length of the trace, $t_2 - t_1$.

\begin{itemize}
\item {\bf Base:} $t_2 - t_1 = 0$. This happens only when $\phi$
does not hold in $s_{t_1}(M)$. As such,
because of rule (\ref{pi1tr_while_false}),
$trans(n_q,t_1,t_2)$ is in $M$. The base case is proved.

\item {\bf Step:} Assume that we have proved the conclusion for
this case for $0 \le t_2 - t_1 < l$. We will show that it is also
correct for $t_2 - t_1 = l$. Since $t_2 - t_1 > 0$, we conclude
that $\phi$ holds in $s_{t_1}$ and there exists $t_1 < t' \le t_2$
such that $s_{t_1}a_{t_1}\ldots s_{t'}$ is a trace of $q_1$ and
$s_{t'}a_{t'}\ldots s_{t_2}$ is a trace of $q$. We have $\mu(q_1)
< \mu(q)$, $t' - t_1 \le t_2 - t_1$ and $t_2 - t' < t_2 - t_1=l$.
By inductive hypothesis, $trans(n_{q_1}, t_1, t')$ and
$trans(n_q,t',t_2)$ are in $M$. By Theorem \ref{the-lt-sem},
$hf(n_\phi,t_1)$ is in $M$ and from the rule
(\ref{pi1tr_while_true}), $trans(n_q,t_1,t_2)$ is in $M$.
\end{itemize}
\item $q = \pick(X, \{c_1,\ldots,c_n\}, q_1)$. So,
there exists an integer $j$, $1 \le j \le n$,
such that the trace of $q$ is a trace of $q_1(c_j)$.
Since $\mu(q_1(c_j)) < \mu(q)$, we have that
$trans(n_{q_1(c_j)}, t_1, t_2) \in M$. This, together with the
fact that $choiceArgs(n_q,n_{q_1(c_j)}) \in r(p)$, and the
rule (\ref{pi1tr_pick}) imply that $trans(n_q,t_1,t_2)$ is in $M$.

\item $q = p(c_1,\ldots,c_n)$ for some procedure
$(p(X_1,\ldots,X_n), q_1)$. This implies that
$s_{t_1}(M)a_{t_1}\ldots a_{t_2-1}s_{t_2}(M)$ is a trace of
$q_1(c_1,\ldots,c_n)$. Since $\mu(q_1(c_1,\ldots,c_n)) < \mu(q)$,
we have that $trans(n_{q_1}, t_1, t_2) \in M$. Since $n_{q_1} =
p(c_1,\ldots,c_n)$ and $n_q = q$, we have that $trans(n_{q}, t_1,
t_2) \in M$. This proves the inductive step for this case as well.
\end{itemize}

\ni
The above cases prove the inductive step of (*).
The conclusion of the lemma follows.
\qed

\st We now prove the Theorem \ref{th2}.

\setcounter{theorem}{5}
\begin{theorem}
{\rm
Let $(D,\Gamma)$ be a consistent action theory and $p$ be a
program. Then,
\begin{itemize}
\item[(i)] for every  answer set $M$ of $\;\Pi^{Golog}_n$ with
$occ(a_i,i) \in M$ for $i \in \{0,\ldots,n-1\}$, $\,s_0(M)a_0
\ldots a_{n-1} s_n(M)\,$ is a trace of $p$; and

\item[(ii)] if $\,s_0a_0\ldots a_{n-1}s_n\,$ is a trace
of $p$ then there exists an answer set $M$ of $\;\Pi^{Golog}_n$
such that $s_j = s_j(M)$ and $\;occ(a_i,i) \in M$ for $j \in
\{0,\ldots,n\}$ and $\,i \in \{0,\ldots,n-1\}$.
\end{itemize}
}
\end{theorem}
\ni \proof (i) follows from Lemma \ref{lemtrace1} and (ii) follows
from Lemma \ref{lemtrace2}. \qed

\section*{Appendix A.4 - Proof of Theorem \ref{th3}}

Let $p$ now be a general program. To prove Theorem \ref{th3}, we
will extend the Lemmas \ref{lemtrace1}-\ref{lemtrace2} to account
for general programs. Similarly to the proofs of Lemmas
\ref{lemtrace1}-\ref{lemtrace2}, we will split $\Pi^{HTN}_n$ by
the set $U = lit(\pi_{f}) \cup r(p)$. Thus $M$ is an answer set of
$\Pi^{HTN}_n$ iff $M = X \cup Y$ where $X$ is an answer set of
$\pi_{f} \cup r(p)$ and $Y$ is an answer set of the program
$e_U(\Pi^{HTN}_n \setminus b_U(\Pi^{HTN}_n), X)$, denoted by
$\pi_4$, which consists of the rules of program $\pi_3$ (with the
difference that a program is now a general program) and the
following rules:

\begin{eqnarray}
trans(n,t_1,t_2)& \la &
     \naf nok(n,t_1,t_2). \label{pi1thtn1}\\
 && (\mbox{if } htn(n,s,c) \in X) \nonumber \\
1\{begin(n, i, t_3, t_1, t_2)\; : \; \hspace*{0.2in}& & \nonumber \\
between(t_3,t_1,t_2)\}1 &\la  &
trans(n,t_1,t_2). \label{pi1thtn2}\\
&& (\mbox{if } htn(n,s,c) \in X, \; in(i, s) \in X) \nonumber \\
1\{end(n, i, t_3, t_1, t_2)\; : \; \hspace*{0.2in} & & \nonumber \\
between(t_3,t_1,t_2)\}1   &\la  &
    trans(n,t_1,t_2). \label{pi1thtn3} \\
&& (\mbox{if } htn(n,s,c) \in X, \; in(i, s) \in X) \nonumber \\
used(n,t,t_1,t_2)& \la &
    begin(n, i, b, t_1, t_2), \label{pi1thtn11} \\
&&  end(n, i, e, t_1, t_2), \nonumber \\
&&  b \le t \le e.  \nonumber \\
&& (\mbox{if } htn(n,s,c) \in X, \; in(i, s) \in X) \nonumber \\
not\_used(n,t,t_1,t_2)& \la & \naf used(n,t,t_1,t_2).
\label{pi1thtn12} \\
overlap(n, t, t_1, t_2) & \la &
    begin(n, i_1, b_1, t_1, t_2), \label{pi1thtn14} \\
&&  end(n, i_1, e_1, t_1, t_2), \nonumber \\
&&   begin(n, i_2, b_2, t_1, t_2), \nonumber \\
&&  end(n, i_2, e_2, t_1, t_2), \nonumber \\
&&   b_1 <  t \le e_1,  b_2 < t \le e_2.  \nonumber \\
&& (\mbox{if } htn(n,s,c) \in X, \; in(i_1, s) \in X, \; in(i_2,s)\in X)
\nonumber \\
nok(n,t_1,t_2) & \la &
     t_3 > t_4,
     begin(n, i, t_3, t_1, t_2), \label{pi1thtn4}\\
&&   end(n, i, t_4, t_1, t_2). \nonumber \\
&& (\mbox{if } htn(n,s,c) \in X, \; in(i, s) \in X) \nonumber \\
nok(n,t_1,t_2)& \la &
     t_3 \leq t_4,
     begin(n, i, t_3, t_1, t_2), \label{pi1thtn5}\\
&&   end(n, i, t_4, t_1, t_2), \nonumber \\
&&      \naf trans(i,t_3,t_4). \nonumber \\
&& (\mbox{if } htn(n,s,c) \in X, \; in(i, s) \in X) \nonumber \\
nok(n,t_1,t_2) & \la &
    t_1 \le t \le t_2, not\_used(n,t, t_1,t_2).
    \label{pi1thtn13} \\
&& (\mbox{if } htn(n,s,c) \in X) \nonumber \\
nok(n,t_1,t_2) & \la &
    t_1 \le t \le t_2, overlap(n,t, t_1,t_2).
    \label{pi1thtn15} \\
&& (\mbox{if } htn(n,s,c) \in X) \nonumber \\
nok(n,t_1,t_2) & \la &
    begin(n, i_1, b_1, t_1, t_2),
\label{pi1thtn7}\\
 && begin(n, i_2, b_2, t_1, t_2),
\nonumber \\
 &&  b_1 > b_2.\nonumber \\
&& (\mbox{if } htn(n,s,c) \in X, \; in(i_1, s) \in X, \;
   in(i_2, s)\in X, \nonumber \\
&& in(o, c)\in X, \; order(o, i_1, i_2)\in X)
) \nonumber \\
nok(n,t_1,t_2) & \la &
        end(n, i_1, e_1, t_1, t_2),\label{pi1thtn8} \\
 &&     begin(n, i_2, b_2, t_1, t_2),
        e_1 < t_3 < b_2. \nonumber \\
&& (\mbox{if } htn(n,s,c) \in X, \; in(i_1, s) \in X,\;
in(i_2, s)\in X, \nonumber \\
&& in(o, c)\; \in X, \; maintain(o, f, i_1, i_2)\in X, \nonumber \\
 && \mbox{and } hf(f,t_3)\not\in X) \nonumber \\
nok(n,t_1,t_2) & \la &
        begin(n, i, b, t_1, t_2),
        end(n, i, e, t_1, t_2),\label{pi1thtn9}\\
&& (\mbox{if } htn(n,s,c) \in X, \; in(i, s) \in X,\nonumber \\
&& in(o,c)\in X, precondition(o, f, i) \in X \nonumber \\
&&  \mbox{and } hf(f, b) \not\in X ) \nonumber \\
nok(n,t_1,t_2) & \la &
        begin(n, i, b, t_1, t_2),
        end(n, i, e, t_1, t_2). \label{pi1thtn10} \\
&& (\mbox{if } htn(n,s,c) \in X, \; in(i, s) \in X,\nonumber \\
&& in(o, c) \in X, \; postcondition(o, f, i) \in X, \nonumber \\
&&  \mbox{and } hf(f, e) \not\in X
        ) \nonumber
\end{eqnarray}

\ni We will continue to use the complexity of
program defined in the last appendix and extend it to allow the
HTN-construct by adding the following to the definition of
$\mu(q)$.
\begin{itemize}
\item For $q = (S,C)$, $\mu(q) = 1+\Sigma_{p \in S} \mu(p)$.

\end{itemize}

\ni Notice that every literal of the program $\pi_4$ has the first
parameter as a program\footnote{ More precisely,  a program
name.}. Hence, we can associate $\mu(q)$ to each literal $l$ of
$\pi_4$ where $q$ is the first parameter of $l$. For instance,
$\mu(trans(q,t_1,t_2)) = \mu(q)$ or $\mu(nok(q,t_1,t_2)) = \mu(q)$
etc.. Since we will continue to use splitting theorem in our
proofs, the following observation is useful.

\begin{observation}
\label{obs2} The two cardinality constraint rules
(\ref{pi1thtn2})-(\ref{pi1thtn3}) can be replaced by the following
normal logic program rules:

\begin{eqnarray*}
begin(n,i,t,t_1,t_2) & \la & trans(n,t_1,t_2),\\
& & t_1 \le t \le t_3 \le t_2, \naf nbegin(n,i,t,t_1,t_2). \\
nbegin(n,i,t,t_1,t_2) & \la & trans(n,t_1,t_2),\\
& &  t_1 \le t \le t_2, t_1 \le  t_3 \le t_2,t \ne t_3,
begin(n,i,t_3,t_1,t_2). \\
end(n,i,t,t_1,t_2) & \la & trans(n,t_1,t_2),\\
& & t_1 \le t \le t_3 \le t_2, \naf nend(n,i,t,t_1,t_2). \\
nend(n,i,t,t_1,t_2) & \la & trans(n,t_1,t_2),\\
& &  t_1 \le t \le t_2, t_1 \le  t_3 \le t_2,t \ne t_3,
end(n,i,t_3,t_1,t_2).
\end{eqnarray*}
That is, let $\pi^*$ be the program obtained from $\pi_4$ by
replacing the rules (\ref{pi1thtn2})-(\ref{pi1thtn3}) with the
above set of rules. Then, $M$ is an answer set of $\pi_4$ iff $M'
= M \cup \{nbegin(n,i,t_3,t_1,t_2) \mid begin(n,i,t,t_1,t_2) \in
M, t \ne t_3, t_1 \le t,t_3 \le t_2\} \cup \{nend(n,i,t_3,t_1,t_2)
\mid t \ne t_3, \; t_1 \le t,t_3 \le t_2, end(n,i,t,t_1,t_2) \in
M\}$ is an answer set of $\pi^*$.
\end{observation}

\ni
The next lemma generalizes Lemma \ref{lemtrace1}.

\begin{lemma}
\label{lemtrace3} Let $q$ be a general program, $Y$ be an answer
set of the program $e_U(\Pi^{HTN}_n \setminus b_U(\Pi^{HTN}_n),
X)$ (i.e. program $\pi_4$), and $t_1,t_2$ be two time points such
that $q \neq \Null$\ and $trans(n_q,t_1,t_2) \in Y$. Then,
$s_{t_1}(M)a_{t_1}s_{t_1+1}(M)\ldots a_{t_2-1}s_{t_2}(M)$ is a
trace of $q$ where $M = X \cup Y$.
\end{lemma}
\proof Let $U_k = \{l \mid l \in lit(\pi_4), \; \mu(l) \le k\}$.

\st From observation \ref{obs2}, we know that we can use the
splitting theorem on $\pi_4$. It is easy to see that $\lan U_k
\ran_{k < \infty}$ is a splitting sequence of $\pi_4$. From the
finiteness of $\pi_4$ and the splitting sequence theorem, we have
that $Y = \bigcup_{i < \infty} Y_i$ where

\begin{enumerate}

\item $Y_0$ is an answer set  of the program $b_{U_0}(\pi_4)$ and

\item for every integer $i$, $Y_{i+1}$ is an answer set for
$e_{U_i}(b_{U_{i+1}}(\pi_4) \setminus b_{U_i}(\pi_4), \bigcup_{j \le i}
Y_j)$.
\end{enumerate}

\ni We prove the lemma by induction over $\mu(q)$.

\st {\bf Base:} $\mu(q) = 0$. From $trans(n_q,t_1,t_2) \in Y$, we
have that $trans(n_q,t_1,t_2) \in Y_0$. It is easy to see that
$b_{U_0}(\pi_4)$ consists of all the rules of $\pi_4$ whose
program has level $0$. It follows from Lemma \ref{lemtrace1}
$s_{t_1}(M)a_{t_1}s_{t_1+1}(M)\ldots a_{t_2-1}s_{t_2}(M)$ is a
trace of $q$. The base case is proved.

\st {\bf Step:} Assume that we have proved the lemma for $\mu(q) =
k$. We prove it for $\mu(q) = k+1$. From the fact that
$trans(n_q,t_1,t_2) \in M$ and $\mu(n_q) = k+1$ we have that
$trans(n_q,t_1,t_2) \in Y_{k+1}$ where $Y_{k+1}$ is an answer set
of the program $e_{U_k}(b_{U_{k+1}}(\pi_4) \setminus
b_{U_k}(\pi_4), \bigcup_{j \le k} Y_k)$ which consists of rules of
the form (\ref{pi1thtn1})-(\ref{pi1thtn10}) and
(\ref{pi1tr_proc})-(\ref{pi1tr_pick}) whose program has the level
$k+1$, i.e., $\mu(q) = k+1$. Because $trans(n_q,t_1,t_2) \in Y$ we
know that there exists a rule that supports $trans(n_q,t_1,t_2)$.
Let $r$ be such a rule. There are following cases:

\begin{itemize}

\item $r$ is a rule of the form (\ref{pi1tr_proc})-(\ref{pi1tr_pick}),
the argument is similar to the argument using in the inductive
step for the corresponding case in Lemma \ref{lemtrace1}. Notice a
minor difference though: in Lemma \ref{lemtrace1}, we do not need
to use $\mu(q)$.

\item $r$ is a rule of the form (\ref{pi1thtn1}),
which implies that $q = (S,C)$ where $S$ is a set of programs and
$C$ is a set of constraints $C$. By definition of answer sets, we
know that $nok(n_q, t_1, t_2) \not\in Y_{k+1}$. Furthermore,
because of the rules (\ref{pi1thtn2}) and (\ref{pi1thtn3}), the
fact that $trans(n_q,t_1,t_2) \in Y_{k+1}$ and the definition of
weight constraint rule, we conclude that for each $q_j \in S$
there exists two numbers $j_b$ and $j_e$, $t_1 \le j_b,j_e \le
t_2$ such that $begin(n_q,n_{q_j},j_b,t_1,t_2) \in Y_{k+1}$ and
$end(n_q,n_{q_j},j_e,t_1,t_2) \in Y_{k+1}$. Because of rule
(\ref{pi1thtn5}), we conclude that $trans(n_{q_j},j_b,j_e) \in
\bigcup_{i \le k}Y_i$. Otherwise, we have that $nok(n_q,t_1,t_2)
\in Y_{k+1}$, and hence, $trans(n_q,t_1,t_2) \not\in Y_{k+1}$,
which is a contradiction. By definition of $\mu(q)$, we have that
$\mu(q_j) < \mu(q)$. Thus, by inductive hypothesis, we can
conclude that:  for every $q_j \in S$, there exists two numbers
$j_b$ and $j_e$, $t_1 \le j_b,j_e \le t_2$, $s_{j_b}(M)a_{j_b}
\ldots a_{j_e-1}s_{j_e}(M)$ is a trace of $q_j$.

\st Furthermore, rules (\ref{pi1thtn11})-(\ref{pi1thtn15}) imply
that the set $\{j_b \mid q_j \in S\}$ creates a permutation of
$\{1,\ldots,|S|\}$ that satisfies the first condition of
Definition \ref{deftrace2}.

\st Consider now an ordering $q_{j_1} \prec q_{j_2}$ in $C$. This
implies that the body of rule (\ref{pi1thtn7}) will be satisfied if
$j_{b_1} > j_{b_2}$ which would lead to $trans(n_q,t_1,t_2)
\not\in Y_{k+1}$. Again, this is a contradiction. Hence, we must
have $j_{b_1} \le j_{b_2}$ that means that the permutation $\{j_b
\mid q_j \in S\}$ also satisfies the second condition of
Definition \ref{deftrace2}.

\st Similarly, using (\ref{pi1thtn8})-(\ref{pi1thtn10}) we can
prove that the permutation $\{j_b \mid q_j \in S\}$ also
satisfies the third and fourth conditions of Definition
\ref{deftrace2}.

\st It follows from the above arguments that $s_{t_1}(M) a_{t_1} \ldots
a_{t_2-1} s_{t_2}(M)$ is a trace of $q)$. The inductive step is
proved for this case.
\end{itemize}

\ni The above cases prove the inductive step. This concludes the
lemma. \qed

\st In the next lemma, we generalize the Lemma \ref{lemtrace2}.

\begin{lemma}
\label{lemtrace4} Let $(D,\Gamma)$ be a consistent action theory,
$p$ be a general program, and $s_0 a_0 \ldots a_{n-1}s_n$ be a
trace of $p$. Then, there is an answer set $M$ of $\Pi^{HTN}_n$
such that $s_i(M) = s_i$ and $occ(a_i,i) \in M$ and
$trans(n_p,0,n) \in M$.
\end{lemma}
\proof Based on our discussion on splitting $\Pi^{HTN}_n$ using
$lit(\pi_f) \cup r(q)$ and the fact that $s_0 a_0 \ldots
a_{n-1}s_n$ is also a trace in $D$, we know that there exists an
answer set $X$ of $\pi \cup r(p)$ such that $s_i(X) = s_i$ and
$occ(a_i,i) \in X$. Thus, it remains to be shown that there exists
an answer set $Y$ of $\pi_4$ such that $trans(n_p,0,n) \in Y$.
Similar to the proof of Lemma \ref{lemtrace3}, we use  $\lan U_k
\ran_{k < \infty}$ as a splitting sequence of $\pi_4$ where $U_k =
\{u \mid u \in lit(\pi_4), \; \mu(u) \le k\}$. From the splitting
sequence theorem, we have that $Y = \bigcup_{ i < \infty} Y_i$
where

\begin{enumerate}

\item $Y_0$ is an answer set  of the program $b_{U_0}(\pi_4)$ and

\item for every integer $i$, $Y_{i+1}$ is an answer set for
$e_{U_i}(b_{U_{i+1}}(\pi_4) \setminus b_{U_i}(\pi_4), \bigcup_{j
\le i} Y_j)$.
\end{enumerate}

\ni We prove the lemma by induction over $\mu(q)$. Similar to Lemma
\ref{lemtrace2}, we prove this by proving a stronger conclusion:

\begin{itemize}
\item[(*)] There exists an answer set $Y = \bigcup_{ i < \infty} Y_i$
of $\pi_4$ such that for every program $q \not = \mathbf{null}$
occurring in $p$, $s_{t_1} a_{t_1} \ldots a_{t_2-1}s_{t_2}$ is a
trace of $q$ iff $trans(n_q,t_1,t_2) \in Y_{\mu(q)}$. (the states
$s_i$ and the actions $a_i$ are defined as in the Lemma's
statement)
\end{itemize}

\ni We will prove (*) by induction over $\mu(q)$.

\st {\bf Base:} $\mu(q) = 0$. Similar to the base case in Lemma
\ref{lemtrace2} .

\st {\bf Step:} Assume that we have proved (*) for $\mu(q) \le k$.
We need to prove (*) for $\mu(q) = k+1$. We will construct an
answer set of $\pi^+ = e_{U_k}(b_{U_{k+1}}(\pi_4) \setminus
b_{U_k}(\pi_4), \bigcup_{j \le k} Y_k)$ such that for every
program $q$ occurring in $p$ with $\mu(q) = k+1$,  if $s_{t_1}
a_{t_1} \ldots a_{t_2-1}s_{t_2}$ is a trace of $q$ then
$trans(n_q,t_1,t_2) \in Y_{k+1}$.

\st Let $Y_{k+1}$ be the set of atoms defined as follows.

\begin{itemize}

\item For every program $q$ with $\mu(q) = k+1$,
if $q$ is not of the form $(S,C)$ and
$s_{t_1} a_{t_1} \ldots a_{t_2-1}s_{t_2}$ is a trace
of $q$, $Y_{k+1}$ contains $trans(n_q,t_1,t_2)$.

\item For every program $q$ with $\mu(q) = k+1$,
$q = (S,C)$, and $s_{t_1} a_{t_1} \ldots a_{t_2-1}s_{t_2}$ is a
trace of $q$. By definition, there exists a permutation
$\{j_1,\ldots,j_{|S|}\}$ of $\{1,\ldots,|S|\}$ satisfying the
conditions (a)-(d) of Item 8 (Definition \ref{deftrace2}).
Consider such a permutation. To simplify the notation, let us
denote the begin- and end-time of a program $q_j \in S$ in the
trace of $q$ by $b_j$ and $e_j$, respectively, i.e.,
$s_{b_j}a_{b_j} \ldots s_{e_j}$ is a trace of $q_j$. Then,
$Y_{k+1}$ contains  $trans(n_q,t_1,t_2)$ and the following atoms:

\begin{enumerate}

\item $begin(n_q,n_{q_j},b_{j},t_1,t_2)$ for every $q_j \in S$,

\item $end(n_q,n_{q_j},e_{j},t_1,t_2)$ for every $q_j \in S$, and

\item $used(n_q,t,t_1,t_2)$ for for every $q_j \in S$
and $b_j \le t \le e_j$ .
\end{enumerate}

\item $Y_{k+1}$ does not contain any other atoms except those
mentioned above.
\end{itemize}

\ni It is easy to see that $Y_{k+1}$ satisfies (*) for every
program $q$ with $\mu(q) = k+1$. Thus, we need to show that
$Y_{k+1}$ is indeed an answer set of $\pi^+$. First, we prove that
$Y_{k+1}$ is closed under $(\pi^+)^{Y_{k+1}}$. We consider the
following cases:

\begin{itemize}

\item $r$ is a rule of the form (\ref{pi1tr_proc}). Obviously, if
$r$ belongs to $(\pi^+)^{Y_{k+1}}$, then $q = q_1;q_2$ and there
exists a $t_1 \le t' \le t_2$ such that $trans(n_{q_1},t_1,t')$
and $trans(n_{q_2},t',t_2)$ belong to $\bigcup_{j \le k} Y_k$
because $\mu(q_1) < \mu(q)$ and $\mu(q_2) < \mu(q)$. By inductive
hypothesis, $s_{t_1} a_{t_1} \ldots s_{t'}$ is a trace of $q_1$
and $s_{t'} a_{t'} \ldots s_{t_2}$ is a trace of $q_2$. By
Definition \ref{deftrace1},
 $s_{t_1} a_{t_1} \ldots s_{t_2}$ is a trace of $q$. By construction
of $Y_{k+1}$ we have that $trans(n_q,t_1,t_2) \in Y_{k+1}$. This
shows that $Y_{k+1}$ is closed under $r$. Similar arguments
conclude that $Y_{k+1}$ is closed under the rule of the form
(\ref{pi1tr_choice})-(\ref{pi1tr_pick}).

\item $r$ is a rule of the form (\ref{pi1thtn1}) of
$(\pi^+)^{Y_{k+1}}$. Then, $q = (S,C)$ and by construction of
$Y_{k+1}$, if $s_{t_1} a_{t_1} \ldots s_{t_2}$ is a trace of $q$
then we have $trans(n_q,t_1,t_2) \in Y_{k+1}$. Thus. $Y_{k+1}$ is
closed under the rules of the form (\ref{pi1thtn1}) too.

\item $r$ is a rule of the form
(\ref{pi1thtn2}) and  (\ref{pi1thtn3}). $Y_{k+1}$ is also closed
under $r$ because whenever $trans(n_q,t_1,t_2) \in Y_{k+1}$, we
now that there is a trace $s_{t_1} a_{t_1} \ldots s_{t_2}$ of $q$,
and hence, by Definition \ref{deftrace2}, we conclude the
existence of the begin- and end-time points $b_j$ and $e_j$ of
$q_j$, respectively. By construction of $Y_{k+1}$, we have that
$begin(n_q,n_{q_j},b_j,t_1,t_2)$ and
$end(n_q,n_{q_j},e_j,t_1,t_2)$ belong to $Y_{k+1}$ and for each
$q_j$, there is a unique atom of this form in $Y_{k+1}$. Hence,
$Y_{k+1}$ is closed under rules of the form (\ref{pi1thtn2}) and
(\ref{pi1thtn3}).

\item $r$ is a rule of the form (\ref{pi1thtn12})-(\ref{pi1thtn10}).
The construction of $Y_{k+1}$ ensures that the body of $r$ is not
satisfied by
$Y_{k+1}$, and hence, $Y_{k+1}$ is closed under $r$.

\item $r$ is a rule of the form (\ref{pi1thtn11}). Because
$used(n_q,t,t_1,t_2)$ belongs to $Y_{k+1}$ for every $t$, $t_1 \le t \le
t_2$.
we have that $Y_{k+1}$ is closed under $r$ too.

\end{itemize}

\ni The conclusion that $Y_{k+1}$ is closed under  $(\pi^+)^{Y_{k+1}}$
follows from the above cases.

\st To complete the proof, we need to show that $Y_{k+1}$ is
minimal. Assume the contrary, there exists a proper subset $Y'$ of
$Y_{k+1}$ such that $Y'$ is closed under $(\pi^+)^{Y_{k+1}}$. Let
$u \in Y_{k+1}\setminus Y'$. Since $u \in Y_{k+1}$, we have the
following cases:

\begin{itemize}

\item $u$ is the head of a rule of the form
(\ref{pi1tr_proc})-(\ref{pi1tr_pick}). By definition of $\pi^+$, we know
that
a rule of this form belongs
to $\pi^+$ iff its body is empty. Thus, from the closeness of $Y'$
we have that $u \in Y'$. This contradicts the fact that $u \not\in Y'$.

\item $u$ is the head of a rule of the form (\ref{pi1thtn1}). Similar
to the above case, we can conclude that $u \in Y'$ which again
contradicts the fact that $u \not\in Y'$.

\item $u$ is the head of a rule $r$ of the form (\ref{pi1thtn2}).
Because of $u \in Y_{k+1}$ we conclude that $trans(n_q,t_1,t_2)
\in Y_{k+1}$. The above case concludes that $trans(n_q,t_1,t_2)
\in Y'$. Since the body of $r$ is true, we conclude that there
exists some $q_j \in S$ such that $Y'$ does not contain an atom
of the form $begin(n_q,n_{q_j},b_j,t_1,t_2)$. Thus, $Y'$ is not
closed under $r$. This contradicts the assumption that $Y'$ is
closed under  $(\pi^+)^{Y_{k+1}}$.

\item $u$ is the head of a rule $r$ of the form (\ref{pi1thtn3}).
Similar to the above case, we can prove that it violates the
assumption that $Y'$ is closed under  $(\pi^+)^{Y_{k+1}}$.

\item $u$ is the head of a rule $r$ of the form (\ref{pi1thtn11}).
Because $u \in Y_{k+1}$ we know that the body of $r$ is satisfied
by $Y_{k+1}$, and hence, $r$ belongs to $(\pi^+)^{Y_{k+1}}$.
Again, because of the closeness of $Y'$, we conclude that
$u \in Y'$ which violates the assumption that $u \not\in Y'$.

\end{itemize}

\ni The above cases imply that $Y'$ is not closed under
$(\pi^+)^{Y_{k+1}}$. Thus, our assumption that $Y_{k+1}$ is not
minimal is incorrect. Together with the closeness of $Y_{k+1}$, we
have that $Y_{k+1}$ is indeed an answer set of $\pi^+$. The
inductive step is proved since $Y_{k+1}$ satisfies (*) for every
program $q$ with $\mu(q) = k+1$. This proves the lemma. \qed

\begin{theorem}
{\rm Let $(D,\Gamma)$ be a consistent action theory and $p$ be a
general program. Then,
\begin{itemize}
\item[(i)] for every  answer set $M$ of $\;\Pi^{HTN}_n$ with
$occ(a_i,i) \in M$ for $i \in \{0,\ldots,n-1\}$, $\,s_0(M)a_0
\ldots a_{n-1} s_n(M)\,$ is a trace of $p$; and

\item[(ii)] if $\,s_0a_0\ldots a_{n-1}s_n\,$ is a trace
of $p$ then there exists an answer set $M$ of $\;\Pi^{HTN}_n$ such
that $\,s_j = s_j(M)$ and $\;occ(a_i,i) \in M$ for $j \in
\{0,\ldots,n\}$ and $\,i \in \{0,\ldots,n-1\}$ and $trans(n_p,0,n)
\in M$.
\end{itemize}
}
\end{theorem}
\proof (i) follows from Lemma  \ref{lemtrace3} and (ii) follows
from Lemma \ref{lemtrace4}. \qed

\section*{Appendix B - Splitting Theorem}

In this appendix, we review the basics of the Splitting Theorem
\cite{lif94a}.
Because programs in this paper do not contain classical negations,
some of the definitions have been modified from the original
presentation in \cite{lif94a}.

Let $r$ be a rule%
\[
a_0 \la a_1,\ldots,a_m,\naf a_{m+1},\ldots, a_n.
\]
By $head(r)$, $body(r)$, and $lit(r)$ we denote
$a_0$, $\{a_1,\ldots,a_n\}$, and
$\{a_0, a_1,\ldots,a_n\}$, respectively.
$pos(r)$ and $neg(r)$ denote the set
$\{a_1,\ldots,a_m\}$ and $\{a_{m+1},\ldots,a_n\}$,
respectively.

\st
For a program $\Pi$ over the language $\call{\cal P}$,
a set of atoms of $\call{\cal P}$,
$A$, is a splitting set of $\Pi$ if for every rule $r \in \Pi$,
if $head(r) \in  A$ then $lit(r) \subseteq A$.

\st
Let $A$ be a splitting set of $\Pi$. The {\em bottom of $\Pi$
relative to A}, denoted by $b_A(\Pi)$, is the program consisting
of all rules $r \in \Pi$ such that $lit(r) \subseteq A$.

\st
Given a splitting set $A$ for $\Pi$,
and a set $X$ of atoms from $lit(b_A(\Pi))$,
the {\em partial  evaluation of $\Pi$ by X
with respect to A}, denoted by $e_A(\Pi,X)$,
is the program obtained from $\Pi$ as follows.
For each rule $r \in \Pi \setminus b_A(\Pi)$ such that

\begin{enumerate}

\item $pos(r) \cap A \subseteq X$;

\item $neg(r) \cap A$ is disjoint from $X$;

\end{enumerate}

\ni
we create a rule $r'$ in $e_A(\Pi,X)$ such that

\begin{enumerate}

\item $head(r') = head(r)$, and

\item $pos(r') = pos(r)   \setminus A$,

\item $neg(r') = neg(r)    \setminus A$.
\end{enumerate}

\ni
Let $A$ be a splitting set of $\Pi$. A {\em solution to $\Pi$
with respect to A} is a pair $\langle X,Y \rangle$ of sets of
atoms satisfying the following two properties:

\begin{enumerate}

\item $X$ is an answer set of $b_A(\Pi)$; and

\item $Y$ is an answer set of $e_A(\Pi \setminus b_A(\Pi),X)$;

\end{enumerate}

\ni
The splitting set theorem is as follows.

\begin{theorem}
[Splitting Set Theorem, \cite{lif94a}] \label{spl1} Let $A$ be a
splitting set for a program $\Pi$. A set $S$ of atoms is a
consistent answer set of $\Pi$ iff $S = X \cup Y$ for some
solution $\langle X,Y \rangle$ to $\Pi$ with respect to $A$. \qed
\end{theorem}

\ni A {\em sequence} is a family whose index set is an initial
segment of ordinals $\{\alpha \ | \ \alpha < \mu\}$. A sequence
$\langle A_\alpha \rangle_{\alpha < \mu}$ of sets is {\em monotone
} if $A_\alpha \subseteq A_\beta$ whenever $\alpha < \beta$, and
{\em continuous} if, for each limit ordinal $\alpha < \mu$,
$A_\alpha = \bigcup_{\gamma< \alpha} A_\gamma$.

\st A {\em splitting sequence} for a program $\Pi$ is a nonempty,
monotone, and continuous sequence $\langle A_\alpha
\rangle_{\alpha < \mu}$ of splitting sets of $\Pi$ such that
$lit(\Pi) = \bigcup_{\alpha < \mu} A_\alpha$.

\st Let $\langle A_\alpha \rangle_{\alpha < \mu}$ be a splitting
sequence of the program $\Pi$. A {\em solution to $\Pi$ with
respect to A} is a sequence $\langle E_\alpha \rangle_{\alpha <
\mu}$ of set of atoms satisfying the following conditions.

\begin{enumerate}

\item $E_0$ is an answer set  of the program $b_{A_0}(\Pi)$;

\item for any $\alpha$ such that $\alpha + 1 < \mu$, $E_{\alpha+1}$
is an answer set for $e_{A_\alpha}(b_{A_{\alpha+1}}(\Pi) \setminus
b_{A_\alpha}(\Pi), \bigcup_{\gamma \le \alpha}E_\gamma)$; and

\item For any limit ordinal $\alpha < \mu$,
$E_\alpha = \emptyset$.

\end{enumerate}

\ni
The splitting set theorem is generalized for splitting sequence next.

\begin{theorem}
[Splitting Sequence Theorem, \cite{lif94a}] \label{spl2} Let $A =
\langle A_\alpha \rangle_{\alpha < \mu}$ be a splitting sequence
of the program $\Pi$. A set of atoms $E$ is an  answer
set of $\Pi$ iff $E = \bigcup_{\alpha < \mu} E_\alpha$ for some
solution $\langle E_\alpha \rangle_{\alpha < \mu}$ to $\Pi$ with
respect to $A$. \qed
\end{theorem}

\ni
To apply Theorems \ref{spl1}-\ref{spl2} to programs with
constraints of the form (\ref{lprule2}), we need to modify
the notation of the bottom of a program relative to a set
of atoms as follows.

Let $\Pi = \Pi_1 \cup \Pi_2$ be a program with constrains where
$\Pi_1$ is a set of rules of the form (\ref{lprule1})
and $\Pi_2$ is a set of rules of the form (\ref{lprule2}).
For a splitting set $A$ of $\Pi$, we define
$b_A(\Pi) = b_A(\Pi_1) \cup c_A(\Pi_2)$ where
$c_A(\Pi_2) = \{r \mid r \in \Pi_2,\; lit(r) \subseteq A\}$.

We can prove that Theorems \ref{spl1}-\ref{spl2} hold for
programs with constraints. For example, if $A$ is
a splitting of the program $\Pi$, then
$S$ is an answer set of $\Pi$ iff $A = X \cup Y$ where
$X$ is an answer set of $b_A(\Pi)$ and
$Y$ is an answer set of $e_A(\Pi \setminus b_A(\Pi), X)$.
The proof of the modified theorems is based on two
observations: (i) a set $A$ of atoms from $lit(\Pi)$ is a splitting
set of $\Pi$ iff it is a splitting set of $\Pi_1$ (because
$\bot \not\in lit(\Pi)$); and
(ii) a set of atoms $S$ is an answer set of $\Pi$ iff
$S$ is answer set of $\Pi_1$ and $S$ satisfies the rules
of $\Pi_2$.

\begin{received}
Received July 2002;
revised December 2003;
accepted April 2004
\end{received}

\end{document}